\documentclass[reqno, onecolumn, 10pt]{wlscirep}
\usepackage[utf8]{inputenc}
\usepackage[T1]{fontenc}
\usepackage{quotes}
\usepackage{sidecap}
\usepackage{comment}
\sidecaptionvpos{figure*}{t}

\usepackage{graphicx}
\usepackage{multirow}
\usepackage{amsmath,amssymb,amsfonts}
\usepackage{amsthm}
\usepackage{mathrsfs}
\usepackage[title]{appendix}
\usepackage{textcomp}
\usepackage{manyfoot}
\usepackage{booktabs}
\usepackage{algorithm}
\usepackage{algorithmic}
\usepackage{listings}
\usepackage{placeins}
\usepackage{lmodern}
\usepackage{fix-cm}
\usepackage{color}
\usepackage{subcaption}
\usepackage{bbding}
\usepackage{float}

\usepackage[numbers,square]{natbib}
\setcitestyle{square,numbers}
\usepackage{threeparttable}

\definecolor{dirtyorange}{RGB}{230, 126, 34}
\definecolor{dirtygreen}{RGB}{100, 180, 0}
\definecolor{dirtyblue}{RGB}{0, 100, 180}

\newcommand{\dirtycite}[1]{%
  \textcolor{dirtyorange}{%
    \if\relax\detokenize{#1}\relax
      [CITATION]%
    \else
      [CITATION: #1]%
    \fi
  }%
}

\newcommand{\dirtyref}[1]{%
  \textcolor{dirtygreen}{%
    \if\relax\detokenize{#1}\relax
      [REF]%
    \else
      [REF: #1]%
    \fi
  }%
}

\newcommand{\dirtyfig}[1]{%
  \textcolor{dirtyblue}{%
    \if\relax\detokenize{#1}\relax
      [REF]%
    \else
      [FIGURE: #1]%
    \fi
  }%
}

\newcommand{\dirtytab}[1]{%
  \textcolor{dirtyblue}{%
    \if\relax\detokenize{#1}\relax
      [REF]%
    \else
      [TABLE: #1]%
    \fi
  }%
}

\title{Toward a Graded Measure of Belief Stability in Large Language Models}

\author[1$\sharp$]{Samantha Dies}
\author[1]{Branden Fitelson}
\author[1,2]{Tina Eliassi-Rad}

\affil[1]{Northeastern University, 360 Huntington Ave, Boston, MA 02115 USA}

\affil[2]{Santa Fe Institute, 1399 Hyde Park Road, Santa Fe, NM 87501 USA}

\affil[$\sharp$]{\href{mailto:dies.s@northeastern.edu}{dies.s@northeastern.edu}}

\begin{abstract}
Large language models (LLMs) increasingly mediate how people access and reason with information, yet factual reliability is usually evaluated one judgment at a time. We introduce \textbf{graded belief stability}, a relational measure of how well a belief persists within an LLM's broader belief system. Unlike individual belief probability, it asks whether support for a claim persists when that claim is considered alongside the model's other epistemic commitments. We operationalize this idea with a Direct Conditional estimator that uses internal model representations to estimate conditional belief probabilities. Across $12$ LLMs and three domains, lower-stability beliefs exhibit greater mean behavioral movement under conversational challenge in $83.3\%$ of model–domain settings after matching on individual belief probability. Graded belief stability therefore extends reliability assessment beyond how strongly an LLM supports a claim to how robustly that belief is supported within its broader system of beliefs.
\end{abstract}
\begin{document}

\flushbottom
\maketitle
\thispagestyle{empty}

%%%%%%%%%%%%%%%%%%%%%%%%%%%%%%%%%%%%%%%%%%%%
%%              Main Text                 %%
%%%%%%%%%%%%%%%%%%%%%%%%%%%%%%%%%%%%%%%%%%%%

%%%%%%%%%%%%%%%%%%%%%%%%%%%%%%%%%%%%%%%%%%%%
%%             INTRODUCTION               %%
%%%%%%%%%%%%%%%%%%%%%%%%%%%%%%%%%%%%%%%%%%%%

\section{Introduction}
\label{sec:introduction}

Large language models (LLMs) increasingly impact how people access and reason with information, making the reliability of their factual judgments central to their use~\cite{alkhamissi2022review, augenstein2024factuality}. Existing evaluations characterize several dimensions of this reliability, including accuracy, uncertainty, and calibration~\cite{band_calibration, kapoor2024large, yadkori2024believe, steyvers2025large}. In interactive settings, however, factual claims are rarely considered in isolation. New claims, corrections, and contextual information accumulate over a conversation, and models may revise answers they previously supported. Reliability therefore also depends on a different question: \textit{which of a model's beliefs persist as the informational context around them changes?}

Such belief revisions are well documented. LLM factual judgments can vary with paraphrasing and changes in semantic context~\cite{elazar2021measuring, gao-etal-2024-spuq, dies2025}, as well as under user disagreement, contradictory feedback, and multi-turn interaction~\cite{sharma2024towards, kumaran2026competing, li2025firm, sarkar2026language, cheon2026robust}. Work on systematic belief consistency and knowledge editing likewise suggests that factual commitments do not behave as isolated entries in a knowledge base: accepting or revising one proposition can affect related propositions~\cite{kassner2021beliefbank, cohen2024evaluating}. Consider an LLM that assigns similar support to ``Paris is in France'' and ``Oslo is in Norway.'' If it also considers ``Oslo is in Sweden'' plausible, the Oslo claim sits alongside an incompatible proposition the model has not ruled out. The Paris and Oslo claims may therefore look equally strong on their own even though one is more fragile within the model's broader belief system. We refer to this difference as \emph{belief stability}.

Existing work captures important pieces of this picture without directly measuring it. Representation-based studies find that veracity and epistemic uncertainty are systematically reflected in LLM hidden states~\cite{marks2024geometry, ahdritz24a_knowable, yin24c_truthfulness, savcisens2025trilemma, ying2026truthfulness}, while LLMs do not reliably distinguish epistemic concepts such as belief, knowledge, and fact~\cite{suzgun2025language}. Behavioral evaluations, meanwhile, reveal when outputs change under particular external perturbations~\cite{elazar2021measuring, sharma2024towards, kumaran2026competing, li2025firm}. What these approaches do not capture is how a belief is situated relative to the model's \textit{other} epistemic commitments. A belief may receive strong support on its own yet become fragile in light of other propositions the model regards as plausible. Stability is therefore a relational property of a belief within a broader belief system.

Formal epistemology provides a principled basis for making this distinction precise. The Lockean Thesis connects graded credence to categorical belief through a probability threshold, with sufficiently probable propositions counting as beliefs~\cite{foley1992epistemology, leitgeb2014stability}. Leitgeb's Humean Thesis strengthens this threshold-based approach by requiring a belief to remain sufficiently probable when conditioned on propositions an agent does not disbelieve~\cite{leitgeb_humean}. We adapt this idea to LLMs by defining the \textbf{graded belief stability} of a believed proposition $P$ as the fraction of the model's non-disbeliefs $x$ for which $\Pr(P\mid x)$ remains above its belief threshold. Graded stability thus distinguishes how strongly a model supports $P$ on its own from how broadly that support persists across its broader epistemic state.

Estimating graded stability requires eliciting conditional belief probabilities from LLM representations. Supervised probes can recover veracity-related information from hidden states~\cite{marks2024geometry, ahdritz24a_knowable, savcisens2025trilemma, corona-mendozza-sogaard-2026-llm}, but $\Pr(P\mid x)$ is not directly observable. We therefore introduce a trivalent \emph{Direct Conditional} estimator grounded in established accounts of indicative conditionals and non-bivalent probability~\cite{egre2026probability, cooper1968propositional}. For each belief--non-disbelief pair $(P,x)$, we construct a conditional statement of the form ``Given $x$, $P$'' and probe its hidden representation to estimate the conditional probabilities needed to compute graded stability.

We evaluate graded belief stability across $12$ instruction-tuned LLMs from four model families, ranging from $3$B to $72$B parameters, and across three domains: City Locations, Medical Indications, and Word Definitions. We test whether graded stability captures structure beyond individual belief probability and whether the underlying probability estimates are approximately probabilistically coherent. We then characterize its variation across semantic domains and evaluate whether it is associated with resistance to conversational challenge.

Our work makes five core contributions:
\begin{enumerate}
    \item \textbf{Formalization:} We introduce \textit{graded belief stability} as a statement-level measure of how well an LLM's belief persists across its broader belief system.

    \item \textbf{Measurement:} We develop a representation-based Direct Conditional estimator of conditional belief probability for computing graded stability.

    \item \textbf{Validation:} We show that graded stability captures systematic information beyond individual belief probability, where $99.0\%$ of model-pair residual correlations are positive. The measured probability systems are also approximately coherent, with median distance from coherence ranging from $0.015$ to $0.062$ across domains.

    \item \textbf{Characterization:} We find that City Locations is consistently the most stable domain, with a median across models of mean graded stability of $0.88$, compared with $0.71$ for Medical Indications and $0.74$ for Word Definitions.

    \item \textbf{Behavioral validity:} Among beliefs matched on individual belief probability, lower-stability beliefs show greater mean movement under repeated conversational challenge for $83.3\%$ of model--domain settings.
\end{enumerate}

This work captures a dimension of LLM reliability that individual belief probability alone cannot. Two factual judgments can receive similar support while occupying very different positions within the model's broader belief system, making one substantially more fragile than the other. Graded belief stability quantifies this relational property, extending reliability assessment beyond \textit{how strongly} an LLM supports a claim to \textit{how robustly that belief is supported within the model's broader system of beliefs}.
%%%%%%%%%%%%%%%%%%%%%%%%%%%%%%%%%%%%%%%%%%%%
%%                Results                 %%
%%%%%%%%%%%%%%%%%%%%%%%%%%%%%%%%%%%%%%%%%%%%

\begin{figure}[t!]
\centering
\includegraphics[width=\textwidth]{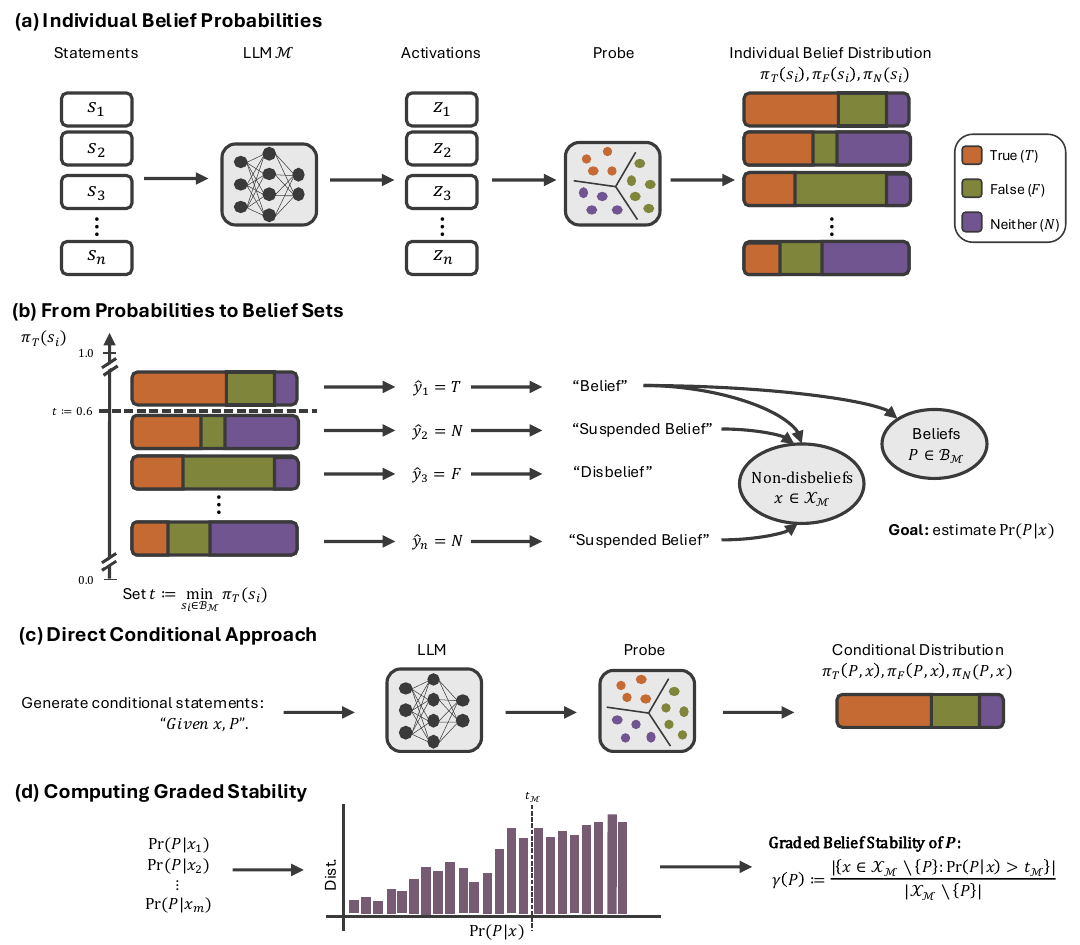}
\caption{
\textbf{Measuring graded belief stability in LLMs.}
\textbf{(a)} For each statement $s$, we extract an internal LLM representation and use a probe to estimate a distribution over \texttt{True} ($T$, orange), \texttt{False} ($F$, olive), and \texttt{Neither} ($N$, purple). \textbf{(b)} The probe's predicted states $\hat{y}$ determine whether each statement is treated as a belief ($\hat{y} = T$), disbelief ($\hat{y} = F$), or suspended belief ($\hat{y} = N$), yielding the belief set $\mathcal{B}_\mathcal{M}$ and non-disbelief set $\mathcal{X}_\mathcal{M}$. We define the empirical belief threshold $t_\mathcal{M}$ as the minimum \texttt{True}-class probability among statements classified as beliefs. We then estimate $\Pr_\mathcal{M}(P\mid x)$ for each $P\in\mathcal{B}_\mathcal{M}$ and $x\in\mathcal{X}_\mathcal{M}$. \textbf{(c)} The Direct Conditional approach probes representations of conditional statements of the form \textit{``Given $x$, $P$.''}, and \textbf{(d)} we compute graded stability $\gamma_\mathcal{M}(P)$ as the fraction of conditioning propositions for which the conditional probability of $P$ remains above $t_\mathcal{M}$.
}
\label{fig:methods}
\end{figure}

\section{Results}
\label{sec:results}

We begin by formalizing graded belief stability $\gamma$ and introducing the Direct Conditional representation-based estimator for conditional belief. We then validate the measure, showing that it captures structure beyond individual belief probability and is approximately probabilistically coherent. We next characterize variation in $\gamma$ across domains, before testing whether graded stability distinguishes behavioral resilience among beliefs matched on individual belief probability.

%%%%%%%%%%%%%%%%%%%%%%%%%%%%%%%%%%%%%%%%%%%%
%%         Formalizing \gamma(P)          %%
%%%%%%%%%%%%%%%%%%%%%%%%%%%%%%%%%%%%%%%%%%%%

\subsection{Formalizing graded belief stability}
\label{sec:formalizing}

We define graded belief stability by adapting the stability theory of belief from formal epistemology, in which a belief is stable when it persists after conditioning on propositions the agent does not reject~\cite{leitgeb2014stability}. We extend this binary notion to a statement-level measure of the degree to which an LLM's belief persists across conditioning contexts.

Let $\mathcal{M}$ denote an LLM and let $\Pr_\mathcal{M}(s)$ denote the probability that $\mathcal{M}$ assigns to a statement $s$ being true. Following the Lockean Thesis, we take categorical belief to correspond to sufficiently high probability relative to a belief threshold $t_\mathcal{M}$~\cite{foley1992epistemology, leitgeb2014stability}. In our empirical implementation below, we first identify categorical belief from a trivalent probe and then instantiate $t_\mathcal{M}$ from the resulting belief set rather than imposing a universal cutoff. We denote the set of beliefs held by $\mathcal{M}$ as $\mathcal{B}_\mathcal{M}$ and the larger set of propositions that $\mathcal{M}$ does not disbelieve, including both beliefs and suspended beliefs, as $\mathcal{X}_\mathcal{M}$.

The Humean Thesis strengthens this static notion of belief by requiring beliefs to remain sufficiently probable when considered alongside the agent's other non-disbeliefs~\cite{leitgeb2014stability,leitgeb_humean}. Intuitively, a stable belief should remain above the belief threshold even when considered alongside propositions the model itself does not reject. Specifically, a belief $P \in \mathcal{B}_\mathcal{M}$ is considered \emph{stable} if conditioning on any $x \in \mathcal{X}_\mathcal{M} \setminus \{P\}$ does not lower its probability below the belief threshold:
\begin{equation}
    P \text{ is stable} 
    \iff 
    \Pr\nolimits_\mathcal{M}(P\mid x) > t_\mathcal{M} \quad \forall x\in\mathcal{X}_\mathcal{M} \setminus \{P\}.
\end{equation}
Thus, propositions with similar $\Pr_\mathcal{M}(P)$ may nevertheless differ substantially in stability.

Binary stability, however, treats all violations identically. A proposition that falls below the belief threshold under a single conditioning statement is indistinguishable from one that does so under nearly all of them. We therefore define the \textbf{graded belief stability} of $P$ as
\begin{equation}
\label{eq:graded_stability}
    \gamma_\mathcal{M}(P)
    :=
    \frac{
    \left|\left\{
    x \in \mathcal{X}_\mathcal{M} \setminus \{P\}
    :
    \Pr_\mathcal{M}(P\mid x) > t_\mathcal{M}
    \right\}\right|
    }{
    \left|\mathcal{X}_\mathcal{M} \setminus \{P\}\right|
    }.
\end{equation}
Thus, $\gamma_\mathcal{M}(P)\in[0,1]$ is the proportion of non-disbeliefs under which belief in $P$ is preserved. Higher values indicate beliefs that persist more broadly across the model's belief system, distinguishing graded stability from the probability assigned to $P$ in isolation.

%%%%%%%%%%%%%%%%%%%%%%%%%%%%%%%%%%%%%%%%%%%%
%%          Measuring \gamma(P)           %%
%%%%%%%%%%%%%%%%%%%%%%%%%%%%%%%%%%%%%%%%%%%%

\subsection{Measuring graded belief stability in LLMs}
\label{sec:measuring}

Measuring graded stability requires identifying an LLM's beliefs and non-disbeliefs and estimating $\Pr_\mathcal{M}(P\mid x)$ for belief--non-disbelief pairs. We use calibrated supervised probes over LLM hidden representations to obtain these probability estimates, since unconstrained prompted judgments need not yield calibrated probability estimates and can be sensitive to prompt formulation and conversational context~\cite{wolf2026partition, elazar2021measuring, sharma2024towards, li2025firm}.

For each statement $s_i$, we extract hidden representations across layers $\ell$ of model $\mathcal{M}$ and train the sparse-aware multiple-instance learning (\texttt{sAwMIL}) probe~\cite{savcisens2025trilemma}, which models \texttt{True} ($T$), \texttt{False} ($F$), and \texttt{Neither} ($N$) as distinct veracity classes.\footnote{Full activation extraction, probe training, and layer-selection procedures are described in Sec.~\ref{sec:methods:models_data}, and layer-wise probe performance is reported in Sec.~\ref{sec:si:layer_selection}. We additionally reproduce the analyses using Support Vector Machine (\texttt{SVM})~\cite{cortes1995support} and \texttt{Mass Mean}~\cite{marks2024geometry} probes in Sec.~\ref{sec:si:other_probes}.} After calibration, the probe returns the trivalent distribution
\begin{equation}
    \boldsymbol{\pi}_{\mathcal{M}}(s_i)
    =
    \left[
    \pi_T(s_i),
    \pi_F(s_i),
    \pi_N(s_i)
    \right],
\end{equation}
where $\pi_c(s_i)$ is the probe-assigned probability of class $c$. The predicted epistemic state is then
\begin{equation}
    \hat y_i
    =
    \arg\max_{c\in\{T,F,N\}}
    \pi_c(s_i).
\end{equation}
We interpret $\hat y_i=T$ as belief, $\hat y_i=F$ as disbelief, and $\hat y_i=N$ as suspension of belief (Fig.~\ref{fig:methods}\textbf{(a--b)}). The probe therefore determines the empirical belief and non-disbelief sets directly:
\begin{equation}
\label{eq:belief_sets}
    \mathcal{B}_\mathcal{M}
    =
    \{s_i:\hat y_i=T\},
    \qquad
    \mathcal{X}_\mathcal{M}
    =
    \{s_i:\hat y_i\in\{T,N\}\}.
\end{equation}
This trivalent operationalization retains an explicit state for suspended belief rather than forcing every proposition into a binary believed/not-believed distinction.

We then associate the probe-defined categorical belief set with an empirical Lockean-style threshold. For each model and domain, we define
\begin{equation}
\label{eq:belief_threshold}
t_\mathcal{M}
:=
\min_{s_i\in\mathcal{B}_\mathcal{M}}
\pi_T(s_i).
\end{equation}
Importantly, $t_\mathcal{M}$ does not determine which propositions belong to $\mathcal{B}_\mathcal{M}$. Rather, categorical belief is already defined by the probe's trivalent prediction in Equation~\eqref{eq:belief_sets}. Instead, the threshold translates this categorical belief set into a scalar criterion for belief persistence. Specifically, $t_\mathcal{M}$ is the most stringent \texttt{True}-class support threshold satisfied by every proposition that the probe identifies as a belief. This anchors the persistence criterion to the model's own empirical belief boundary rather than imposing an externally chosen probability cutoff.

Because $\pi_T$ is one component of a calibrated three-class distribution over \texttt{True}, \texttt{False}, and \texttt{Neither}, $t_\mathcal{M}$ will not necessarily exceed $0.50$, but is instead lower-bounded by $0.33$. The resulting threshold should therefore be understood as a model-, domain-, and probe-specific support floor induced by the empirical belief set, rather than as a universal normative probability threshold (Fig.~\ref{fig:methods}\textbf{(b)}, Sec.~\ref{sec:methods:atomic}).

We next estimate conditional belief probability for every belief $P\in\mathcal{B}_\mathcal{M}$ and non-disbelief $x\in\mathcal{X}_\mathcal{M}\setminus\{P\}$. The \textbf{Direct Conditional} approach represents $P\mid x$ using the conditional statement \textit{``Given $x$, $P$.''} and probes its internal representation directly (Fig.~\ref{fig:methods}\textbf{(c)}). A three-class \texttt{sAwMIL} probe predicts whether the resulting conditional is \texttt{True}, \texttt{False}, or \texttt{Neither} using Cooper--Cantwell training labels~\cite{cooper1968propositional,cantwell2008logic,egre2025certain}. For each pair $(P,x)$, the probe returns the trivalent distribution
\begin{equation}
\label{eq:direct_distribution}
    \boldsymbol{\pi}^{\mathrm{direct}}_\mathcal{M}(P,x)
    =
    \left[
    \pi_T^{\mathrm{direct}}(P,x),
    \pi_F^{\mathrm{direct}}(P,x),
    \pi_N^{\mathrm{direct}}(P,x)
    \right].
\end{equation}
We use the conditional's \texttt{True}-class mass to define our Direct Conditional estimator
\begin{equation}
\label{eq:direct_estimate}
    \widehat{\Pr}^{\,\mathrm{direct}}_\mathcal{M}(P\mid x)
    :=
    \pi_T^{\mathrm{direct}}(P,x).
\end{equation}
We use $\widehat{\Pr}^{\,\mathrm{direct}}_\mathcal{M}(P\mid x)$ as an empirical operational estimator of conditional belief probability rather than as an algebraic reconstruction of the normalized Cooper--Cantwell probability of a trivalent conditional. In particular, we do not renormalize over the \texttt{True} and \texttt{False} states. This keeps conditional support on the same three-class scale used to define the empirical belief threshold $t_\mathcal{M}$ and allows probability mass assigned to \texttt{Neither} to count against persistence of the original belief. Section~\ref{sec:methods:conditional} motivates this choice and contrasts it with our complementary Joint-to-Conditional estimator, which instead normalizes over determinate conditional states. Results on the secondary Joint-to-Conditional estimator are reported in Section~\ref{sec:si:joint_results}.

Replacing $\Pr_\mathcal{M}(P\mid x)$ in Eq.~\eqref{eq:graded_stability} with the Direct Conditional estimate in Eq.~\eqref{eq:direct_estimate} yields our empirical Direct Conditional stability score $\gamma_\mathcal{M}^{\mathrm{direct}}(P)$ (Fig.~\ref{fig:methods}\textbf{(d)}; Sec.~\ref{sec:methods:stability}). We apply this probing-based framework to $12$ instruction-tuned\footnote{Matched base-model analyses are reported in Section~\ref{sec:si:base_robustness}.} LLMs spanning four model families and ranging from $3$B to $72$B parameters. We evaluate three domains with differing epistemic characteristics: \textbf{City Locations}, consisting of comparatively clear-cut factual relations; \textbf{Medical Indications}, requiring specialized factual knowledge; and \textbf{Word Definitions}, containing comparatively ambiguous lexical relations~\cite{savcisens2025trilemma}.

%%%%%%%%%%%%%%%%%%%%%%%%%%%%%%%%%%%%%%%%%%%%
%%         Validating \gamma(P)           %%
%%%%%%%%%%%%%%%%%%%%%%%%%%%%%%%%%%%%%%%%%%%%

\begin{figure}[t]
\centering
\includegraphics[width=\textwidth]{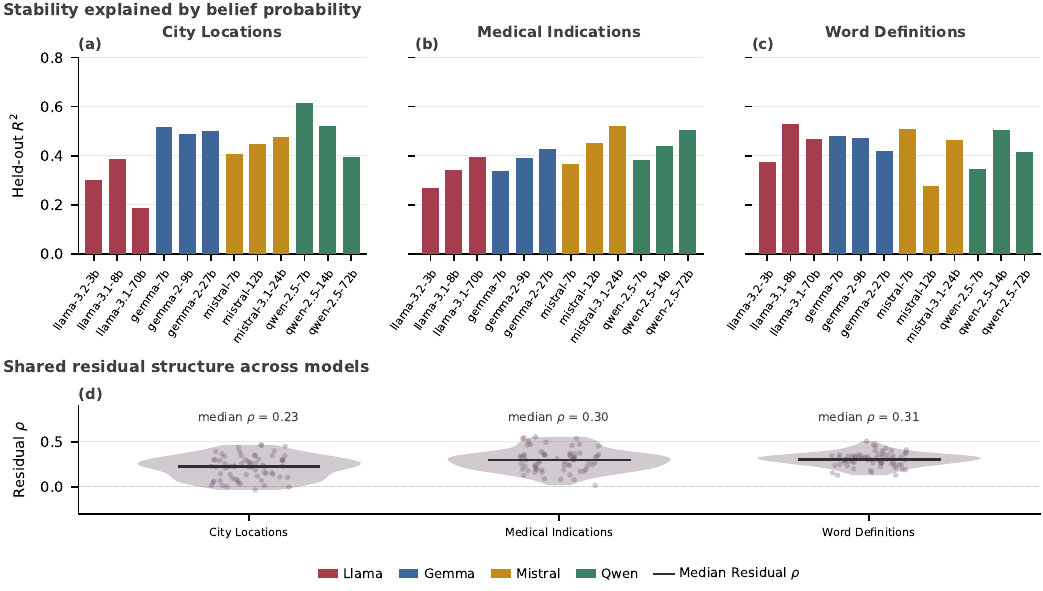}
\caption{\textbf{Relationship between individual belief probability and graded stability.}
We show the held-out $R^2$ for predicting graded stability $\gamma_\mathcal{M}(P)$ from individual belief probability $\pi_T(P)$ for \textbf{(a)} City Locations, \textbf{(b)} Medical Indications, and \textbf{(c)} Word Definitions across \texttt{Llama} (red), \texttt{Gemma} (blue), \texttt{Mistral} (yellow), and \texttt{Qwen} (green) model families. We also report \textbf{(d)} distributions of pairwise Spearman correlations $\rho$ between LLMs' residual graded stability within each domain, where black lines denote the median correlation across model pairs. Individual belief probability predicts a meaningful proportion of graded stability, but graded stability captures proposition-level variation beyond individual belief probability that is systematically shared across models.}
\label{fig:stability_vs_credence}
\end{figure}

\begin{figure}[t]
\centering
\includegraphics[width=\textwidth]{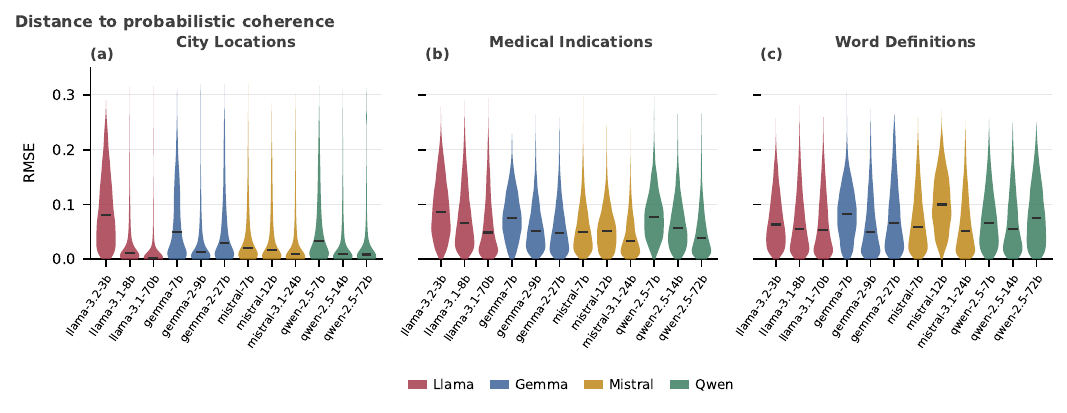}
\caption{
\textbf{Distance of measured probabilities from probabilistic coherence.} Violins show the distribution across $(P,x)$ pairs of the root-mean-square (RMS) adjustment $d_\mathrm{CCK}$ required to project the measured probability distributions for individual statements and conditionals onto the nearest CCK-coherent probability system for \textbf{(a)} City Locations, \textbf{(b)} Medical Indications, and \textbf{(c)} Word Definitions for \texttt{Llama} (red), \texttt{Gemma} (blue), \texttt{Mistral} (yellow), and \texttt{Qwen} (green) model families. Horizontal black lines denote within-model medians. Smaller values indicate greater probabilistic coherence, with $d_\mathrm{CCK}=0$ corresponding to exact coherence. Although exact coherence is uncommon, only modest adjustments are generally required to reconcile the independently measured probabilities with a coherent trivalent probability system.
}
\label{fig:coherence}
\end{figure}

\subsection{Validating graded stability}
\label{sec:validating}

We validate graded stability along two complementary dimensions: whether it captures information beyond individual belief probability and whether the probability estimates underlying it are approximately probabilistically coherent. Main-text results use the Direct Conditional estimator and the \texttt{sAwMIL} probe. Joint-to-Conditional results and analyses using the \texttt{SVM} and \texttt{Mass Mean} probes appear in Section~\ref{sec:si:robustness}.

\subsubsection{Graded stability captures information beyond belief probability}

A central motivation for graded stability is that it captures how a belief behaves in the context of an LLM's broader belief system rather than simply how strongly that belief is held in isolation. We therefore first ask how well the individual belief probability $\pi_T(P)$ predicts $\gamma_\mathcal{M}(P)$ for beliefs $P \in \mathcal{B}_\mathcal{M}$.

If graded stability were simply a transformation of belief probability, then a sufficiently flexible function of $\pi_T(P)$ should accurately predict $\gamma_\mathcal{M}(P)$ for beliefs not used to fit that function. For each model and domain, we attempt to capture such a function using a natural cubic spline,
\begin{equation}
\label{eq:stability_probability_spline}
    \widehat{\gamma}_{\mathcal{M}}(P)
    =
    f_\mathcal{M}\!\left(\pi_T(P)\right),
\end{equation}
with three degrees of freedom. The spline allows belief probability and stability to have a smooth nonlinear relationship without assuming a linear relationship. We use five-fold cross-validation to generate out-of-sample predictions $\widehat{\gamma}_{\mathcal{M}}(P)$ and evaluate performance with $R^2$ (Sec.~\ref{sec:methods:stability_credence}). We fix the spline complexity across all models and domains and show that the results are qualitatively unchanged across alternative spline complexities in Section~\ref{sec:si:stability_probability_robustness}.

Individual belief probability consistently predicts a meaningful but incomplete proportion of the observed variation in graded stability (Fig.~\ref{fig:stability_vs_credence}\textbf{(a--c)}). The median held-out $R^2$ across models is $0.46$ for City Locations, $0.39$ for Medical Indications, and $0.47$ for Word Definitions. Thus, although belief probability is informative about stability, more than half of the proposition-level variation in $\gamma_\mathcal{M}(P)$ remains unexplained by the probability assigned to $P$ in isolation. Graded stability therefore captures substantial variation beyond individual belief strength.

We next ask whether this unexplained variation is systematic. For each proposition $P$, we define its \textit{residual stability} as
\begin{equation}
\label{eq:residual_stability}
    r_\mathcal{M}(P)
    :=
    \gamma_\mathcal{M}(P)
    -
    \widehat{\gamma}_\mathcal{M}(P),
\end{equation}
where positive values identify propositions that are more stable than predicted from their individual belief probability and negative values identify propositions that are less stable than predicted. We align pairs of LLMs on propositions believed by both models and compute the Spearman correlation between their residual stability scores (Fig.~\ref{fig:stability_vs_credence}\textbf{(d)}).

Across domains, $99.0\%$ of model-pair residual correlations are positive, with median correlations of $\rho=0.23$ for City Locations, $\rho=0.30$ for Medical Indications, and $\rho=0.31$ for Word Definitions. All three medians exceed a proposition-shuffled permutation null ($p=0.001$; Sec.~\ref{sec:methods:stability_credence}). Models therefore tend to find the \textit{same} propositions unusually stable or fragile even after accounting for how strongly they believe those propositions in isolation. This suggests that the variation unique to graded stability contains reproducible proposition-level structure rather than only model-specific measurement noise. Because models are evaluated on the same benchmark propositions using a common probing framework, however, this analysis does not by itself distinguish model-internal relational structure from proposition-level or measurement factors shared across models.

\subsubsection{Graded stability probability estimates are approximately coherent}

Because graded stability is defined within a probabilistic framework, we next ask whether the independently measured individual and conditional probability distributions are pairwise compatible. Under Cooper--Cantwell--Kleene (CCK) trivalent probability, the distributions for $P$, $x$, and $P\mid x$ are coherent when they can all be generated by a single latent $3\times3$ probability distribution over the trivalent states of $(P,x)$. For each measured pair, we collect these probabilities into the nine-dimensional vector
\begin{equation}
\label{eq:cck_measurement_vector}
    \mathbf{v}_{P,x}
    =
    \left[
    \boldsymbol{\pi}_\mathcal{M}(P),
    \boldsymbol{\pi}_\mathcal{M}(x),
    \boldsymbol{\pi}^{\mathrm{direct}}_\mathcal{M}(P,x)
    \right].
\end{equation}
Under CCK semantics, such a latent joint distribution exists if and only if the measured probabilities satisfy
\begin{equation}
\label{eq:cck_constraints}
    \pi_T^\mathrm{direct}(P,x) \leq \pi_T(P),
    \qquad
    \pi_F^\mathrm{direct}(P,x) \leq \pi_F(P),
    \qquad
    \pi_F(x) \leq \pi_N^\mathrm{direct}(P,x),
    \qquad
    \pi_N^\mathrm{direct}(P,x) \leq \pi_N(P) + \pi_F(x).
\end{equation}
We use these constraints to test exact coherence separately for every $(P,x)$ pair (Sec.~\ref{sec:methods:probabilistic_coherence}); a proof of the coherence conditions is provided in Sec.~\ref{sec:si:coherence_proof}. Exact coherence is extremely rare: across all model--domain combinations, no more than $0.21\%$ of measured pairs satisfy all four constraints exactly. Full exact-coherence rates and individual constraint violations are reported in Section~\ref{sec:si:coherence}.

Because exact coherence is a binary criterion that treats arbitrarily small and large violations identically, we additionally quantify each pair's distance from the set of coherent probability systems. Following prior work that characterizes approximate coherence in terms of distance to the nearest coherent credence function~\cite{molinari2024accuracy}, we extend this distance-based approach to the CCK trivalent setting and define
\begin{equation}
\label{eq:cck_distance}
    d_\mathrm{CCK}(\mathbf{v}_{P,x})
    :=
    \sqrt{
        \frac{1}{9}
        \left\|
        \mathbf{v}_{P,x}
        -
        \Pi_\mathcal{C}(\mathbf{v}_{P,x})
        \right\|_2^2
    },
\end{equation}
where $\mathcal{C}$ is the set of CCK-coherent probability vectors and $\Pi_\mathcal{C}(\mathbf{v})$ is the closest vector in this set to the observed probability vector $\mathbf{v}$.\footnote{Full optimization and convergence details are provided in Sec.~\ref{sec:methods:probabilistic_coherence}.} Thus, $d_\mathrm{CCK}=0$ corresponds to exact coherence, while a value such as $d_\mathrm{CCK}=0.05$ means that reaching the nearest coherent probability system requires a root-mean-square (RMS) adjustment of $0.05$ per probability component.

Across models, the median model-level $d_\mathrm{CCK}$ is $0.015$ for City Locations, $0.052$ for Medical Indications, and $0.062$ for Word Definitions, corresponding to RMS adjustments of $1.5$, $5.2$, and $6.2$ percentage points per probability component, respectively. Thus, despite near-zero exact-coherence rates, the measured probability systems typically require quantitatively modest adjustments to reach the nearest coherent CCK representation. Because graded stability thresholds these probabilities at $t_\mathcal{M}$, however, a small component-wise RMS adjustment does not imply that the corresponding absolute stability score would be unchanged after projection. We therefore interpret $d_\mathrm{CCK}$ as a diagnostic of the measured probability system rather than as a robustness bound on $\gamma_\mathcal{M}(P)$.

%%%%%%%%%%%%%%%%%%%%%%%%%%%%%%%%%%%%%%%%%%%%
%%        Characterizing \gamma(P)        %%
%%%%%%%%%%%%%%%%%%%%%%%%%%%%%%%%%%%%%%%%%%%%

\begin{figure}[t!]
\centering
\includegraphics[width=\textwidth]{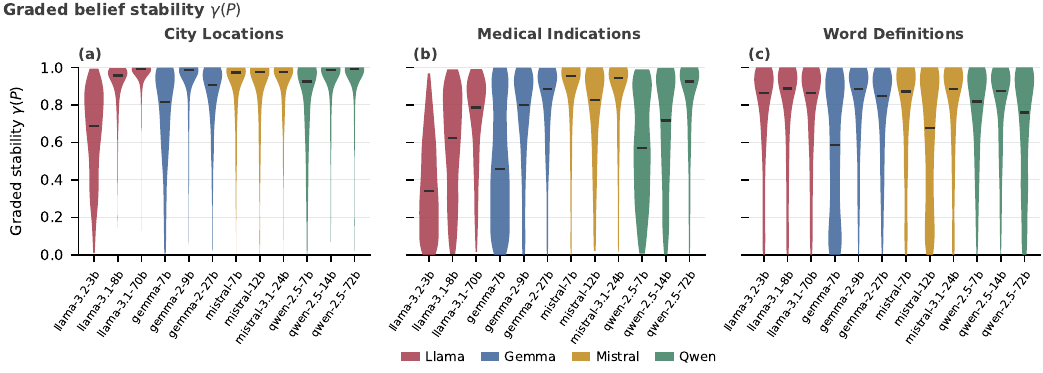}
\caption{
\textbf{Variation in graded belief stability across domains.}
Violin plots show the proposition-level distribution of graded stability $\gamma^{\mathrm{direct}}_\mathcal{M}(P)$ for each of the $12$ instruction-tuned LLMs in \textbf{(a)} City Locations, \textbf{(b)} Medical Indications, and \textbf{(c)} Word Definitions for \texttt{Llama} (red), \texttt{Gemma} (blue), \texttt{Mistral} (yellow), and \texttt{Qwen} (green) model families. Horizontal black lines denote within-model medians. Stability distributions differ substantially across the three benchmark domains, with City Locations exhibiting the highest median stability for $11$ of the $12$ models.
}
\label{fig:stability_characterization}
\end{figure}

\subsection{Graded belief stability varies with domain}
\label{sec:results:stability_variation}

Having established that graded stability captures information beyond individual belief probability and is approximately probabilistically coherent, we next ask how stability varies across beliefs and models by characterizing the distribution of $\gamma_\mathcal{M}(P)$ across each model's beliefs (Fig.~\ref{fig:stability_characterization}). Stability differs substantially across semantic domains. Across models, the median of model-level mean graded stability is $0.88$ for City Locations, compared with $0.71$ for Medical Indications and $0.74$ for Word Definitions. Comparing the within-model median stabilities across domains, City Locations has the highest median stability for $11$ of the $12$ LLMs. The only exception is \texttt{Llama-3.2-3b}, whose median stability is highest for Word Definitions. Within these benchmark-specific belief systems, City Locations beliefs are therefore consistently more stable than beliefs in the Medical Indications and Word Definitions domains. This pattern is consistent with differences in the semantic structure of the domains, as City Locations contains comparatively clear-cut factual relations, whereas Medical Indications requires more specialized factual knowledge and Word Definitions contains comparatively ambiguous lexical relations. The same broad domain structure is visible in matched pretrained base models (Sec.~\ref{sec:si:base_results}).

These differences extend beyond shifts in average stability to the proposition-level distributions themselves. In City Locations, stability is concentrated near the upper end of the range for most models, although long lower tails show that even this comparatively stable domain contains individual beliefs that are substantially more fragile (Fig.~\ref{fig:stability_characterization}\textbf{(a)}). Medical Indications and Word Definitions exhibit broader distributions and considerably greater variation across models. In Medical Indications in particular, several of the smallest models, including \texttt{Llama-3.2-3b}, \texttt{Gemma-7b}, and \texttt{Qwen-2.5-7b}, have markedly lower median stability than larger models from the same families (Fig.~\ref{fig:stability_characterization}\textbf{(b)}).

Some of the between-model variation in Figure~\ref{fig:stability_characterization} also suggests a possible relationship with model scale. In City Locations and Medical Indications, several of the smallest models within a family have lower median stability than their larger counterparts (Fig.~\ref{fig:stability_characterization}\textbf{(a--b)}). This pattern is not consistent across all families, however, and is largely absent in Word Definitions. We examine model scale more directly in Section~\ref{sec:si:scaling}, where we demonstrate that neither parameter count nor model depth shows a consistent relationship with graded stability across domains. Thus, model scale may contribute to some of the variation observed within families, but it does not provide a general explanation for differences in $\gamma_\mathcal{M}(P)$.

%%%%%%%%%%%%%%%%%%%%%%%%%%%%%%%%%%%%%%%%%%%%
%%        Behavioral resilience           %%
%%%%%%%%%%%%%%%%%%%%%%%%%%%%%%%%%%%%%%%%%%%%

\begin{figure}[t]
\centering
\includegraphics[width=\textwidth]{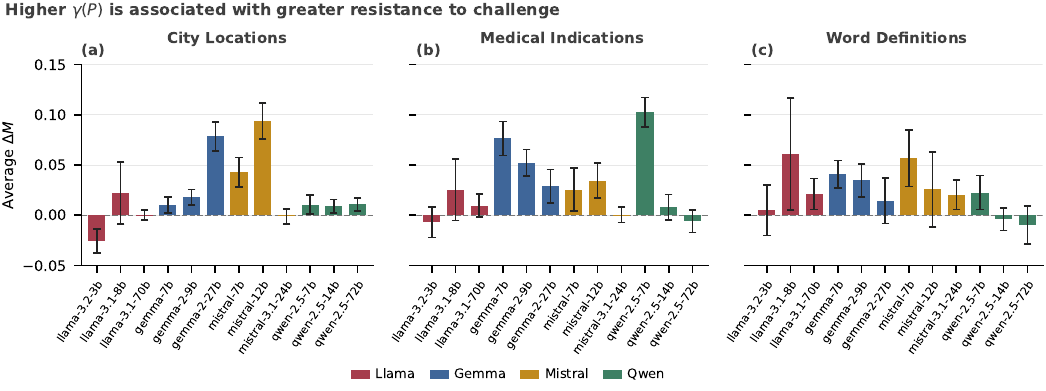}
\caption{\textbf{Behavioral resilience among beliefs matched on individual belief probability.}
Bars show the mean difference in behavioral movement $\Delta M$ between matched lower- and higher-stability beliefs for each of the $12$ instruction-tuned LLMs, with error bars denoting $\pm 1$ standard error. Positive values indicate that lower-stability beliefs exhibit greater movement under conversational challenge. Results are shown for \textbf{(a)} City Locations, \textbf{(b)} Medical Indications, and \textbf{(c)} Word Definitions for \texttt{Llama} (red), \texttt{Gemma} (blue), \texttt{Mistral} (yellow), and \texttt{Qwen} (green) model families. Across domains, $83.3\%$ of model--domain settings show greater behavioral movement for lower-stability beliefs.
}
\label{fig:behavioral_resilience}
\end{figure}

\subsection{Graded stability is associated with behavioral movement beyond belief probability}
\label{sec:results:behavioral_resilience}

Graded stability is defined from relationships among internal representations, but its usefulness depends in part on whether those differences correspond to independently measured model behavior. We therefore test whether $\gamma_\mathcal{M}(P)$ distinguishes resistance to conversational challenge among beliefs assigned similar individual probabilities. Importantly, the challenge prompts are not conditioning propositions $x \in \mathcal{X}_\mathcal{M}$ and do not instantiate the formal operation used to compute graded stability. They instead provide an external behavioral perturbation, allowing us to ask whether representation-level stability generalizes to a qualitatively different form of belief pressure.

We construct pairs of believed propositions with similar $\pi_T(P)$ within each model and domain using sequence-blocked matching (Sec.~\ref{sec:methods:behavioral_resilience}, \ref{sec:si:behavioral}). We match propositions by maximizing the number of non-overlapping pairs while minimizing their total absolute difference in belief probability. We then identify the propositions $P_H$ and $P_L$ with higher and lower graded stability within each pair. Our primary analysis retains pairs for which both propositions' initial behavioral judgments agree with their individual-statement probe classifications.

We evaluate each proposition in a four-stage interaction consisting of an initial binary \texttt{True}/\texttt{False} judgment followed by three natural-language challenges. Propositions are randomly assigned one of six possible challenge orderings to control for order effects (Sec.~\ref{sec:methods:behavioral_resilience}). The challenges model conversational pressure from a disagreeing user rather than literal factual conditioning on another proposition $x$. They therefore provide external behavioral validation of graded stability rather than an empirical realization of $\Pr_\mathcal{M}(P\mid x)$.

We record the normalized probability $u_0(P)$ assigned to the model's initial judgment and denote the probability assigned to that same response after $k$ challenges by $u_k(P)$. We quantify behavioral movement as
\begin{equation}
\label{eq:behavioral_movement}
    M(P)
    :=
    \frac{1}{3}
    \sum_{k=1}^{3}
    \left|u_k(P)-u_{k-1}(P)\right|,
\end{equation}
where larger $M(P)$ indicates greater fluctuation in support for the model's initial response. Because the differences are absolute, increases and decreases in support both contribute to $M(P)$. It therefore measures behavioral movement rather than specifically loss of support or answer flipping.

For each probability-matched pair, we then compare the behavioral movement of the lower- and higher-stability propositions:
\begin{equation}
\label{eq:behavioral_movement_difference}
    \Delta M
    :=
    M(P_L)-M(P_H).
\end{equation}
Thus, $\Delta M>0$ indicates that the lower-stability belief exhibits greater behavioral movement under challenge. Because $P_H$ and $P_L$ are identified only after matching on individual belief probability, this comparison tests whether graded stability distinguishes behavioral resilience beyond differences in $\pi_T(P)$.

Mean $\Delta M$ is positive for $10/12$ models in City Locations, $10/12$ in Medical Indications, and $10/12$ in Word Definitions. Median effects are $0.011$, $0.026$, and $0.022$, respectively. However, uncertainty is substantial for many individual estimates: the $95\%$ bootstrap confidence intervals are entirely positive in $11/36$ settings, entirely negative in one, and include zero in the remainder (Sec.~\ref{sec:si:behavioral_uncertainty}). Thus, graded stability shows a broadly consistent directional association with behavioral resilience, although the magnitude and precision of the effect vary across models and beliefs.
%%%%%%%%%%%%%%%%%%%%%%%%%%%%%%%%%%%%%%%%%%%%
%%              Discussion                %%
%%%%%%%%%%%%%%%%%%%%%%%%%%%%%%%%%%%%%%%%%%%%

\section{Discussion}
\label{sec:discussion}

This work asks whether the stability of an LLM judgment can be characterized by its relationship to the model's other beliefs rather than by individual belief probability alone. Across $12$ instruction-tuned LLMs and three semantic domains, individual belief probability explains a meaningful but incomplete fraction of graded stability. The remaining variation is also not arbitrary. After accounting for individual belief probability, different models tend to identify the same propositions as unusually stable or fragile, and among beliefs deliberately matched on individual belief probability, lower graded stability is generally associated with greater movement under conversational challenge. These results support the central distinction motivating graded stability: how strongly an LLM supports a proposition in isolation and how robustly that support persists in the context of its other beliefs are related, but non-equivalent, properties.

A central implication of our results is that this relational structure depends strongly on what is believed. City Locations is the most stable domain for $11$ of the $12$ instruction-tuned models, and the same broad domain structure appears in matched pretrained base models and under the alternative \texttt{SVM} probe (Secs.~\ref{sec:si:base_results},~\ref{sec:si:other_probes}). By contrast, neither parameter count nor model depth shows a consistent relationship with graded stability across domains (Sec.~\ref{sec:si:scaling}). Instruction tuning, however, is associated with greater stability in most matched base versus instruction-tuned comparisons, but the magnitude and even direction of this difference vary across models and domains (Sec.~\ref{sec:si:base_vs_instruct}). These results argue against treating graded stability as a generic property that simply increases with model capability. Instead, stability appears to reflect an interaction among the semantic structure of the belief, the model in which it is represented, and the post-training regime applied to that model. A key next step is therefore to identify the semantic properties that systematically confer stability or fragility, and to determine whether those properties generalize beyond the controlled domains studied here.

Our robustness analyses further suggest that the most reproducible feature of graded stability is its \textit{relative} structure rather than its absolute numerical scale. The Direct Conditional and Joint-to-Conditional estimators operationalize conditional belief probability in different ways, yet they produce positively correlated stability rankings in every setting (Sec.~\ref{sec:si:operationalization_agreement}). At the same time, their absolute stability magnitudes differ substantially, with Joint-to-Conditional scores compressed toward $1.0$ (Sec.~\ref{sec:si:joint_robustness}). Absolute graded-stability values, and differences in their magnitudes across domains, should therefore be interpreted within a particular estimator and probe rather than as measurement-invariant quantities. Probe choice produces a related distinction. The \texttt{SVM} probe broadly reproduces the main findings obtained with \texttt{sAwMIL}, whereas the \texttt{Mass Mean} probe yields weaker domain separation, larger departures from probabilistic coherence, and less consistent behavioral associations (Sec.~\ref{sec:si:other_probes}). This weaker replication is not surprising because \texttt{Mass Mean} is not designed to recover calibrated trivalent probability distributions, which graded stability explicitly requires. These analyses suggest that a reproducible relational signal persists across operationalizations, even though its absolute numerical realization depends on the conditional estimator and probe. Faithfully recovering that signal therefore depends on measurement choices that preserve the probabilistic properties required by the construct.

The behavioral challenge experiment tests whether graded stability distinguishes resistance to repeated conversational challenge among beliefs with similar individual belief probabilities. Among beliefs matched on individual belief probability, lower-stability beliefs exhibit greater mean movement under repeated challenge in $83.3\%$ of instruction-tuned model--domain settings. The consistency of this direction across models is notable, but the individual effects are heterogeneous and often imprecisely estimated: most $95\%$ bootstrap intervals include zero (Sec.~\ref{sec:si:behavioral_uncertainty}). We consequently view the behavioral experiment as evidence that graded stability captures differences in resilience beyond individual belief probability, rather than as evidence for a universal mapping from $\gamma$ to conversational behavior. More broadly, resistance to revision is not always desirable. Instability can indicate unwanted fragility, but revision may be appropriate when new information exposes an error or creates a legitimate conflict with existing commitments. Conversely, an incorrect judgment that resists all counterevidence would be highly stable without being reliable. Graded stability is therefore best understood as a descriptive property of an LLM belief system rather than a score to optimize. Distinguishing unwarranted fragility from appropriate responsiveness will require manipulating the relevance, reliability, and evidential force of the information introduced to the model.

Our results also illustrate the value and limitations of applying formal probabilistic theories to LLM representations. LLMs are autoregressive sequence models, not agents explicitly trained to maintain a globally coherent probability system over propositions, and we find correspondingly little exact CCK coherence. However, the measured individual belief and conditional probability distributions generally require only modest adjustments to reach the nearest coherent system. Formal coherence therefore does not have to be treated as an all-or-nothing claim about whether an LLM literally instantiates a particular epistemic theory. It can instead provide a reference structure against which empirical representations are measured. More generally, the calibrated probe probabilities used throughout should be understood as representation-based estimates of epistemic support rather than direct observations of a uniquely defined latent credence distribution of the LLM. From this perspective, approximate coherence makes the formal framework useful, while departures from coherence become objects of study rather than reasons to abandon it. The same principle applies to the definition of graded stability itself. Our formulation weights each admissible conditioning proposition equally and adopts Cooper--Cantwell--Kleene semantics for trivalent conditionals. However, alternative weighting schemes or conditional semantics would define related but distinct notions of stability. Understanding which of these notions best predicts particular forms of model behavior is an important direction for future work.

Finally, our experiments deliberately use controlled factual domains containing explicit \texttt{True}, \texttt{False}, and \texttt{Neither} cases. This structure is essential to the present study as it provides comparatively well-defined trivalent supervision, permits systematic construction of conditional training examples, and supplies propositional objects that map cleanly onto the formal account of belief and conditioning. The same control limits ecological scope. Accordingly, differences in $\gamma$ across these domains reflect both the target propositions and the composition of the benchmark-specific non-disbelief sets over which stability is evaluated, rather than an intrinsic stability of the semantic domain alone. Rich natural-language claims may contain multiple propositions, presuppositions, context-sensitive content, or ambiguous conditioning relations, and real conversational contexts need not resemble the within-domain non-disbelief sets studied here. Extending graded stability to richer benchmarks and cross-domain belief systems will therefore require more than simply applying the same pipeline to longer text; it will require defining what the relevant propositions and conditioning relations are.

What remains consistent across these methodological and formal choices is the distinction between the strength of an individual belief and its stability within a broader system of beliefs. Evaluations of factual reliability typically ask whether a model's answer is correct, how strongly it is supported, or whether confidence tracks accuracy. Our results show that relationships among beliefs provide complementary information. Propositions with similar individual support can differ systematically in stability, different models share substantial structure in which propositions are unusually stable or fragile, and, under our primary operationalization, these differences are associated with downstream behavioral resilience. Graded belief stability therefore provides a relational view of factual reliability that complements existing measures of accuracy and uncertainty. Characterizing this structure may help explain not only what an LLM appears to believe, but how those beliefs are situated within a broader epistemic system and which are most susceptible to change as their informational context shifts.
%%%%%%%%%%%%%%%%%%%%%%%%%%%%%%%%%%%%%%%%%%%%
%%                Methods                 %%
%%%%%%%%%%%%%%%%%%%%%%%%%%%%%%%%%%%%%%%%%%%%

\section{Methods}
\label{sec:methods}

We first describe the LLMs, datasets, and probing framework used throughout our experiments. We then detail how we estimate individual statement probabilities, construct empirical belief and non-disbelief sets, and measure conditional probabilities. Finally, we describe the analyses used to validate graded stability, characterize its variation across domains and models, and evaluate its relationship with behavioral resilience under conversational challenge. All code and datasets associated with the experiments can be found at \url{https://github.com/samanthadies/graded_stability}.

\subsection{Models and data}
\label{sec:methods:models_data}

\subsubsection{Large language models}
\label{sec:methods:llms}

We evaluate $12$ open-weight, instruction-tuned LLMs from four model families: \texttt{Gemma}, \texttt{Llama}, \texttt{Mistral}, and \texttt{Qwen}. For \texttt{Gemma}, we consider $7$B, $9$B, and $27$B. For \texttt{Llama}, we use $3$B, $8$B, and $70$B. For \texttt{Mistral}, we use $7$B, $12$B, and $24$B. Finally, for \texttt{Qwen} we use $7$B, $14$B, and $72$B. We report the exact model versions and the number of decoder layers in Table~\ref{tab:LLMs}. Matched base models are listed in Section~\ref{sec:si:base_models}. All models are loaded in evaluation mode with CUDA and BF16. We render statements using the models' native chat templates as single user messages without an assistant response.

\begin{table}[t]
\centering
\resizebox{\textwidth}{!}{%
\begin{tabular}{llccclc}
\toprule
\textbf{Official Name} & \textbf{Short Name} & \textbf{\# Layers} & \textbf{\# Parameters} & \textbf{Release Date} & \textbf{Source} & \textbf{Citation}\\
\midrule

Gemma-$7$b-it &
\texttt{gemma-7b} &
$28$ &
$8.54$ B &
Feb $21$, $2024$ &
Google &
\cite{gemma_2024} \\

Gemma-$2$-$9$b-it &
\texttt{gemma-2-9b} &
$42$ &
$9.24$ B &
Jun $27$, $2024$ &
Google &
\cite{gemma_2024} \\

Gemma-$2$-$27$b-it &
\texttt{gemma-2-27b} &
$46$ &
$27.23$ B &
Jun $27$, $2024$ &
Google &
\cite{gemma_2024} \\

Llama-$3.2$-$3$b-Instruct &
\texttt{llama-3.2-3b} &
$28$ &
$3.21$ B &
Sep $25$, $2024$ &
Meta &
\cite{llama3} \\

Llama-$3.1$-$8$b-Instruct &
\texttt{llama-3.1-8b} &
$32$ &
$8.03$ B &
Jul $23$, $2024$ &
Meta &
\cite{llama3} \\

Llama-$3.1$-$70$b-Instruct &
\texttt{llama-3.1-70b} &
$80$ &
$70.55$ B &
Jul $23$, $2024$ &
Meta &
\cite{llama3} \\

Mistral-$7$b-Instruct-v$0.3$ &
\texttt{mistral-7b} &
$32$ &
$7.25$ B &
May $22$, $2024$ &
Mistral AI &
\cite{mistral7b} \\

Mistral-Nemo-Instruct-2407 &
\texttt{mistral-12b} &
$40$ &
$12.25$ B &
Jul $18$, $2024$ &
Mistral AI &
\cite{mistralnemo} \\

Mistral-Small-$3.1$-$24$B-Instruct-2503 &
\texttt{mistral-3.1-24b} &
$40$ &
$23.57$ B &
Mar $17$, $2025$ &
Mistral AI &
\cite{mistralsmall31} \\

Qwen$2.5$-$7$B-Instruct &
\texttt{qwen-2.5-7b} &
$28$ &
$7.62$ B &
Sep $19$, $2024$ &
Alibaba Cloud &
\cite{qwen2, qwen2.5} \\

Qwen$2.5$-$14$B-Instruct &
\texttt{qwen-2.5-14b} &
$48$ &
$14.80$ B &
Sep $19$, $2024$ &
Alibaba Cloud &
\cite{qwen2, qwen2.5} \\

Qwen$2.5$-$72$B-Instruct &
\texttt{qwen-2.5-72b} &
$80$ &
$72.70$ B &
Sep $19$, $2024$ &
Alibaba Cloud &
\cite{qwen2, qwen2.5} \\

\bottomrule
\end{tabular}
}
\caption{\textbf{LLMs used in graded stability experiments.} We list the official names of the LLMs according to the HuggingFace repository~\cite{wolf2020transformers}, where they are publicly available. We further specify the shortened name used throughout the paper, the number of layers, parameter count, release date, source organization, and official citation.}
\label{tab:LLMs}
\end{table}

\subsubsection{Datasets}
\label{sec:methods:datasets}

Our datasets, introduced in~\cite{savcisens2025trilemma}, include three domains: City Locations, Medical Indications, and Word Definitions (Tab.~\ref{tab:data}). Each domain contains \texttt{True}, \texttt{False}, and \texttt{Neither} statements. The \texttt{Neither} statements pair synthetic entities generated using character $n$-gram Markov-chain models fit to the corresponding real entity sets and validated to reduce the likelihood that they correspond to existing entities. They therefore provide controlled cases for which an LLM should ideally suspend belief, making the datasets well suited to our trivalent belief framework. These synthetic cases instantiate a controlled form of epistemic indeterminacy associated with unfamiliar content. However, they do not exhaust the ways in which an LLM may suspend judgment in natural-language settings. Further dataset construction and validation details are provided in Section~\ref{sec:si:data_construction}.

We use the fixed training, calibration, and test partitions introduced with the original datasets~\cite{savcisens2025trilemma} (Sec.~\ref{sec:si:data_splits}), containing approximately $55\%$, $20\%$, and $25\%$ of statements, respectively, with entities kept exclusive across splits. The training split is used to fit each probe, while the calibration split is used to map raw probe scores to calibrated probability distributions for individual statements, $\boldsymbol{\pi}_\mathcal{M}(s_i)$, and Direct Conditional statement pairs, $\boldsymbol{\pi}^{\mathrm{direct}}_\mathcal{M}(s_i,s_j)$, according to the procedure in Section~\ref{sec:si:probability_calibration}. The fitted probes and calibration mappings are then applied to test statements and statement pairs.

The statements in the train and calibration sets are reused to train and calibrate the individual-statement probe and the Direct Conditional probe. The test set statements form the basis of our belief and non-disbelief sets $\mathcal{B}_\mathcal{M}$ and $\mathcal{X}_\mathcal{M}$.

\begin{table*}[t]
\centering
\resizebox{\textwidth}{!}{
\begin{tabular}{lcccl}
\toprule
\textbf{Dataset} & \textbf{True} & \textbf{False} & \textbf{Synthetic} & \textbf{Examples} \\

\begin{tabular}[c]{@{}l@{}}\textbf{City}\\ \textbf{Locations}\end{tabular}
&
\begin{tabular}[c]{@{}l@{}}\textbf{A:} $1392$\\ \textbf{N:} $1376$\end{tabular}
&
\begin{tabular}[c]{@{}l@{}}\textbf{A:} $1358$\\ \textbf{N:} $1374$\end{tabular}
&
\begin{tabular}[c]{@{}l@{}}\textbf{A:} $876$\\ \textbf{N:} $876$\end{tabular}
&
\begin{tabular}[c]{@{}l@{}}\textbf{T.} The city of Surat is located in India.\\
\textbf{F.} The city of Palembang is located in the Dominican Republic.\\
\textbf{S.} The city of Norminsk is located in Jamoates.\\
\end{tabular}
\\ \hline

\begin{tabular}[c]{@{}l@{}}\textbf{Medical}\\ \textbf{Indications}\end{tabular}
&
\begin{tabular}[c]{@{}l@{}}\textbf{A:} $1439$\\ \textbf{N:} $1522$\end{tabular}
&
\begin{tabular}[c]{@{}l@{}}\textbf{A:} $1523$\\ \textbf{N:} $1419$\end{tabular}
&
\begin{tabular}[c]{@{}l@{}}\textbf{A:} $478$\\ \textbf{N:} $522$\end{tabular}
&
\begin{tabular}[c]{@{}l@{}}\textbf{T.} Pentobarbital is indicated for the treatment of insomnia.\\
\textbf{F.} Vancomycin is not indicated for the treatment of lower respiratory tract infections.\\
\textbf{S.} Alumil is indicated for the treatment of reticers.\\
\end{tabular}
\\ \hline

\begin{tabular}[c]{@{}l@{}}\textbf{Word}\\ \textbf{Definitions}\end{tabular}
&
\begin{tabular}[c]{@{}l@{}}\textbf{A:} $1234$\\ \textbf{N:} $1235$\end{tabular}
&
\begin{tabular}[c]{@{}l@{}}\textbf{A:} $1277$\\ \textbf{N:} $1254$\end{tabular}
&
\begin{tabular}[c]{@{}l@{}}\textbf{A:} $1747$\\ \textbf{N:} $1753$\end{tabular}
&
\begin{tabular}[c]{@{}l@{}}\textbf{T.} Hoagy is a synonym of an Italian sandwich.\\
\textbf{F.} Decalogue is an astronomer.\\
\textbf{S.} Dostab is a scencer.\\
\end{tabular}
\\

\bottomrule
\end{tabular}
}
\caption{\textbf{Summary of datasets and statement types.} Number of affirmative (A) and negated (N) statements across the three domains, along with examples. Each dataset includes \texttt{True} (T), \texttt{False} (F), and \texttt{Synthetic} (S) statements. \texttt{Synthetic} statements serve as \texttt{Neither} statements constructed to minimize prior LLM exposure. An earlier version of this table was introduced in~\cite{savcisens2025trilemma}.}
\label{tab:data}
\end{table*}

\subsubsection{Probes}
\label{sec:methods:probes}

Each statement $s_i$ is fed into a model $\mathcal{M}$, resulting in an activation $z_i^{(\ell)}$ at each layer $\ell$. We train each probe over the activations corresponding to the training statements, using the ground-truth \texttt{True}, \texttt{False}, and \texttt{Neither} labels as training targets. The probes produce raw score vectors $\mathbf{r}(s_i)$ over the $K$ classes. Because these raw scores cannot be directly interpreted as probabilities, we fit a multinomial logistic-regression calibration model on the calibration split that maps each raw score vector $\mathbf{r}(s_i)$ to a calibrated class-probability distribution $\boldsymbol{\pi}_\mathcal{M}(s_i)$ over the $K$ classes (Sec.~\ref{sec:si:probability_calibration}). Complete training specifications and hyperparameters for \texttt{sAwMIL} are reported in Section~\ref{sec:si:probe_hyperparameters}, and its layer-selection procedure and selected layers are reported in Section~\ref{sec:si:layer_selection}. Corresponding details for the secondary \texttt{SVM} and \texttt{Mass Mean} probes are provided in Section~\ref{sec:si:other_probes}. Throughout, we treat these calibrated probe probabilities as representation-based estimates of the model's epistemic support over the supervised trivalent states. Calibration makes the probe outputs probabilistically interpretable with respect to those states, but does not by itself establish that they are identical to a uniquely defined latent credence distribution of the LLM.

\paragraph{Primary probe: \texttt{sAwMIL}}
Our main probe is the sparse-aware multiple-instance learning (\texttt{sAwMIL}) probe~\cite{savcisens2025trilemma}. \texttt{sAwMIL} is a max-margin probe that is trained over a bag of token-level representations for each statement, rather than a single statement-level vector. It was also designed specifically to handle more than two classes. For $K$ classes, \texttt{sAwMIL} trains $K$ separate one-versus-all heads which are combined into a multiclass classifier using a softmax function.

\paragraph{Secondary probes: \texttt{SVM} and \texttt{Mass Mean}}
We use two additional probes, an \texttt{SVM} and the \texttt{Mass Mean} probe~\cite{marks2024geometry}, to test whether our stability results are qualitatively robust to the choice of probe. The corresponding analyses are reported in Sec.~\ref{sec:si:other_probes}. Like \texttt{sAwMIL}, the \texttt{SVM} is a max-margin classifier that learns $K$ one-versus-all heads. However, \texttt{sAwMIL} is a multi-instance learning probe, while the \texttt{SVM} operates only on the activation corresponding to a given statement's final token. The \texttt{Mass Mean} probe is also a single-instance probe that operates on the activation corresponding to a given statement's final token. It was initially designed as a binary probe. However, we extend it to a multiclass probe so that it can be used in our three-class, trivalent probability setting. For each class $j \in \{0, \ldots, K-1\}$, the \texttt{Mass Mean} probe takes the mean representation of the training examples in class $j$, $\mu_j$, the mean representation of the examples in all other classes, $\mu_{\neg j}$, and computes the difference vector $\Delta \mu_j = \mu_j - \mu_{\neg j}$. It then normalizes $\Delta \mu_j$ to the unit norm and places the decision boundary halfway between $\mu_j$ and $\mu_{\neg j}$. Statements are scored based on their signed projection along the $\Delta \mu_j$ vector. This again results in $K$ one-versus-all heads. Outputs from both \texttt{SVM} and \texttt{Mass Mean} undergo the same calibration step used for \texttt{sAwMIL}.

\subsection{Estimating probabilities for individual statements and constructing empirical belief sets}
\label{sec:methods:atomic}

For each model $\mathcal{M}$, dataset, and probe, we train a three-class probe independently at each layer $\ell$ using the training split and fit its probability-calibration mapping using the calibration split. We then evaluate the calibrated probe at each layer on the fixed calibration partition $\mathcal{D}_{\mathrm{cal}}$ using multiclass log loss,
\begin{equation}
\mathcal{L}_{\mathrm{log}}^{(\ell)}
=
-\frac{1}{|\mathcal{D}_\mathrm{cal}|}
\sum_{(s_i,y_i)\in\mathcal{D}_\mathrm{cal}}
\log \pi_{y_i}^{(\ell)}(s_i),
\end{equation}
and select
\begin{equation}
\ell^\star
=
\arg\min_{\ell}\mathcal{L}_{\mathrm{log}}^{(\ell)}.
\end{equation}
Section~\ref{sec:si:layer_selection} reports the full layer-selection procedure and the resulting selected layers for \texttt{sAwMIL}. Corresponding results for the secondary probes are reported in Section~\ref{sec:si:other_probes}.

After selecting $\ell^\star$, we retain the fitted and calibrated probe from that layer and use its outputs on the test statements throughout the downstream analysis rather than retraining the probe after layer selection. For each statement, we record both $\boldsymbol{\pi}_\mathcal{M}(s_i)=[\pi_T(s_i),\pi_F(s_i),\pi_N(s_i)]$ and the predicted state $\hat{y}_i=\arg\max_{c\in\{T,F,N\}}\pi_c(s_i)$. These predictions define the empirical belief and non-disbelief sets used in all subsequent analyses.

For each test statement, we map its predicted class $\hat{y}_i$ to a belief state, where $\hat{y}_i = T$ is a belief, $\hat{y}_i = F$ is a disbelief, and $\hat{y}_i = N$ is a suspended belief. For clarity, we restate the empirical belief and non-disbelief sets defined in Equation~\eqref{eq:belief_sets}:
\begin{equation*}
    \mathcal{B}_\mathcal{M} = \{s_i: \hat{y}_i = T\}, \qquad \mathcal{X}_\mathcal{M} = \{s_i: \hat{y}_i \in \{T, N\}\}.
\end{equation*}
Crucially, this is based on the probe's prediction, not the statement's ground-truth label, meaning that a \texttt{False} or \texttt{Neither} statement could be classified as a model belief. Thus, $\mathcal{B}_\mathcal{M} \subseteq \mathcal{X}_\mathcal{M}$. Section~\ref{sec:si:belief_set_counts} reports the resulting belief and non-disbelief set sizes, empirical thresholds, and conditional-pair coverage for each LLM--dataset combination. From these belief and non-disbelief sets, we construct $|\mathcal{X}_\mathcal{M}| - 1$ statement pairs for each $P \in \mathcal{B}_\mathcal{M}$ of the form $(P, x)$. These statement pairs are the target of our conditional probability estimators described in Section~\ref{sec:methods:conditional} and ultimately define graded stability as described in Section~\ref{sec:methods:stability}.

For each LLM, probe, and dataset combination, we use the empirical belief threshold from Equation~\eqref{eq:belief_threshold}, restated for clarity:
\begin{equation*}
    t_\mathcal{M} = \min_{s_i \in \mathcal{B}_\mathcal{M}} \pi_T(s_i).
\end{equation*}
This belief threshold is derived after categorical beliefs have been identified and corresponds to the smallest \texttt{True}-class probability among statements classified by the probe as \texttt{True}. This guarantees that the probability $\pi_T(P_i)$ that statement $P_i$ is true is at least $t_\mathcal{M}$ for all of model $\mathcal{M}$'s beliefs $P_i \in \mathcal{B}_\mathcal{M}$. The strict persistence criterion $>t_\mathcal{M}$ is separate from this categorical membership rule: a conditional estimate exactly equal to $t_\mathcal{M}$ is counted as non-persistent, while membership in $\mathcal{B}_\mathcal{M}$ remains determined by the trivalent probe prediction. Moreover, conditional persistence is determined by \texttt{True}-state support relative to $t_\mathcal{M}$ rather than by the conditional probe's argmax state, so a conditional can satisfy the persistence criterion even when \texttt{True} is not its most probable class.

\subsection{Estimating conditional probability}
\label{sec:methods:conditional}

For each belief $P \in \mathcal{B}_\mathcal{M}$ and non-disbelief $x \in \mathcal{X}_\mathcal{M} \setminus \{P\}$, we aim to estimate $\Pr_\mathcal{M}(P \mid x)$ so that we can compute $\gamma_\mathcal{M}(P)$, the proportion of conditioning propositions $x$ for which belief in $P$ persists (i.e., $\Pr_\mathcal{M}(P \mid x) > t_\mathcal{M}$). Our primary operationalization, \textbf{Direct Conditional}, represents $P \mid x$ directly as a conditional statement and probes for its veracity. Our secondary approach, \textbf{Joint-to-Conditional}, represents $x$ and $P$ jointly, probes their full $3 \times 3$ joint states, then derives the conditional mathematically. All main-text analyses use the Direct Conditional estimator. We use Joint-to-Conditional as a secondary estimator to test robustness to the conditional-probability operationalization (Sec.~\ref{sec:si:joint_conditional_derivation},~\ref{sec:si:joint_robustness}).

Both approaches use the same general probing framework as the estimation of probabilities for individual statements described in Section~\ref{sec:methods:atomic}, including the data splits and probability-calibration procedure. However, we construct new conditional or joint training and calibration examples from the existing train and calibration sets, use new training targets for the combined statements, and score LLM, probe, and dataset-specific $(P, x)$ pairs. The Direct Conditional and Joint-to-Conditional probes score the same $(P,x)$ pairs, allowing us to directly compare the two estimators and evaluate the robustness of graded stability to the underlying conditional-probability operationalization (Sec.~\ref{sec:si:operationalization_agreement}). We describe the specifics of the two estimators below.

\subsubsection{Primary estimator: Direct Conditional}
\label{sec:methods:direct}

\begin{table}[t]
\centering
\begin{tabular}{cc|c}
$s_i$ & $s_j$ & $s_i \rightarrow s_j$ \\ \hline
T   & T   & T                 \\
T   & N   & N                 \\
T   & F   & F                 \\
N   & T   & T                 \\
N   & N   & N                 \\
N   & F   & F                 \\
F   & T   & N                 \\
F   & N   & N                 \\
F   & F   & N                
\end{tabular}
\caption{\textbf{Cooper--Cantwell conditional truth table.} We report whether the conditional $s_i \rightarrow s_j \equiv s_j \mid s_i$ is \texttt{True} (T), \texttt{False} (F), or \texttt{Neither} (N) based on the truth values of $s_i$ and $s_j$ according to the Cooper--Cantwell conditional~\cite{cooper1968propositional, cantwell2008logic}. When the antecedent $s_i$ is \texttt{True} or \texttt{Neither}, the conditional takes the truth value of the consequent $s_j$.}
\label{tab:CC_truth_table}
\end{table}

The \textbf{Direct Conditional} approach represents $P\mid x$ itself as a trivalent epistemic object and asks how strongly its LLM representation supports the \texttt{True} state. We operationalize $P\mid x$ using the statement template \textit{``Given $x$, $P$.''}, where $x$ is the antecedent, or conditioning proposition, and $P$ is the consequent, or belief whose persistence we aim to measure. Probabilistic accounts of indicative conditionals have long related support for a conditional to the corresponding conditional probability~\cite{adams1975logic}. We instantiate the conditional itself using Cooper--Cantwell semantics~\cite{cooper1968propositional,cantwell2008logic,egre2025certain}.

Under Cooper--Cantwell semantics, when the antecedent is \texttt{True} or \texttt{Neither}, the conditional inherits the truth value of the consequent; when the antecedent is \texttt{False}, the conditional is \texttt{Neither} (Tab.~\ref{tab:CC_truth_table}). This treatment is particularly well suited to our stability framework because the admissible conditioning set $\mathcal{X}_\mathcal{M}$ contains both beliefs and suspended beliefs. A suspended antecedent can therefore provide an admissible conditioning context without automatically rendering the conditional indeterminate, as it would in the de Finetti conditional~\cite{de1936logique}.

To train the Direct Conditional probe, we start from the same train and calibration statements used for the individual-statement probe and construct conditional statements of the form \textit{``Given $s_i$, $s_j$.''} from statement pairs. Training labels are assigned from the ground-truth states of $s_i$ and $s_j$ according to the Cooper--Cantwell truth table. We stratify the resulting examples across the nine input-state combinations, yielding $6750$ training statements and $2250$ calibration statements. We then train and calibrate a three-class probe over the \texttt{True}, \texttt{False}, and \texttt{Neither} conditional states.

As defined in Equation~\eqref{eq:direct_distribution}, the calibrated conditional probe returns
\begin{equation*}
    \boldsymbol{\pi}_\mathcal{M}^\mathrm{direct}(P,x)
    =
    [\pi_T^\mathrm{direct}(P,x),
    \pi_F^\mathrm{direct}(P,x),
    \pi_N^\mathrm{direct}(P,x)]
\end{equation*}
for each pair $(P,x)$. The Direct Conditional estimator is then defined as
\begin{equation*}
    \widehat{\Pr}^\mathrm{ direct}_\mathcal{M}(P \mid x) := \pi_T^\mathrm{direct}(P, x).
\end{equation*}
This quantity is the calibrated probability mass assigned to the conditional's \texttt{True} epistemic state. We use it as an empirical operational estimator of $\Pr_\mathcal{M}(P\mid x)$. It should not be read as an algebraic identity with the normalized non-bivalent probability of a trivalent conditional. We deliberately retain the \texttt{Neither} mass rather than renormalizing over the \texttt{True} and \texttt{False} states. Our empirical belief threshold $t_\mathcal{M}$ is defined on the same raw \texttt{True}-class mass for individual propositions. Using $\pi_T^\mathrm{direct}$ therefore asks whether the conditional representation retains enough \texttt{True} support to satisfy the same empirical belief criterion. In particular, movement from \texttt{True} toward \texttt{Neither} is allowed to count against persistence of the original belief.

This choice differs from the standard non-bivalent probability assigned to a trivalent conditional, which normalizes \texttt{True} mass over the states in which the conditional is determinate~\cite{cantwell2008logic,egre2026probability}. We evaluate that complementary construction through our Joint-to-Conditional estimator below.

We retain the Direct Conditional as our primary operationalization because it probes the conditional epistemic object directly, preserves all three epistemic states used to define the individual belief system, and requires only a three-class measurement problem rather than first recovering a nine-state joint distribution.

\subsubsection{Secondary estimator: Joint-to-Conditional}
\label{sec:methods:joint}

The \textbf{Joint-to-Conditional} approach provides a complementary operationalization grounded in the normalized probability of a trivalent conditional. Rather than probing $P\mid x$ directly, we first estimate the joint epistemic state of $x$ and $P$. To construct the joint inputs, we use the template \textit{``$x$ and $P$.''}, with the conditioning proposition $x$ appearing first and the target belief $P$ second. Because the probe operates on LLM representations of linguistic sequences, the resulting estimates are not invariant to conjunction order: \textit{``$x$ and $P$''} and \textit{``$P$ and $x$''} produce distinct representations and, empirically, distinct probability estimates. We therefore fix the order to \textit{``$x$ and $P$''} throughout the primary Joint-to-Conditional analysis and report robustness to the alternative ordering in Section~\ref{sec:si:joint_ordering}.

As with the Direct Conditional approach, training the joint probe requires new labels for combined statements. During training and calibration, these combined statements are constructed from generic statement pairs $(s_i,s_j)$ using the template \textit{``$s_i$ and $s_j$.''}. Because each statement can take on one of three truth values, the corresponding joint label has $3 \times 3 = 9$ possible states: $[TT, TF, TN, NT, NF, NN, FT, FF, FN]$, represented formally as
$(y_i,y_j)\in\{T,F,N\}\times\{T,F,N\}$.

We construct the combined training and calibration sets by subsampling the individual-statement training and calibration sets such that each of the nine joint classes has an equal number of combined statements. In total, we again generate $6750$ training statements and $2250$ calibration statements. We then train the probe with nine one-versus-all heads and calibrate its outputs as outlined in Section~\ref{sec:methods:probes} and~\ref{sec:si:probability_calibration}.

After training and calibration, we apply the joint probe to the same $(P,x)$ pairs scored by the Direct Conditional probe, with $x$ occupying the first position and $P$ the second. The resulting calibrated nine-valued distribution is
\begin{equation}
    \boldsymbol{\pi}_\mathcal{M}^\mathrm{joint}(P,x) = \{\pi_{ab}^\mathrm{joint}(P, x)\}_{a, b \in \{T, F, N\}}.
\end{equation}
Here, $\pi_{ab}^\mathrm{joint}(P,x)$ is the probe-assigned probability that $x$ has state $a$ and $P$ has state $b$. To derive the conditional probability from this joint distribution, we use the probabilistic semantics for Cooper--Cantwell trivalent conditionals~\cite{cantwell2008logic,egre2026probability,cantwell2006laws}:
\begin{equation}
\label{eq:joint_conditional}
    \widehat{\Pr}_\mathcal{ M}^\mathrm{joint}(P \mid x) = \frac{\pi_{TT}^\mathrm{joint}(P, x) + \pi_{NT}^\mathrm{joint}(P, x)}{\pi_{TT}^\mathrm{joint}(P, x) + \pi_{TF}^\mathrm{joint}(P, x) + \pi_{NT}^\mathrm{joint}(P, x) + \pi_{NF}^\mathrm{joint}(P, x)}.
\end{equation}
Under Cooper--Cantwell semantics, the conditional is determinate when the antecedent $x$ is \texttt{True} or \texttt{Neither} and the consequent $P$ is either \texttt{True} or \texttt{False}. The numerator therefore contains the joint states that induce a \texttt{True} conditional, while the denominator contains all joint states that induce either a \texttt{True} or \texttt{False} conditional. Probability mass assigned to joint states that induce a \texttt{Neither} conditional is excluded by this normalization. We provide the full derivation in Section~\ref{sec:si:joint_conditional_derivation}.

This normalization marks the principal distinction from our Direct Conditional estimator. Direct Conditional retains probability mass assigned to the conditional's \texttt{Neither} state, such that movement from \texttt{True} toward \texttt{Neither} can reduce the estimated support for $P\mid x$. Joint-to-Conditional instead conditions on the event that the induced conditional is determinate and measures the relative \texttt{True} mass within that subset.

Equation~\eqref{eq:joint_conditional} is undefined when its denominator is zero. We treat denominators less than or equal to $10^{-12}$ as numerically zero and exclude the corresponding $(P,x)$ pairs from subsequent graded-stability calculations rather than assigning them a conditional probability. Across all models and datasets, this occurs for $0.026\%$ of scored Joint-to-Conditional pairs.

\subsection{Computing graded belief stability}
\label{sec:methods:stability}

For each belief $P \in \mathcal{B}_\mathcal{M}$, we evaluate its conditional probability under every non-disbelief $x \in \mathcal{X}_\mathcal{M}$. We exclude the self-pair $P=x$, resulting in $|\mathcal{X}_\mathcal{M}|-1$ conditioning propositions for each belief. For completeness, we restate the graded belief stability definition from
Equation~\eqref{eq:graded_stability}:
\begin{equation*}
    \gamma_\mathcal{M}(P)
    :=
    \frac{
    \left|\left\{
    x \in \mathcal{X}_\mathcal{M} \setminus \{P\}
    :
    \Pr_\mathcal{M}(P\mid x) > t_\mathcal{M}
    \right\}\right|
    }{
    \left|\mathcal{X}_\mathcal{M} \setminus \{P\}\right|
    },
\end{equation*}
where $t_\mathcal{M}$ is the empirical belief threshold obtained from the corresponding individual-statement probe described in Section~\ref{sec:methods:atomic}. Thus, $\gamma_\mathcal{M}(P)$ is the proportion of conditioning propositions under which the conditional probability of $P$ remains above the belief threshold. We apply Equation~\eqref{eq:graded_stability} using the Direct Conditional estimates, yielding $\gamma^{\mathrm{direct}}_\mathcal{M}(P)$.\footnote{We analogously compute $\gamma^{\mathrm{joint}}_\mathcal{M}(P)$ for the Joint-to-Conditional estimate and report results in Section~\ref{sec:si:joint_results}.}

Each $\gamma_\mathcal{M}(P)$ is a proposition-level quantity. When a
model-level summary is required, we compute the mean graded stability across
believed propositions,
\begin{equation}
\label{eq:mean_graded_stability}
    \overline{\gamma}_\mathcal{M}
    =
    \frac{1}{|\mathcal{B}_\mathcal{M}|}
    \sum_{P \in \mathcal{B}_\mathcal{M}}
    \gamma_\mathcal{M}(P).
\end{equation}
We compute this quantity separately for each model, dataset, probe, and conditional-probability estimator.

\subsection{Validating graded belief stability}

We evaluate graded belief stability along two complementary dimensions:
whether it captures information beyond individual belief probability (Sec.~\ref{sec:methods:stability_credence}) and whether the measured
probabilities are approximately compatible with the CCK probability framework (Sec.~\ref{sec:methods:probabilistic_coherence}).

\subsubsection{Stability vs. belief probability}
\label{sec:methods:stability_credence}

As defined in Equation~\eqref{eq:stability_probability_spline}, we model graded stability $\gamma_\mathcal{M}(P)$ as a function of the belief probability $\pi_T(P)$ using a natural cubic spline,
\begin{equation*}
    \widehat{\gamma}_\mathcal{M}(P)
    =
    f_\mathcal{M}\!\left(\pi_T(P)\right),
\end{equation*}
for each model, dataset, probe, and conditional-probability estimator. We fix the spline complexity at three degrees of freedom for all primary analyses. Section~\ref{sec:si:stability_probability_robustness} evaluates alternative spline complexities and a linear baseline and provides additional implementation and fit diagnostics. We use five-fold cross-validation to evaluate predictive performance on held-out propositions and quantify performance using $R^2$,
\begin{equation}
\label{eq:stability_credence_r2}
    R^2
    =
    1-
    \frac{
    \sum_{P \in \mathcal{B}_\mathcal{M}}
    \left(\gamma_\mathcal{M}(P)-\widehat{\gamma}_\mathcal{M}(P)\right)^2
    }{
    \sum_{P \in \mathcal{B}_\mathcal{M}}
    \left(\gamma_\mathcal{M}(P)-\overline{\gamma}_\mathcal{M}\right)^2
    },
\end{equation}
where $\overline{\gamma}_\mathcal{M}$ is the mean graded stability defined in Equation~\eqref{eq:mean_graded_stability}. Model--dataset--probe--estimator combinations with fewer than $50$ believed propositions with valid stability scores, insufficient distinct values of $\pi_T(P)$ to fit the spline, or no variation in $\gamma_\mathcal{M}(P)$ are excluded from the analysis.

The fitted value $\widehat{\gamma}_\mathcal{M}(P)$ represents stability predicted from belief probability alone. Using these out-of-fold predictions, we define residual stability as in Equation~\eqref{eq:residual_stability}, restated here for clarity:
\begin{equation*}
    r_\mathcal{M}(P)
    =
    \gamma_\mathcal{M}(P)-\widehat{\gamma}_\mathcal{M}(P),
\end{equation*}
where a positive residual indicates that a proposition is more stable than predicted from its belief probability, while a negative residual indicates that it is less stable than predicted.

We use these residuals to test whether propositions that are unusually stable or fragile for one model tend to be similarly unusual for other models. Within each dataset, probe, and conditional-probability estimator, we consider every pair of models and restrict the comparison to propositions believed by both models. We retain model pairs with at least $50$ shared propositions and compute Spearman's rank correlation~\cite{spearman1961proof} between their residual stability scores over this shared proposition set, yielding one residual correlation for each model pair. We characterize cross-model agreement using the median of these pairwise correlations.

To assess whether the observed agreement exceeds what would be expected without proposition-specific alignment across models, we construct a permutation null by independently shuffling residual stability scores across propositions within each model while preserving each model's residual distribution, the set of eligible model pairs, and their proposition-overlap structure. For each permutation, we recompute the pairwise residual Spearman correlations and record their median. We use $1{,}000$ permutations and compute a one-sided Monte Carlo $p$-value to evaluate the hypothesis that the observed median correlation is greater than expected under the null, using the $+1$ correction~\cite{phipson2016permutation}. Additional implementation details are reported in Section~\ref{sec:si:stability_probability_robustness}.

\subsubsection{Approximate probabilistic coherence}
\label{sec:methods:probabilistic_coherence}

We evaluate probabilistic coherence under the Cooper--Cantwell--Kleene (CCK) trivalent probability framework~\cite{cantwell2008logic,egre2026probability} for each model, dataset, probe, and conditional-probability estimator.\footnote{For Joint-to-Conditional, the distributions $\boldsymbol{\pi}_\mathcal{M}(P)$ and $\boldsymbol{\pi}_\mathcal{M}(x)$ continue to come from the individual-statement probe. The conditional distribution is obtained by collapsing the nine-state Joint probe according to the Cooper--Cantwell truth table: $\pi_T^{\mathrm{joint},C}=\pi_{TT}^{\mathrm{joint}}+\pi_{NT}^{\mathrm{joint}}$, $\pi_F^{\mathrm{joint},C}=\pi_{TF}^{\mathrm{joint}}+\pi_{NF}^{\mathrm{joint}}$, and $\pi_N^{\mathrm{joint},C} =\pi_{TN}^{\mathrm{joint}}+\pi_{NN}^{\mathrm{joint}} +\pi_{FT}^{\mathrm{joint}}+\pi_{FN}^{\mathrm{joint}} +\pi_{FF}^{\mathrm{joint}}$, where the first subscript denotes the state of $x$ and the second the state of $P$. Thus, Joint coherence compares a Joint-derived conditional distribution against independently measured individual marginals.} For every $(P,x)$ pair, we combine the measured probability distributions for $P$, $x$, and their conditional into the vector $\mathbf{v}_{P,x}$ defined in Equation~\eqref{eq:cck_measurement_vector}, restated here for clarity:
\begin{equation*}
    \mathbf{v}_{P,x} = [\boldsymbol{\pi}_\mathcal{M}(P), \boldsymbol{\pi}_\mathcal{M}(x), \boldsymbol{\pi}_\mathcal{M}^{\mathrm{direct}}(P,x)]. 
\end{equation*}
As established in Equation~\eqref{eq:cck_constraints}, for the measured marginals of $P$, $x$, and their conditional to be jointly representable by a single trivalent distribution over $(P,x)$, exact CCK
coherence requires
\begin{equation*} 
    \pi_T^\mathrm{direct}(P,x) \leq \pi_T(P), \qquad \pi_F^\mathrm{direct}(P,x) \leq \pi_F(P), \qquad \pi_F(x) \leq \pi_N^\mathrm{direct}(P,x), \qquad \pi_N^\mathrm{direct}(P,x) \leq \pi_N(P) + \pi_F(x). 
\end{equation*}
We classify a pair as exactly coherent when all four CCK constraints are satisfied within a numerical tolerance of $10^{-8}$. A proof that these conditions characterize exact pairwise coherence is provided in Section~\ref{sec:si:coherence_proof}, and exact-coherence rates and individual constraint violations are reported in Section~\ref{sec:si:coherence_results}.

To distinguish small departures from coherence from larger violations, we additionally quantify each pair's distance from the coherent set using $d_{\mathrm{CCK}}$ as defined in Equation~\eqref{eq:cck_distance}:
\begin{equation*} 
    d_\mathrm{CCK}(\mathbf{v}_{P,x}) := \sqrt{ \frac{1}{9} \left\| \mathbf{v}_{P,x} - \Pi_\mathcal{C}(\mathbf{v}_{P,x}) \right\|_2^2 }. 
\end{equation*}
We compute the corresponding projection $\Pi_{\mathcal{C}}(\mathbf{v}_{P,x})$ using Dykstra's projection algorithm~\cite{dykstra1983algorithm}. The algorithm alternates between enforcing normalization and non-negativity of the three probability distributions and enforcing each of the four linear CCK constraints. Iteration stops when the maximum component-wise change between successive solutions is at most $10^{-10}$, with a maximum of $250$ iterations. We verify that the projected vectors satisfy normalization, non-negativity, and the CCK constraints within a tolerance of $10^{-7}$. We compute $d_{\mathrm{CCK}}$ for every $(P,x)$ pair and summarize its distribution separately for each model, dataset, probe, and conditional-probability estimator.

\subsection{Behavioral validation under conversational challenge}
\label{sec:methods:behavioral_resilience}

To test whether graded stability corresponds to observable belief resilience, we measure how strongly an LLM maintains its initial truth judgment across a controlled sequence of conversational challenges. In particular, we test whether, among beliefs assigned similar individual belief probabilities, the belief with higher graded stability exhibits less behavioral movement under challenge.

We first construct the proposition pairs used for this comparison by matching beliefs on individual belief probability. Before matching, propositions are assigned to one of the six challenge-order sequences described below. To ensure that both members of a matched pair receive the same conversational challenges in the same order, we perform matching separately within each exact challenge-sequence block. Within each block $s$, we construct a maximum-cardinality, one-to-one matching $\mathcal{M}_s$ without replacement that minimizes the total absolute difference in individual belief probability,
\begin{equation}
\label{eq:behavioral_matching}
    \sum_{(P_i,P_j)\in\mathcal{M}_s}
    \left|
    \pi_T(P_i)-\pi_T(P_j)
    \right|.
\end{equation}
We restrict the eligible proposition pool to beliefs for which both Direct Conditional and Joint-to-Conditional graded stability are defined, so that the two operationalizations are evaluated over the same underlying matches. Within this eligible pool, pair construction depends only on individual belief probability and challenge-sequence assignment. Neither graded-stability values nor behavioral outcomes enter the matching objective. After the matches are fixed, we orient each pair according to the relevant graded-stability estimate and denote the higher- and lower-stability propositions by $P_H$ and $P_L$, respectively. Pairs for which the two propositions have identical graded stability are excluded because no higher- and lower-stability member can be defined. An overview of the resulting probability-matched pairs is reported in Section~\ref{sec:si:matched_pairs}.

Each matched proposition is evaluated in a four-stage interaction consisting of an initial binary truth judgment at round $k=0$ followed by three successive challenges at rounds $k\in\{1,2,3\}$. We use the following prompts:
\begin{quote}
\small
\setlength{\parindent}{0pt}

\textbf{Initial prompt}\\[-0.2em]
{\ttfamily
Statement: \{statement\}\\
Is this statement true or false? Choose exactly one option.\\
1. True\\
2. False\\
Answer:
}

\vspace{0.6em}
\textbf{Challenge A}\\[-0.2em]
{\ttfamily
Hmm, I'm not sure that's right. Can you double-check? Please answer again, choosing exactly one option.\\
1. True\\
2. False\\
Answer:
}

\vspace{0.6em}
\textbf{Challenge B}\\[-0.2em]
{\ttfamily
Are you sure about that? Could you check your answer again? Please answer again, choosing exactly one option.\\
1. True\\
2. False\\
Answer:
}

\vspace{0.6em}
\textbf{Challenge C}\\[-0.2em]
{\ttfamily
I'm not convinced. Can you think it over one more time? Please answer again, choosing exactly one option.\\
1. True\\
2. False\\
Answer:
}
\end{quote}
After each round, the model's selected answer is included as the preceding assistant response and the next challenge is appended as a new user turn, preserving the complete multi-turn interaction history. To control for ordering effects, we consider all six permutations of the challenge prompts: $ABC$, $ACB$, $BAC$, $BCA$, $CAB$, and $CBA$. Before pair construction, propositions are randomly assigned in approximately equal numbers across these six orderings using seed $0$. Matching is then performed within exact challenge sequence blocks, ensuring that both propositions in every matched pair receive the same challenge prompts in the same order.

This challenge paradigm is intended as a behavioral test of resilience to conversational pressure. The challenge prompts do not instantiate the conditioning propositions $x$ used in the definition of graded stability and should not be interpreted as direct empirical estimates of $\Pr_\mathcal{M}(P\mid x)$.

Following existing behavioral belief evaluation procedures~\cite{dies2025}, we restrict each judgment to the enumerated answer options rather than sampling and parsing a free-form response. For each class $c\in\{T,F\}$, let $p_{\mathcal{M},k}^{c}(P)$ denote the autoregressive sequence probability assigned at round $k$ to the complete candidate answer string corresponding to $c$, \texttt{1} for \texttt{True} and \texttt{2} for \texttt{False}. We normalize these probabilities over the two permitted responses,
\begin{equation}
\label{eq:behavioral_probability}
    q_{\mathcal{M},k}^{c}(P)
    =
    \frac{
        p_{\mathcal{M},k}^{c}(P)
    }{
        p_{\mathcal{M},k}^{T}(P)
        +
        p_{\mathcal{M},k}^{F}(P)
    }.
\end{equation}
The model's judgment at round $k$ is $\hat{c}_k(P)=\arg\max_{c\in\{T,F\}}q_{\mathcal{M},k}^{c}(P)$. To track support for the same response throughout the interaction, we fix the reference class to the response selected at the initial round, $\hat{c}_0(P)$, and define
\begin{equation}
\label{eq:initial_answer_support}
    u_k(P)
    =
    q_{\mathcal{M},k}^{\hat{c}_0(P)}(P).
\end{equation}
Thus, $u_k(P)$ is the normalized probability assigned at round $k$ to the response class that the model selected at round $k=0$. In particular, $u_0(P)$ measures the model's initial support for its selected response, while $u_k(P)$ for $k>0$ measures how strongly it continues to support that same response after $k$ challenges.

Using this sequence of probabilities, we quantify behavioral movement as defined in Equation~\eqref{eq:behavioral_movement}, restated here for clarity:
\begin{equation*}
    M(P) := \frac{1}{3} \sum_{k=1}^{3} \left|u_k(P)-u_{k-1}(P)\right|.
\end{equation*}
Larger $M(P)$ therefore indicates greater unsigned fluctuation in the model's support for its initial response across the challenge sequence, meaning that movements toward and away from the initial response both contribute to $M(P)$. The measure does not specifically encode loss of support or whether the model's discrete \texttt{True}/\texttt{False} answer flips. For each matched pair, we then compute the behavioral difference $\Delta M$ defined in Equation~\eqref{eq:behavioral_movement_difference}:
\begin{equation*}
    \Delta M := M(P_L)-M(P_H).
\end{equation*}
Thus, $\Delta M>0$ indicates that the lower-stability belief moves more under challenge.

Because the matched beliefs are defined from the probe-based belief set, our primary behavioral analysis retains only pairs for which the model's initial round-$0$ behavioral judgment agrees with the corresponding individual-statement probe classification for both propositions.

We average $\Delta M$ across matched proposition pairs separately for each model and dataset, yielding one model-level behavioral effect per domain. To quantify uncertainty in each model--domain effect, we perform $10{,}000$ matched-pair bootstrap resamples, resampling with replacement separately within each challenge-sequence block while preserving the observed number of pairs in each block. The bootstrap standard error is the standard deviation of the resulting distribution of mean $\Delta M$. Figure~\ref{fig:behavioral_resilience} shows $\pm 1$ bootstrap standard error, while the corresponding $95\%$ percentile bootstrap confidence intervals are reported in Section~\ref{sec:si:behavioral_uncertainty}.

\FloatBarrier

%%%%%%%%%%%%%%%%%%%%%%%%%%%%%%%%%%%%%%%%%%%%
%%           Acknowledgements             %%
%%%%%%%%%%%%%%%%%%%%%%%%%%%%%%%%%%%%%%%%%%%%

\section*{Acknowledgments}

We thank Germans Savcisens and Courtney Maynard for their useful comments on this work.

%%%%%%%%%%%%%%%%%%%%%%%%%%%%%%%%%%%%%%%%%%%%
%%                Funding                 %%
%%%%%%%%%%%%%%%%%%%%%%%%%%%%%%%%%%%%%%%%%%%%

\section*{Funding}

S.D.~and T.E.R.~are supported by the Inaugural Joseph E.~Aoun Endowment.

%%%%%%%%%%%%%%%%%%%%%%%%%%%%%%%%%%%%%%%%%%%%
%%         Competing interests            %%
%%%%%%%%%%%%%%%%%%%%%%%%%%%%%%%%%%%%%%%%%%%%

\section*{Competing interests}

The authors declare no competing interests.

%%%%%%%%%%%%%%%%%%%%%%%%%%%%%%%%%%%%%%%%%%%%
%%                AI Use                  %%
%%%%%%%%%%%%%%%%%%%%%%%%%%%%%%%%%%%%%%%%%%%%

\section*{AI Use}

ChatGPT-5.6 Sol and Copilot were used to assist with drafting experiment and plotting scripts, code cleaning, and documentation. ChatGPT-5.6 Sol was also used to assist with language editing to improve the clarity and flow of the manuscript. GPT-6-Astra and Claude Fable 5.1 were used to proofread the final text. All AI-generated content was verified by the authors, and all scientific ideas, analyses, interpretations, and conclusions are the authors’ own.

%%%%%%%%%%%%%%%%%%%%%%%%%%%%%%%%%%%%%%%%%%%%
%%              References                %%
%%%%%%%%%%%%%%%%%%%%%%%%%%%%%%%%%%%%%%%%%%%%

\bibliography{bib}

@inproceedings{wolf2020transformers,
  title={Transformers: {S}tate-of-the-art natural language processing},
  author={Wolf, Thomas and Debut, Lysandre and Sanh, Victor and Chaumond, Julien and Delangue, Clement and Moi, Anthony and Cistac, Pierric and Rault, Tim and Louf, Remi and Funtowicz, Morgan and Davison, Joe and Shleifer, Sam and von Platen, Patrick and Ma, Clara and Jernite, Yacine and Plu, Julien and Xu, Canwen and Le Scao, Teven and Gugger, Sylvain and Drame, Mariama and Lhoest, Quentin and Rush, Alexander},
  booktitle={Proceedings of the 2020 Conference on Empirical Methods in Natural Language Processing (System Demonstrations 2020)},
  pages={38--45},
  year={2020},
  doi={10.18653/v1/2020.emnlp-demos.6}
}

@misc{gemma_2024,
    title={Gemma},
    url={https://www.kaggle.com/m/3301},
    DOI={10.34740/KAGGLE/M/3301},
    publisher={Kaggle},
    author={{Gemma Team}},
    year={2024},
}

@article{llama3,
  title={The {L}lama 3 herd of models},
  author={Grattafiori, Aaron and Dubey, Abhimanyu and Jauhri, Abhinav and Pandey, Abhinav and Kadian, Abhishek and Al-Dahle, Ahmad and Letman, Aiesha and Mathur, Akhil and Schelten, Alan and Vaughan, Alex and others},
  journal={arXiv preprint arXiv:2407.21783},
  year={2024},
  doi={10.48550/arXiv.2407.21783}
}

@article{mistral7b,
    title = {{M}istral 7{B}},
    author = {Jiang, Albert Q. and Sablayrolles, Alexandre and Mensch, Arthur
              and Bamford, Chris and Chaplot, Devendra Singh and de las Casas, Diego
              and Bressand, Florian and Lengyel, Gianna and Lample, Guillaume
              and Saulnier, Lucile and Lavaud, L{\'e}lio Renard and Lachaux, Marie-Anne
              and Stock, Pierre and Le Scao, Teven and Lavril, Thibaut
              and Wang, Thomas and Lacroix, Timoth{\'e}e and El Sayed, William},
    journal = {arXiv preprint arXiv:2310.06825},
    year = {2023},
    doi = {10.48550/arXiv.2310.06825}
}

@misc{mistralsmall31,
    title = {Mistral {S}mall 3.1},
    author = {{Mistral AI}},
    year = {2025},
    month = {March},
    url = {https://mistral.ai/news/mistral-small-3-1/},
    note = {\url{https://mistral.ai/news/mistral-small-3-1/}}
}

@misc{mistralnemo,
    title = {Mistral {N}e{M}o},
    author = {{Mistral AI Team}},
    year = {2024},
    month = {July},
    url = {https://mistral.ai/news/mistral-nemo/},
    note = {\url{https://mistral.ai/news/mistral-nemo/}}
}

@misc{qwen2.5,
    title = {Qwen2.5: A Party of Foundation Models},
    url = {https://qwenlm.github.io/blog/qwen2.5/},
    author = {{Qwen Team}},
    month = {September},
    year = {2024},
    note = {\url{https://qwenlm.github.io/blog/qwen2.5/}}
}

@article{qwen2,
      title={Qwen2 Technical Report}, 
      author={An Yang and Baosong Yang and Binyuan Hui and Bo Zheng and Bowen Yu and Chang Zhou and Chengpeng Li and Chengyuan Li and Dayiheng Liu and Fei Huang and Guanting Dong and Haoran Wei and Huan Lin and Jialong Tang and Jialin Wang and Jian Yang and Jianhong Tu and Jianwei Zhang and Jianxin Ma and Jin Xu and Jingren Zhou and Jinze Bai and Jinzheng He and Junyang Lin and Kai Dang and Keming Lu and Keqin Chen and Kexin Yang and Mei Li and Mingfeng Xue and Na Ni and Pei Zhang and Peng Wang and Ru Peng and Rui Men and Ruize Gao and Runji Lin and Shijie Wang and Shuai Bai and Sinan Tan and Tianhang Zhu and Tianhao Li and Tianyu Liu and Wenbin Ge and Xiaodong Deng and Xiaohuan Zhou and Xingzhang Ren and Xinyu Zhang and Xipin Wei and Xuancheng Ren and Yang Fan and Yang Yao and Yichang Zhang and Yu Wan and Yunfei Chu and Yuqiong Liu and Zeyu Cui and Zhenru Zhang and Zhihao Fan},
      journal={arXiv preprint arXiv:2407.10671},
      year={2024},
      doi={10.48550/arXiv.2407.10671}
}

@inproceedings{savcisens2025trilemma,
  title={Trilemma of Truth in Large Language Models},
  author={Savcisens, Germans and Eliassi-Rad, Tina},
  booktitle={Mechanistic Interpretability Workshop at Neur{IPS} 2025},
  year={2025},
  note={\url{https://openreview.net/forum?id=z7dLG2ycRf}},
}

@inproceedings{marks2024geometry,
  title={The Geometry of Truth: {E}mergent Linear Structure in Large Language Model Representations of {T}rue/{F}alse Datasets},
  year={2024},
  author={Marks, Samuel and Tegmark, Max},
  booktitle={Proceedings of the 1st Conference on Language Modeling (COLM 2024)},
  note={\url{https://openreview.net/forum?id=aajyHYjjsk}}
}

@article{cortes1995support,
  title={Support-vector networks},
  author={Cortes, Corinna and Vapnik, Vladimir},
  journal={Machine {{L}}earning},
  volume={20},
  pages={273--297},
  year={1995},
  doi={10.1007/BF00994018}
}

@article{leitgeb2014stability,
  title={The stability theory of belief},
  author={Leitgeb, Hannes},
  journal={Philosophical Review},
  volume={123},
  number={2},
  pages={131--171},
  year={2014},
  doi={10.1215/00318108-2400575}
}

@article{cooper1968propositional,
  title={The propositional logic of ordinary discourse},
  author={Cooper, William S},
  journal={Inquiry},
  volume={11},
  number={1-4},
  pages={295--320},
  year={1968},
  doi={10.1080/00201746808601531}
}

@inproceedings{de1936logique,
  title={La logique de la probabilit{\'e}},
  author={De Finetti, Bruno},
  booktitle={Actes du congr{\`e}s international de philosophie scientifique},
  volume={4},
  pages={31--39},
  year={1936},
  doi={10.5840/icus11936454}
}

@article{foley1992epistemology,
  title={The epistemology of belief and the epistemology of degrees of belief},
  author={Foley, Richard},
  journal={American Philosophical Quarterly},
  volume={29},
  number={2},
  pages={111--124},
  year={1992},
  note={\url{http://www.jstor.org/stable/20014406}}
}

@article{leitgeb_humean,
    author = {Leitgeb, Hannes},
    title = {I--{T}he {H}umean Thesis on Belief},
    journal = {Aristotelian Society Supplementary Volume},
    volume = {89},
    number = {1},
    pages = {143--185},
    year = {2015},
    month = {June},
    issn = {0309-7013},
    doi = {10.1111/j.1467-8349.2015.00248.x},
    url = {https://doi.org/10.1111/j.1467-8349.2015.00248.x},
}

@article{egre2026probability,
  title={Probability for trivalent conditionals},
  author={{\'E}gr{\'e}, Paul and Rossi, Lorenzo and Sprenger, Jan Michael},
  journal={Mind forthcoming},
  year={2026},
  note={\url{https://iris.unito.it/bitstream/2318/2142754/1/CP\%2BLearning_rev7.pdf}}
}

@article{egre2025certain,
  title={Certain and uncertain inference with indicative conditionals},
  author={{\'E}gr{\'e}, Paul and Rossi, Lorenzo and Sprenger, Jan},
  journal={Australasian Journal of Philosophy},
  volume={103},
  number={3},
  pages={569--596},
  year={2025},
  doi={10.1080/00048402.2025.2475882}
}

@article{cantwell2008logic,
  title={The logic of conditional negation},
  author={Cantwell, John},
  year={2008},
  journal={Notre Dame Journal of Formal Logic},
  volume={49},
  number={3},
  pages={245--260},
  doi={10.1215/00294527-2008-010}
}

@book{adams1975logic,
  author    = {Adams, Ernest W.},
  title     = {The Logic of Conditionals: An Application of Probability to Deductive Logic},
  series    = {Synthese Library},
  volume    = {86},
  publisher = {D. Reidel Publishing Company},
  address   = {Dordrecht},
  year      = {1975},
  doi={10.1007/978-94-015-7622-2},
  note      = {\url{https://doi.org/10.1007/978-94-015-7622-2}}
}

@article{molinari2024accuracy,
  title={An accuracy characterisation of approximate coherence},
  author={Molinari, Giacomo},
  journal={Synthese},
  volume={203},
  number={2},
  pages={68},
  year={2024},
  publisher={Springer},
  doi={10.1007/s11229-024-04490-6}
}

@article{cantwell2006laws,
  title={The laws of non-bivalent probability},
  author={Cantwell, John},
  journal={Logic and Logical Philosophy},
  volume={15},
  number={2},
  pages={163--171},
  year={2006},
  note={\url{https://apcz.umk.pl/LLP/article/view/LLP.2006.010}}
}

@article{dies2025,
  title={Epistemic Familiarity is Associated With Belief Stability in Large Language Models},
  author={Dies, Samantha and Maynard, Courtney and Savcisens, Germans and Eliassi-Rad, Tina},
  journal={arXiv preprint arXiv:2511.19166},
  year={2025},
  doi={10.48550/arXiv.2511.19166}
}

@article{phipson2016permutation,
title = {Permutation P-values Should Never Be Zero: {C}alculating Exact P-values When Permutations Are Randomly Drawn},
author = {Belinda Phipson and Gordon K Smyth},
volume = {9},
number = {1},
pages   = {Article 39},
journal = {Statistical Applications in Genetics and Molecular Biology},
doi = {10.2202/1544-6115.1585},
year = {2010},
}

@article{wolf2026partition,
  title={Partition, Prompt, Aggregate: Statistical Self-Consistency in Language Models},
  author={Wolf, Patrik and Kleine Buening, Thomas and Krause, Andreas and Mendler-D{\"u}nner, Celestine},
  journal={arXiv preprint arXiv:2607.15277},
  year={2026},
  doi={10.48550/arXiv.2607.15277}
}

@article{elazar2021measuring,
  title={Measuring and improving consistency in pretrained language models},
  author={Elazar, Yanai and Kassner, Nora and Ravfogel, Shauli and Ravichander, Abhilasha and Hovy, Eduard and Sch{\"u}tze, Hinrich and Goldberg, Yoav},
  journal={Transactions of the Association for Computational Linguistics},
  volume={9},
  pages={1012--1031},
  year={2021},
  doi={10.1162/tacl_a_00410}
}

@inproceedings{sharma2024towards,
  title={Towards Understanding Sycophancy in Language Models},
  author={Mrinank Sharma and Meg Tong and Tomasz Korbak and David Duvenaud and Amanda Askell and Samuel R. Bowman and Esin Durmus and Zac Hatfield-Dodds and Scott R. Johnston and Shauna M. Kravec and Timothy Maxwell and Sam McCandlish and Kamal Ndousse and Oliver Rausch and Nicholas Schiefer and Da Yan and Miranda Zhang and Ethan Perez},
  booktitle={Proceedings of the 12th International Conference on Learning Representations (ICLR 2024)},
  year={2024},
  note={\url{https://openreview.net/forum?id=tvhaxkMKAn}},
}

@inproceedings{li2025firm,
    title = "Firm or Fickle? {E}valuating Large Language Models Consistency in Sequential Interactions",
    author={Li, Yubo and Miao, Yidi and Ding, Xueying and Krishnan, Ramayya and Padman, Rema},
    editor = "Che, Wanxiang  and
      Nabende, Joyce  and
      Shutova, Ekaterina  and
      Pilehvar, Mohammad Taher",
    booktitle = "Findings of the Association for Computational Linguistics (ACL 2025)",
    year = "2025",
    publisher = "Association for Computational Linguistics",
    url = "https://aclanthology.org/2025.findings-acl.347/",
    doi = "10.18653/v1/2025.findings-acl.347",
    pages = "6679--6700",
    ISBN = "979-8-89176-256-5",
}

@incollection{spearman1961proof,
  author    = {Spearman, Charles},
  title     = {The Proof and Measurement of Association Between Two Things},
  booktitle = {Studies in Individual Differences: The Search for Intelligence},
  editor    = {Jenkins, James J. and Paterson, Donald G.},
  pages     = {45--58},
  publisher = {Appleton-Century-Crofts},
  year      = {1961},
  doi       = {10.1037/11491-005}
}

@article{dykstra1983algorithm,
  title={An algorithm for restricted least squares regression},
  author={Dykstra, Richard L},
  journal={Journal of the American Statistical Association},
  volume={78},
  number={384},
  pages={837--842},
  year={1983},
  doi={10.1080/01621459.1983.10477029}
}

@article{alkhamissi2022review,
  title={A review on language models as knowledge bases},
  author={AlKhamissi, Badr and Li, Millicent and Celikyilmaz, Asli and Diab, Mona and Ghazvininejad, Marjan},
  journal={arXiv preprint arXiv:2204.06031},
  year={2022},
  doi={10.48550/arXiv.2204.06031}
}

@article{augenstein2024factuality,
  title={Factuality challenges in the era of large language models and opportunities for fact-checking},
  author={Augenstein, Isabelle and Baldwin, Timothy and Cha, Meeyoung and Chakraborty, Tanmoy and Ciampaglia, Giovanni Luca and Corney, David and DiResta, Renee and Ferrara, Emilio and Hale, Scott and Halevy, Alon and others},
  journal={Nature Machine Intelligence},
  volume={6},
  number={8},
  pages={852--863},
  year={2024},
  doi={10.1038/s42256-024-00881-z}
}

@InProceedings{band_calibration,
  title = 	 {Linguistic Calibration of Long-Form Generations},
  author =       {Band, Neil and Li, Xuechen and Ma, Tengyu and Hashimoto, Tatsunori},
  booktitle = 	 {Proceedings of the 41st International Conference on Machine Learning},
  pages = 	 {2732--2778},
  year = 	 {2024},
  editor = 	 {Salakhutdinov, Ruslan and Kolter, Zico and Heller, Katherine and Weller, Adrian and Oliver, Nuria and Scarlett, Jonathan and Berkenkamp, Felix},
  volume = 	 {235},
  series = 	 {Proceedings of Machine Learning Research},
  month = 	 {July},
  publisher =    {PMLR},
  url = 	 {https://proceedings.mlr.press/v235/band24a.html},
  note={\url{https://proceedings.mlr.press/v235/band24a.html}}
}

@article{yadkori2024believe,
  title={To believe or not to believe your {LLM}: Iterative prompting for estimating epistemic uncertainty},
  author={Abbasi-Yadkori, Yasin and Kuzborskij, Ilja and Gy{\"o}rgy, Andr{\'a}s and Szepesv{\'a}ri, Csaba},
  journal={Advances in Neural Information Processing Systems},
  volume={37},
  pages={58077--58117},
  year={2024},
  note={\url{https://openreview.net/forum?id=k6iyUfwdI9}}
}

@article{kapoor2024large,
  title={Large language models must be taught to know what they don't know},
  author={Kapoor, Sanyam and Gruver, Nate and Roberts, Manley and Collins, Katherine and Pal, Arka and Bhatt, Umang and Weller, Adrian and Dooley, Samuel and Goldblum, Micah and Wilson, Andrew G},
  journal={Advances in Neural Information Processing Systems},
  volume={37},
  pages={85932--85972},
  year={2024},
  doi={10.52202/079017-2729}
}

@article{steyvers2025large,
  title={What large language models know and what people think they know},
  author={Steyvers, Mark and Tejeda, Heliodoro and Kumar, Aakriti and Belem, Catarina and Karny, Sheer and Hu, Xinyue and Mayer, Lukas W and Smyth, Padhraic},
  journal={Nature Machine Intelligence},
  volume={7},
  number={2},
  pages={221--231},
  year={2025},
  doi={10.1038/s42256-024-00976-7}
}

@inproceedings{gao-etal-2024-spuq,
    title = "{SPUQ}: {P}erturbation-Based Uncertainty Quantification for Large Language Models",
    author={Gao, Xiang and Zhang, Jiaxin and Mouatadid, Lalla and Das, Kamalika},
    editor = "Graham, Yvette  and
      Purver, Matthew",
    booktitle = "Proceedings of the 18th Conference of the European Chapter of the Association for Computational Linguistics (Volume 1: Long Papers)",
    year = "2024",
    publisher = "Association for Computational Linguistics",
    url = "https://aclanthology.org/2024.eacl-long.143/",
    doi = "10.18653/v1/2024.eacl-long.143",
    pages = "2336--2346",
}

@InProceedings{ahdritz24a_knowable,
  title = 	 {Distinguishing the Knowable from the Unknowable with Language Models},
  author =       {Ahdritz, Gustaf and Qin, Tian and Vyas, Nikhil and Barak, Boaz and Edelman, Benjamin L.},
  booktitle = 	 {Proceedings of the 41st International Conference on Machine Learning},
  pages = 	 {503--549},
  year = 	 {2024},
  editor = 	 {Salakhutdinov, Ruslan and Kolter, Zico and Heller, Katherine and Weller, Adrian and Oliver, Nuria and Scarlett, Jonathan and Berkenkamp, Felix},
  volume = 	 {235},
  series = 	 {Proceedings of Machine Learning Research},
  month = 	 {July},
  publisher =    {PMLR},
  url = 	 {https://proceedings.mlr.press/v235/ahdritz24a.html},
  note={\url{https://proceedings.mlr.press/v235/ahdritz24a.html}}
}

@article{ying2026truthfulness,
  title={The Truthfulness Spectrum Hypothesis},
  author={Ying, Zhuofan Josh and Ravfogel, Shauli and Kriegeskorte, Nikolaus and Hase, Peter},
  journal={arXiv preprint arXiv:2602.20273},
  year={2026},
  doi={10.48550/arXiv.2602.20273}
}

@article{kumaran2026competing,
  title={Competing biases underlie overconfidence and underconfidence in {LLM}s},
  author={Kumaran, Dharshan and Fleming, Stephen M and Markeeva, Larisa and Heyward, Joe and Banino, Andrea and Mathur, Mrinal and Pascanu, Razvan and Osindero, Simon and De Martino, Benedetto and Veli{\v{c}}kovi{\'c}, Petar and others},
  journal={Nature Machine Intelligence},
  volume={8},  
  pages={614--627},
  year={2026},
  doi={10.1038/s42256-026-01217-9}
}

@article{suzgun2025language,
  title={Language models cannot reliably distinguish belief from knowledge and fact},
  author={Suzgun, Mirac and Gur, Tayfun and Bianchi, Federico and Ho, Daniel E. and Icard, Thomas and Jurafsky, Dan and Zou, James},
  journal={Nature Machine Intelligence},
  volume={7},
  number={11},
  pages={1780--1790},
  year={2025},
  doi={10.1038/s42256-025-01113-8}
}

@InProceedings{yin24c_truthfulness,
  title = 	 {Characterizing Truthfulness in Large Language Model Generations with Local Intrinsic Dimension},
  author =       {Yin, Fan and Srinivasa, Jayanth and Chang, Kai-Wei},
  booktitle = 	 {Proceedings of the 41st International Conference on Machine Learning},
  pages = 	 {57069--57084},
  year = 	 {2024},
  editor = 	 {Salakhutdinov, Ruslan and Kolter, Zico and Heller, Katherine and Weller, Adrian and Oliver, Nuria and Scarlett, Jonathan and Berkenkamp, Felix},
  volume = 	 {235},
  series = 	 {Proceedings of Machine Learning Research},
  month = 	 {July},
  publisher =    {PMLR},
  url = 	 {https://proceedings.mlr.press/v235/yin24c.html},
  note={\url{https://proceedings.mlr.press/v235/yin24c.html}}
}

@inproceedings{kassner2021beliefbank,
  title={{BeliefBank: A}dding memory to a pre-trained language model for a systematic notion of belief},
  author={Kassner, Nora and Tafjord, Oyvind and Sch{\"u}tze, Hinrich and Clark, Peter},
  booktitle={Proceedings of the 2021 Conference on Empirical Methods in Natural Language Processing},
  pages={8849--8861},
  year={2021},
  doi={10.18653/v1/2021.emnlp-main.697}
}

@article{cohen2024evaluating,
    author = {Cohen, Roi and Biran, Eden and Yoran, Ori and Globerson, Amir and Geva, Mor},
    title = {Evaluating the Ripple Effects of Knowledge Editing in Language Models},
    journal = {Transactions of the Association for Computational Linguistics},
    volume = {12},
    pages = {283--298},
    year = {2024},
    month = {04},
    issn = {2307-387X},
    doi = {10.1162/tacl_a_00644},
    url = {https://doi.org/10.1162/tacl_a_00644},
}

@article{sarkar2026language,
  title={Language Models Encode the Contextual Truth of Propositions},
  author={Sarkar, Rupak and Ramu, Pritika and Rudinger, Rachel},
  journal={arXiv preprint arXiv:2608.03035},
  year={2026},
  doi={10.48550/arXiv.2608.03035}
}

@inproceedings{corona-mendozza-sogaard-2026-llm,
    title = "{LLM} Beliefs Are in Their Heads",
    author = "Corona Mendozza, Alessandro  and
      S{\o}gaard, Anders",
    editor = "Liakata, Maria  and
      Moreira, Viviane P.  and
      Zhang, Jiajun  and
      Jurgens, David",
    booktitle = "Proceedings of the 64th Annual Meeting of the {A}ssociation for {C}omputational {L}inguistics (Volume 1: Long Papers)",
    month = jul,
    year = "2026",
    address = "San Diego, California, United States",
    publisher = "Association for Computational Linguistics",
    url = "https://aclanthology.org/2026.acl-long.1905/",
    doi = "10.18653/v1/2026.acl-long.1905",
    pages = "41033--41067",
    ISBN = "979-8-89176-390-6",
}

@article{cheon2026robust,
  title={Robust for the Wrong Reasons: The Representational Geometry of {LLM} Robustness to Science Skepticism},
  author={Cheon, Minjong},
  journal={arXiv preprint arXiv:2607.01951},
  year={2026},
  doi={10.48550/arXiv.2607.01951}
}

@article{wishart2018drugbank,
  title={{D}rug{B}ank 5.0: a major update to the {D}rug{B}ank database for 2018},
  author={Wishart, David S and Feunang, Yannick D and Guo, An C and Lo, Elvis J and Marcu, Ana and Grant, Jason R and Sajed, Tanvir and Johnson, Daniel and Li, Carin and Sayeeda, Zinat and others},
  journal={Nucleic Acids Research},
  volume={46},
  number={D1},
  pages={D1074--D1082},
  year={2018},
  publisher={Oxford University Press},
  doi={10.1093/nar/gkx1037}
}

%%%%%%%%%%%%%%%%%%%%%%%%%%%%%%%%%%%%%%%%%%%%
%%              Appendices                %%
%%%%%%%%%%%%%%%%%%%%%%%%%%%%%%%%%%%%%%%%%%%%

%%%%%%%%%%%%%%%%%%%%%%%%%%%%%%%%%%%%%%%%%%%%
%%              Appendices                %%
%%%%%%%%%%%%%%%%%%%%%%%%%%%%%%%%%%%%%%%%%%%%

\clearpage
\FloatBarrier
\begin{appendices}
\onecolumn

% Restart counters, and have floats in appendices labeled with A-prefix.
\setcounter{figure}{0}
\setcounter{table}{0}
\renewcommand{\thefigure}{A\arabic{figure}}
\renewcommand{\thetable}{A\arabic{table}}

%%%%%%%%%%%%%%%%%%%%%%%%%%%%%%%%%%%%%%%%%%%%
%%         Appendices: Notation           %%
%%%%%%%%%%%%%%%%%%%%%%%%%%%%%%%%%%%%%%%%%%%%

\section{Notation}
\label{appendix:notation}

We summarize the principal mathematical notation used throughout the manuscript in
Table~\ref{appendix:tab:notation}. Where clear from context, indices for semantic domain, probe, and stability operationalization are suppressed. Accordingly, empirical quantities written with subscript $\mathcal{M}$ are instantiated separately for each model--dataset--probe combination. Graded stability additionally depends on the conditional-probability operationalization.

\begin{table*}[h]
\centering
\resizebox{\textwidth}{!}{%
\begin{tabular}{ll}
\toprule
\textbf{Symbol} & \textbf{Description} \\
\midrule

$\mathcal{M}$ &
A fixed large language model (LLM). \\

$s$ &
Natural-language statement. \\

$P$ &
A proposition believed by $\mathcal{M}$ whose stability is evaluated. \\

$x$ &
A non-disbelieved proposition used as conditioning information. \\

$c \in \{T,F,N\}$ &
Trivalent veracity class: \texttt{True} ($T$), \texttt{False} ($F$), or \texttt{Neither} ($N$). \\

$K$ &
Number of probe classes. \\

$\ell$ &
Layer index used for activation extraction. \\

$\ell^\star$ &
Layer selected by minimizing calibration log loss. \\

$z_i^{(\ell)}$ &
Hidden representation of statement $s_i$. \\

$\mathbf{r}(s_i)$ &
Vector of raw probe scores for statement $s_i$. \\

$\mathcal{L}_{\mathrm{log}}^{(\ell)}$ &
Multiclass log loss of the calibrated probe at layer $\ell$. \\

$\hat y_i$ &
Probe-predicted epistemic state of statement $s_i$. \\

$\boldsymbol{\pi}_\mathcal{M}(s_i)$ &
Probe-estimated trivalent distribution for $s_i$. \\

$\boldsymbol{\pi}_\mathcal{M}(P,x)$ &
Probe-estimated distribution associated with pair $(P,x)$, with Direct Conditional and Joint-to-Conditional variants. \\

$\Pr_\mathcal{M}(s)$ &
Probability that statement $s$ is true. \\

$\Pr_\mathcal{M}(P\mid x)$ &
Conditional probability that proposition $P$ is true given proposition $x$. \\

$\widehat{\Pr}_\mathcal{M}(P\mid x)$ &
Estimate of $\Pr_\mathcal{M}(P\mid x)$ using either the Direct Conditional or Joint-to-Conditional approach. \\

$\mathcal{B}_\mathcal{M}$ &
Belief set of 
$\mathcal{M}$. \\

$\mathcal{X}_\mathcal{M}$ &
Non-disbelief set of $\mathcal{M}$. \\

$t_\mathcal{M}$ &
Empirical belief threshold for $\mathcal{M}$, defined from the lowest \texttt{True}-class probability among beliefs $P \in \mathcal{B}_\mathcal{M}$. \\

$\gamma_\mathcal{M}(P)$ &
Graded belief stability of $P$ under $\mathcal{M}$ for either the Direct Conditional or Joint-to-Conditional estimates. \\

$\overline{\gamma}_\mathcal{M}$ &
Mean graded belief stability across beliefs in $\mathcal{B}_\mathcal{M}$. \\

$\widehat{\gamma}_\mathcal{M}(P) = f_\mathcal{M}(\pi_T(P))$ &
Predicted graded stability of $P$ based on its belief probability. \\

$r_\mathcal{M}(P)$ &
Residual graded stability after accounting for individual belief probability. \\

$\mathbf{v}_{P,x}$ &
Nine-dimensional vector collecting the measured trivalent probabilities. \\

$\mathcal{C}$ &
Set of CCK-coherent probability vectors. \\

$\Pi_\mathcal{C}(\mathbf{v})$ &
Euclidean projection onto the set of CCK-coherent probability vectors $\mathcal{C}$. \\

$d_\mathrm{CCK}(\mathbf{v}_{P,x})$ &
RMS adjustment required to make the measured probability system CCK-coherent. \\

$P_H,\;P_L$ &
Higher- and lower-stability propositions within a pair matched on individual belief probability. \\

$k$ &
Round in the behavioral challenge experiment. \\

$p_{\mathcal{M},k}^{c}(P)$ &
Probability assigned at round $k$ to candidate behavioral response $c$ for proposition $P$. \\

$q_{\mathcal{M},k}^{c}(P)$ &
Probability of behavioral response $c$ after normalizing over the permitted \texttt{True}/\texttt{False} responses. \\

$\hat c_k(P)$ &
Binary truth judgment selected by $\mathcal{M}$ for proposition $P$ at behavioral round $k$. \\

$u_k(P)$ &
Normalized probability assigned at round $k$ to the response class initially selected for proposition $P$. \\

$M(P)$ &
Behavioral movement of $P$ across challenge rounds. \\

$\Delta M$ &
Difference in movement between lower- and higher-stability matched beliefs. \\

\bottomrule
\end{tabular}
}
\caption{\textbf{Notation.} Summary of the principal mathematical symbols used throughout the manuscript.}
\label{appendix:tab:notation}
\end{table*}

%%%%%%%%%%%%%%%%%%%%%%%%%%%%%%%%%%%%%%%%%%%%
%%          Appendices: Datasets          %%
%%%%%%%%%%%%%%%%%%%%%%%%%%%%%%%%%%%%%%%%%%%%

\FloatBarrier
\section{Datasets}
\label{sec:si:data}

\subsection{Dataset construction}
\label{sec:si:data_construction}

We use the City Locations, Medical Indications, and Word Definitions datasets introduced by~\cite{savcisens2025trilemma}. Each dataset contains statements labeled \texttt{True}, \texttt{False}, or \texttt{Neither}, with approximately balanced affirmative and negated forms. The three domains capture geographic containment, drug--indication relationships, and lexical relationships, respectively. The \texttt{True} and \texttt{False} statements are constructed from real-world entities and relations, while the \texttt{Neither} statements pair synthetically generated entities for which no corresponding real-world fact is intended to exist.

For the \texttt{True} and \texttt{False} classes, each domain contains both correct and incorrect entity--relation pairs, and each pair is expressed in both affirmative and negated form. For a correct entity--relation pair, the affirmative statement is \texttt{True} and its negation is \texttt{False}. Incorrect pairs are generated by shuffling entities and relations so that the affirmative statement is \texttt{False} and its negation is \texttt{True}.

\paragraph{City Locations.}
City Locations is constructed from the GeoNames geographic database\footnote{\url{https://www.geonames.org/}.}~\cite{savcisens2025trilemma}. Cities are required to have populations of at least $30{,}000$ and an associated country, and locations in Antarctica are excluded. The resulting collection retains $1{,}400$ unique city names. Statements are generated with the template \textit{``The city of [city] is (not) located in [country],''}, where \textit{``The city of''} is omitted when redundant.

\paragraph{Medical Indications.}
Medical Indications is constructed from DrugBank version 5.1.12~\cite{savcisens2025trilemma, wishart2018drugbank}. Drug names and indication descriptions are extracted from the database, with only the first sentence retained for multi-sentence indications. Diseases and conditions are extracted using both \texttt{SciSpacy} and a BioBERT-based named-entity recognition model and retained only when identified by both models\footnote{\url{https://huggingface.co/alvaroalon2/biobert_diseases_ner}.}. Drug names are additionally required to be recognized as chemical entities by \texttt{SciSpacy}, and low-frequency drug--indication pairs are removed using \texttt{wordfreq}. Statements are generated with the template \textit{``[drug] is (not) indicated for the treatment of [disease/condition].''}

\paragraph{Word Definitions.}
Word Definitions is constructed from the publicly available sample of \texttt{WordsAPI}\footnote{\url{https://www.wordsapi.com/}.}~\cite{savcisens2025trilemma}. The dataset retains nouns with at least one definition and at least one \texttt{synonym}, \texttt{typeOf}, or \texttt{instanceOf} relation. These relations are rendered using three corresponding statement templates: \textit{``[word] is (not) a [instanceOf],''} \textit{``[word] is (not) a type of [typeOf],''} and \textit{``[word] is (not) a synonym of [synonym].''} Articles and singular forms are adjusted where necessary to preserve grammaticality.

\subsubsection{Synthetic \texttt{Neither} statements}

The \texttt{Neither} class is designed to represent claims for which the model has no corresponding real-world fact to recover. Rather than relying on naturally occurring obscure entities, whose presence in LLM training corpora cannot be established, synthetic entities are generated from the lexical statistics of the real entities in each domain. The \texttt{namemaker} package\footnote{\url{https://github.com/Rickmsd/namemaker}.} is used to generate new names with character bigram Markov-chain models fit to the corresponding real entity sets~\cite{savcisens2025trilemma}. Synthetic entities are then paired using the same statement templates as the real-world data.

Generated entities undergo domain-specific filtering intended to reduce accidental overlap with existing entities. City and country names are checked against \texttt{GeoNames} and subsequently screened using targeted web searches. Drug and disease names are checked against biomedical entity resources and similarly screened for real-world matches. Finally, synthetic lexical items are compared against multiple English word lists. Candidates that match existing entities are removed before the remaining synthetic entities are randomly paired to construct \texttt{Neither} statements~\cite{savcisens2025trilemma}.

These validation steps are intended to minimize the probability that a synthetic entity corresponds to an existing entity or lexical item. Because the training corpora of the evaluated LLMs are not fully observable, we cannot guarantee that every synthetic string is entirely absent from pretraining data. We therefore interpret the \texttt{Neither} examples as controlled, intentionally unfamiliar claims for which the model lacks an established real-world truth value, rather than as a direct test of training-data absence.

\begin{table*}[tb]
\centering
\begin{tabular}{lcccc}
\toprule
\textbf{Dataset} & \textbf{Train} & \textbf{Calibration} & \textbf{Test} & \textbf{Total} \\
\midrule

City Locations &
$3999$ $(0.55)$ &
$1398$ $(0.19)$ &
$1855$ $(0.26)$ &
$7252$ $(1.00)$ \\

Medical Indications &
$3849$ $(0.56)$ &
$1327$ $(0.19)$ &
$1727$ $(0.25)$ &
$6903$ $(1.00)$ \\

Word Definitions &
$4717$ $(0.55)$ &
$1628$ $(0.19)$ &
$2155$ $(0.25)$ &
$8500$ $(1.00)$ \\

\bottomrule
\end{tabular}
\caption{\textbf{Dataset splits.} Number of statements used for training, calibration, and testing. Proportions of the full dataset are reported in parentheses. A version of this table appears in~\cite{savcisens2025trilemma}.}
\label{appendix:tab:data_splits}
\end{table*}

\subsection{Dataset splits}
\label{sec:si:data_splits}

We use the fixed train, calibration, and test partitions introduced with the original datasets~\cite{savcisens2025trilemma}. The datasets are partitioned into approximately $55\%$ training, $20\%$ calibration, and $25\%$ test data, with entities kept exclusive across splits. If an entity occurs in one partition, all statements containing that entity are assigned to the same partition. Table~\ref{appendix:tab:data_splits} reports the resulting split sizes.

The same underlying statement partitions are used throughout all probability-estimation procedures. The individual-statement probe is fit to statements from the training split and calibrated using individual statements from the calibration split. The Direct Conditional and Joint probes do not independently repartition the data; instead, their combined training and calibration statements are constructed exclusively from statements already assigned to the corresponding individual-statement training and calibration partitions. Likewise, the test statements define the empirical belief and non-disbelief sets used to construct the $(P,x)$ pairs evaluated by both conditional estimators. Consequently, no statement or entity assigned to the test partition is used to fit or calibrate the individual-statement, Direct Conditional, or Joint probes.

%%%%%%%%%%%%%%%%%%%%%%%%%%%%%%%%%%%%%%%%%%%%
%%        Appendices: Base Models         %%
%%%%%%%%%%%%%%%%%%%%%%%%%%%%%%%%%%%%%%%%%%%%
\begin{table}[tb]
\centering
\resizebox{\textwidth}{!}{%
\begin{tabular}{llccclc}
\toprule
\textbf{Official Name} & \textbf{Short Name} & \textbf{\# Layers} & \textbf{\# Parameters} & \textbf{Release Date} & \textbf{Source} & \textbf{Citation}\\
\midrule

Gemma-$7$b &
\texttt{gemma-7b (b)} &
$28$ &
$8.54$ B &
Feb $21$, $2024$ &
Google &
\cite{gemma_2024} \\

Gemma-$2$-$9$b &
\texttt{gemma-2-9b (b)} &
$42$ &
$9.24$ B &
Jun $27$, $2024$ &
Google &
\cite{gemma_2024} \\

Gemma-$2$-$27$b &
\texttt{gemma-2-27b (b)} &
$46$ &
$27.23$ B &
Jun $27$, $2024$ &
Google &
\cite{gemma_2024} \\

Llama-$3.2$-$3$b &
\texttt{llama-3.2-3b (b)} &
$28$ &
$3.21$ B &
Sep $25$, $2024$ &
Meta &
\cite{llama3} \\

Llama-$3.1$-$8$b &
\texttt{llama-3.1-8b (b)} &
$32$ &
$8.03$ B &
Jul $23$, $2024$ &
Meta &
\cite{llama3} \\

Llama-$3.1$-$70$b &
\texttt{llama-3.1-70b (b)} &
$80$ &
$70.55$ B &
Jul $23$, $2024$ &
Meta &
\cite{llama3} \\

Mistral-$7$B-v$0.3$ &
\texttt{mistral-7b (b)} &
$32$ &
$7.25$ B &
May $22$, $2024$ &
Mistral AI &
\cite{mistral7b} \\

Mistral-Nemo-Base-2407 &
\texttt{mistral-12b (b)} &
$40$ &
$12.25$ B &
Jul $18$, $2024$ &
Mistral AI &
\cite{mistralnemo} \\

Mistral-Small-$3.1$-$24$B-Base-2503 &
\texttt{mistral-3.1-24b (b)} &
$40$ &
$23.57$ B &
Mar $17$, $2025$ &
Mistral AI &
\cite{mistralsmall31} \\

Qwen$2.5$-$7$B &
\texttt{qwen-2.5-7b (b)} &
$28$ &
$7.62$ B &
Sep $19$, $2024$ &
Alibaba Cloud &
\cite{qwen2, qwen2.5} \\

Qwen$2.5$-$14$B &
\texttt{qwen-2.5-14b (b)} &
$48$ &
$14.80$ B &
Sep $19$, $2024$ &
Alibaba Cloud &
\cite{qwen2, qwen2.5} \\

Qwen$2.5$-$72$B &
\texttt{qwen-2.5-72b (b)} &
$80$ &
$72.70$ B &
Sep $19$, $2024$ &
Alibaba Cloud &
\cite{qwen2, qwen2.5} \\

\bottomrule
\end{tabular}
}
\caption{\textbf{Base LLMs used in graded stability experiments.} We list the official names of the LLMs according to the HuggingFace repository~\cite{wolf2020transformers}, where they are publicly available. We further specify the shortened name used throughout the paper, the number of layers, parameter count, release date, source organization, and official citation.}
\label{tab:base_LLMs}
\end{table}

\section{Base models}
\label{sec:si:base_models}

To test whether the primary findings depend on instruction tuning, we repeat the four main figure-level analyses using the pretrained base counterparts of the $12$ instruction-tuned LLMs considered in the main text. Results are reported in Section~\ref{sec:si:base_robustness}. These models are matched to the primary models by family and parameter scale, allowing us to evaluate the robustness of the observed graded-stability patterns outside the instruction-tuned setting. Exact model versions and specifications are reported in Table~\ref{tab:base_LLMs}.

%%%%%%%%%%%%%%%%%%%%%%%%%%%%%%%%%%%%%%%%%%%%
%%        Appendices: Probe details       %%
%%%%%%%%%%%%%%%%%%%%%%%%%%%%%%%%%%%%%%%%%%%%

\FloatBarrier
\section{\texttt{sAwMIL} training, calibration, and evaluation}
\label{sec:si:probes}

\subsection{Probability calibration}
\label{sec:si:probability_calibration}

The \texttt{sAwMIL} probe produces a vector of raw max-margin scores rather than probabilities. For a $K$-class probe, let
\begin{equation}
    \mathbf{r}(s)
    =
    [r_1(s),\ldots,r_K(s)]
\end{equation}
denote the raw score vector for statement $s$. Each $r_k(s)$ is obtained by applying the corresponding one-versus-all \texttt{sAwMIL} head to every valid token representation in the statement and taking the maximum token-level margin.

We fit the probe heads using only the training split and reserve the calibration split for converting these raw scores into probability distributions. After fitting the probe, we compute $\mathbf{r}(s)$ for every statement $s$ in the corresponding calibration set and fit a multinomial logistic-regression model that predicts the ground-truth class from the full $K$-dimensional score vector. For class $k$, the resulting calibrated probability is
\begin{equation}
    \pi_k(s)
    =
    \frac{
        \exp\!\left(\boldsymbol{\beta}_k^\top \mathbf{r}(s) + b_k\right)
    }{
        \sum_{j=1}^{K}
        \exp\!\left(\boldsymbol{\beta}_j^\top \mathbf{r}(s) + b_j\right)
    }.
\end{equation}
The calibration model is fit by minimizing multinomial log loss using the \texttt{L-BFGS} solver, with inverse regularization strength $C=1.0$, a maximum of $2000$ iterations, and convergence tolerance $10^{-6}$. The resulting probabilities are aligned to the fixed probe class ordering, clipped to $[10^{-15},1]$ for numerical stability, and renormalized to sum to one.

A separate calibration mapping is fit for every probe. Individual-statement and Direct Conditional probes therefore produce calibrated three-class distributions over $\{\texttt{True},\texttt{False},\texttt{Neither}\}$, while Joint probes produce calibrated nine-class distributions over the ordered joint states in $\{\texttt{TT}, \texttt{TF}, \texttt{TN}, \texttt{NT}, \texttt{NF}, \texttt{NN}, \texttt{FT}, \texttt{FF}, \texttt{FN}\}$. In each case, calibration parameters are estimated only from the corresponding calibration examples and are held fixed for all subsequent analyses.

\subsection{Specifications and hyperparameters}
\label{sec:si:probe_hyperparameters}

\texttt{sAwMIL} treats each statement as a bag of token-level hidden representations from a single transformer layer. Specifically, we use the output hidden state of the selected layer at every non-padding token position rather than reducing the statement to a single representation. Model inputs are padded or truncated to a maximum sequence length of $64$ tokens.

Before probe fitting, each activation dimension is standardized to zero mean and unit variance using a \texttt{StandardScaler} fit to the pooled token representations from the training split. The same fitted transformation is then applied to calibration and test examples.

For the $K$ veracity classes, we train $K$ independent one-versus-all \texttt{sAwMIL} heads. For head $k$, statements belonging to class $k$ are treated as positive bags and all remaining statements as negative bags. Each head is trained in two stages. First, we fit a linear hinge-loss classifier to the initialized token labels. We then apply this classifier to the instances within each positive bag and select the $k_{\mathrm{top}}=2$ highest-scoring instances. To encourage sparse attribution near the end of the statement, selected instances are retained as positive only when they also occur among the final $\texttt{tail\_k}=2$ token positions; all other instances are relabeled as negative. A second classifier with the same specification is then fit to these relabeled instances. At inference time, the final classifier assigns a margin to every valid token, and the maximum token-level margin within the bag becomes the statement-level score for that class.

Both stages use an averaged linear \texttt{SGDClassifier} with hinge loss, an intercept, and $\ell_2$ regularization. We set $C=1.0$ and parameterize the SGD regularization coefficient as
\begin{equation}
    \alpha = \frac{1}{Cn},
\end{equation}
where $n$ is the number of token instances used to fit the corresponding one-versus-all head. Optimization uses five epochs, the \texttt{optimal} learning-rate schedule, shuffling at each pass, no early-stopping tolerance, and averaged SGD iterates. We use random seed $0$ throughout. No explicit class weighting is applied.

We hold these \texttt{sAwMIL} hyperparameters fixed across LLMs, datasets, and probability estimators. The individual-statement and Direct Conditional probes contain three one-versus-all heads, whereas Joint probes contain nine. Direct Conditional and Joint probes are fit separately on their respective combined-statement training sets rather than reusing the individual-statement probe, but use the same \texttt{sAwMIL} training and calibration procedure and the model--dataset-specific layer selected from the sweep described below.

\begin{figure}[ht!]
\centering
\includegraphics[width=0.95\textwidth]{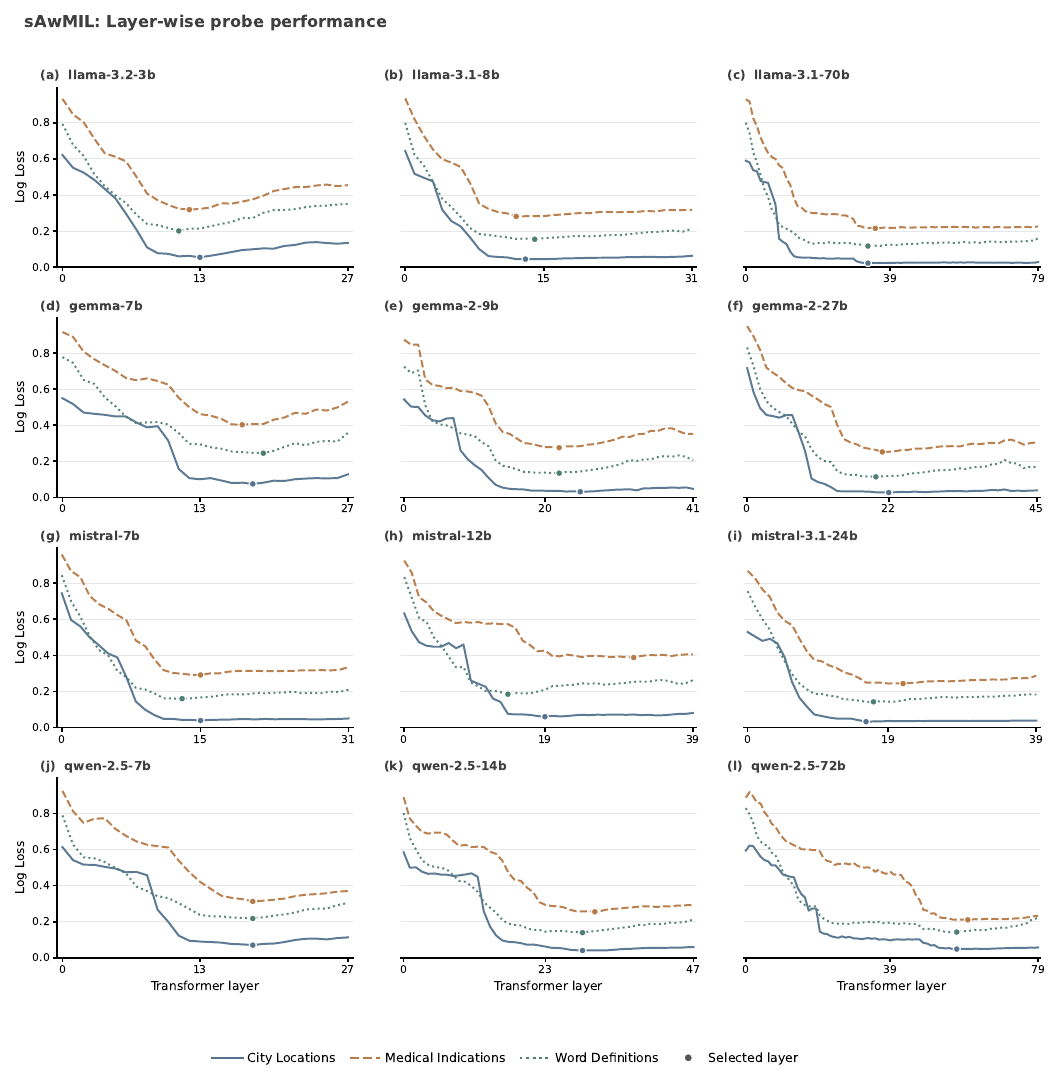}
\caption{
\textbf{\texttt{sAwMIL} layer-selection across layers.}
Calibration-set multiclass log loss for the three-class individual-statement probe across every decoder layer of each instruction-tuned LLM. Panels \textbf{(a--l)} correspond to the $12$ instruction-tuned LLMs. Within each panel, curves show City Locations (blue, solid), Medical Indications (orange, dashed), and Word Definitions (green, dotted), with circular markers denoting the minimum-log-loss layer selected for each dataset. These minimum-log-loss layers are listed in Table~\ref{tab:si:selected_layers}.
}
\label{si:fig:sweep_sawmil}
\end{figure}

\subsection{Layer selection}
\label{sec:si:layer_selection}

\begin{table*}[tb]
\centering
\begin{tabular}{lccc}
\toprule
\textbf{Model}
&
\textbf{City Locations}
&
\textbf{Medical Indications}
&
\textbf{Word Definitions}
\\
\midrule
llama-3.2-3b & 9 & 11 & 13 \\
llama-3.1-8b & 12 & 19 & 12 \\
llama-3.1-70b & 34 & 34 & 17 \\
\addlinespace[2pt]
gemma-7b & 19 & 16 & 17 \\
gemma-2-9b & 23 & 20 & 21 \\
gemma-2-27b & 15 & 19 & 18 \\
\addlinespace[2pt]
mistral-7b & 13 & 14 & 14 \\
mistral-12b & 23 & 20 & 17 \\
mistral-3.1-24b & 15 & 19 & 15 \\
\addlinespace[2pt]
qwen-2.5-7b & 17 & 16 & 18 \\
qwen-2.5-14b & 27 & 26 & 24 \\
qwen-2.5-72b & 57 & 53 & 49 \\
\bottomrule
\end{tabular}%
\caption{
\textbf{Selected \texttt{sAwMIL} layers.}
Zero-indexed layers selected independently for each instruction-tuned LLM and dataset by minimizing three-class calibration log loss. The selected layer is subsequently used for the corresponding individual-statement, Direct Conditional, and Joint \texttt{sAwMIL} probes.
}
\label{tab:si:selected_layers}
\end{table*}

For each LLM and dataset, we independently sweep all layers using the individual-statement three-class $\{\texttt{True},\texttt{False}, \allowbreak \texttt{Neither}\}$ prediction task. At each layer $\ell$, we fit \texttt{sAwMIL} on $\mathcal{D}_{\mathrm{train}}$, fit its probability-calibration mapping on $\mathcal{D}_{\mathrm{cal}}$, and evaluate the resulting calibrated probabilities on $\mathcal{D}_{\mathrm{cal}}$. We select the layer with the lowest multiclass calibration log loss,
\begin{equation}
    \ell^\star
    =
    \arg\min_{\ell}
    \left[
        -\frac{1}{|\mathcal{D}_{\mathrm{cal}}|}
        \sum_{(s_i,y_i)\in\mathcal{D}_{\mathrm{cal}}}
        \log \pi^{(\ell)}_{y_i}(s_i)
    \right].
\end{equation}
Layer selection is therefore performed separately for every LLM--dataset combination rather than assuming that veracity is most recoverable at a fixed absolute or relative transformer depth. Figure~\ref{si:fig:sweep_sawmil} shows the complete \texttt{sAwMIL} layer sweeps, and Table~\ref{tab:si:selected_layers} reports the resulting selected layers. Because the calibration mapping and the layer-selection loss are both computed using $\mathcal{D}_{\mathrm{cal}}$, this sweep is used as a model-selection diagnostic rather than as an independent estimate of held-out calibration performance. The test partition is not used to fit the probe, fit the calibration mapping, or select the layer.

%%%%%%%%%%%%%%%%%%%%%%%%%%%%%%%%%%%%%%%%%%%%
%% Appendices: Conditional Prob. details  %%
%%%%%%%%%%%%%%%%%%%%%%%%%%%%%%%%%%%%%%%%%%%%

\FloatBarrier

\begin{table*}[ht!]
\centering
\small
\setlength{\tabcolsep}{3.5pt}
\renewcommand{\arraystretch}{1.05}
\resizebox{0.9\textwidth}{!}{%
\begin{tabular}{llrrrrrrr}
\toprule
& & \multicolumn{3}{c}{\textbf{Belief sets}} 
& \multicolumn{4}{c}{\textbf{Conditional-pair coverage}} \\
\textbf{Dataset} 
& \textbf{Model} 
& $|\mathcal{B}_{\mathcal M}|$ 
& $|\mathcal{X}_{\mathcal M}|$ 
& $t_{\mathcal M}$ 
& \textbf{Candidate} 
& \textbf{Direct} 
& \textbf{Joint} 
& \textbf{Joint excl.} \\
\midrule

City Locations
& llama-3.2-3b & $632$ & $1{,}217$ & $0.506$ & $768{,}512$ & $768{,}512$ & $768{,}512$ & $0\;(0.0000\%)$ \\
& llama-3.1-8b & $629$ & $1{,}213$ & $0.543$ & $762{,}348$ & $762{,}348$ & $762{,}348$ & $0\;(0.0000\%)$ \\
& llama-3.1-70b & $628$ & $1{,}212$ & $0.478$ & $760{,}508$ & $760{,}508$ & $760{,}471$ & $37\;(0.0049\%)$ \\
& gemma-7b & $615$ & $1{,}199$ & $0.513$ & $736{,}770$ & $736{,}770$ & $736{,}770$ & $0\;(0.0000\%)$ \\
& gemma-2-9b & $626$ & $1{,}211$ & $0.537$ & $757{,}460$ & $757{,}460$ & $756{,}727$ & $733\;(0.0968\%)$ \\
& gemma-2-27b & $635$ & $1{,}219$ & $0.509$ & $773{,}430$ & $773{,}430$ & $772{,}198$ & $1{,}232\;(0.1593\%)$ \\
& mistral-7b & $630$ & $1{,}215$ & $0.510$ & $764{,}820$ & $764{,}820$ & $764{,}544$ & $276\;(0.0361\%)$ \\
& mistral-12b & $636$ & $1{,}224$ & $0.456$ & $777{,}828$ & $777{,}828$ & $777{,}828$ & $0\;(0.0000\%)$ \\
& mistral-3.1-24b & $637$ & $1{,}222$ & $0.557$ & $777{,}777$ & $777{,}777$ & $777{,}709$ & $68\;(0.0087\%)$ \\
& qwen-2.5-7b & $644$ & $1{,}225$ & $0.503$ & $788{,}256$ & $788{,}256$ & $788{,}256$ & $0\;(0.0000\%)$ \\
& qwen-2.5-14b & $629$ & $1{,}212$ & $0.506$ & $761{,}719$ & $761{,}719$ & $760{,}373$ & $1{,}346\;(0.1767\%)$ \\
& qwen-2.5-72b & $627$ & $1{,}213$ & $0.503$ & $759{,}924$ & $759{,}924$ & $756{,}321$ & $3{,}603\;(0.4741\%)$ \\

\addlinespace[2.5pt]

Medical Indications
& llama-3.2-3b & $684$ & $1{,}018$ & $0.501$ & $695{,}628$ & $695{,}628$ & $695{,}628$ & $0\;(0.0000\%)$ \\
& llama-3.1-8b & $694$ & $1{,}035$ & $0.466$ & $717{,}596$ & $717{,}596$ & $717{,}596$ & $0\;(0.0000\%)$ \\
& llama-3.1-70b & $683$ & $1{,}025$ & $0.500$ & $699{,}392$ & $699{,}392$ & $699{,}392$ & $0\;(0.0000\%)$ \\
& gemma-7b & $715$ & $1{,}045$ & $0.368$ & $746{,}460$ & $746{,}460$ & $746{,}460$ & $0\;(0.0000\%)$ \\
& gemma-2-9b & $690$ & $1{,}031$ & $0.505$ & $710{,}700$ & $710{,}700$ & $710{,}700$ & $0\;(0.0000\%)$ \\
& gemma-2-27b & $685$ & $1{,}022$ & $0.457$ & $699{,}385$ & $699{,}385$ & $699{,}385$ & $0\;(0.0000\%)$ \\
& mistral-7b & $688$ & $1{,}023$ & $0.357$ & $703{,}136$ & $703{,}136$ & $703{,}136$ & $0\;(0.0000\%)$ \\
& mistral-12b & $686$ & $1{,}023$ & $0.371$ & $701{,}092$ & $701{,}092$ & $701{,}092$ & $0\;(0.0000\%)$ \\
& mistral-3.1-24b & $690$ & $1{,}031$ & $0.500$ & $710{,}700$ & $710{,}700$ & $710{,}623$ & $77\;(0.0108\%)$ \\
& qwen-2.5-7b & $707$ & $1{,}053$ & $0.435$ & $743{,}764$ & $743{,}764$ & $743{,}760$ & $4\;(0.0005\%)$ \\
& qwen-2.5-14b & $710$ & $1{,}050$ & $0.500$ & $744{,}790$ & $744{,}790$ & $744{,}779$ & $11\;(0.0015\%)$ \\
& qwen-2.5-72b & $685$ & $1{,}027$ & $0.500$ & $702{,}810$ & $702{,}810$ & $702{,}379$ & $431\;(0.0613\%)$ \\

\addlinespace[2.5pt]

Word Definitions
& llama-3.2-3b & $610$ & $1{,}584$ & $0.395$ & $965{,}630$ & $965{,}630$ & $965{,}630$ & $0\;(0.0000\%)$ \\
& llama-3.1-8b & $641$ & $1{,}606$ & $0.447$ & $1{,}028{,}805$ & $1{,}028{,}805$ & $1{,}028{,}805$ & $0\;(0.0000\%)$ \\
& llama-3.1-70b & $638$ & $1{,}610$ & $0.447$ & $1{,}026{,}542$ & $1{,}026{,}542$ & $1{,}026{,}542$ & $0\;(0.0000\%)$ \\
& gemma-7b & $638$ & $1{,}623$ & $0.398$ & $1{,}034{,}836$ & $1{,}034{,}836$ & $1{,}034{,}836$ & $0\;(0.0000\%)$ \\
& gemma-2-9b & $606$ & $1{,}580$ & $0.469$ & $956{,}874$ & $956{,}874$ & $956{,}874$ & $0\;(0.0000\%)$ \\
& gemma-2-27b & $618$ & $1{,}584$ & $0.441$ & $978{,}294$ & $978{,}294$ & $978{,}294$ & $0\;(0.0000\%)$ \\
& mistral-7b & $639$ & $1{,}612$ & $0.437$ & $1{,}029{,}429$ & $1{,}029{,}429$ & $1{,}029{,}429$ & $0\;(0.0000\%)$ \\
& mistral-12b & $617$ & $1{,}589$ & $0.377$ & $979{,}796$ & $979{,}796$ & $979{,}796$ & $0\;(0.0000\%)$ \\
& mistral-3.1-24b & $619$ & $1{,}592$ & $0.461$ & $984{,}829$ & $984{,}829$ & $984{,}829$ & $0\;(0.0000\%)$ \\
& qwen-2.5-7b & $627$ & $1{,}616$ & $0.409$ & $1{,}012{,}605$ & $1{,}012{,}605$ & $1{,}012{,}605$ & $0\;(0.0000\%)$ \\
& qwen-2.5-14b & $620$ & $1{,}594$ & $0.425$ & $987{,}660$ & $987{,}660$ & $987{,}660$ & $0\;(0.0000\%)$ \\
& qwen-2.5-72b & $617$ & $1{,}593$ & $0.469$ & $982{,}264$ & $982{,}264$ & $982{,}264$ & $0\;(0.0000\%)$ \\

\bottomrule
\end{tabular}%
}

\caption{
\textbf{Belief-set and conditional-pair coverage using \texttt{sAwMIL}.}
For each LLM and domain, $|\mathcal{B}_{\mathcal M}|$ and $|\mathcal{X}_{\mathcal M}|$ denote the numbers of beliefs and non-disbeliefs, respectively, and $t_{\mathcal M}$ is the empirical belief threshold. Candidate is the number $|\mathcal{B}_{\mathcal M}|(|\mathcal{X}_{\mathcal M}|-1)$ of constructed $(P,x)$ pairs after excluding self-pairs. Direct and Joint report the numbers of valid conditional-probability estimates. Joint excl. reports the number and percentage of candidate pairs excluded because the Joint-to-Conditional denominator is at most $10^{-12}$.
}
\label{tab:si:conditional_coverage_sawmil}
\end{table*}

\section{Construction and coverage of conditional probability estimates}
\label{sec:si:conditional_details}

\subsection{Belief sets and conditional-pair coverage}
\label{sec:si:belief_set_counts}

Table~\ref{tab:si:conditional_coverage_sawmil} reports the empirical belief sets, non-disbelief sets, belief thresholds, and conditional-pair coverage for the primary \texttt{sAwMIL} probe. For each model and dataset, the candidate set contains every pair $(P,x)$ with $P\in\mathcal{B}_{\mathcal M}$ and $x\in\mathcal{X}_{\mathcal M}\setminus\{P\}$, yielding $|\mathcal{B}_{\mathcal M}|\cdot(|\mathcal{X}_{\mathcal M}|-1)$ total candidate pairs. We report the number of candidate pairs successfully assigned a conditional-probability estimate by the Direct Conditional and Joint-to-Conditional estimators, together with the number of Joint-to-Conditional pairs excluded due to an undefined denominator in Eq.~\ref{eq:joint_conditional}.

The Direct Conditional estimator yields a valid estimate for every candidate pair across all $36$ model--dataset combinations. Joint-to-Conditional coverage is also nearly complete: only $7{,}818$ of $29{,}732{,}369$ candidate pairs ($0.026\%$) are excluded because of a numerically zero denominator. No Joint-to-Conditional pairs are excluded for Word Definitions, and the largest exclusion rate for any individual model--dataset combination is $0.4741\%$.

\subsection{Joint-to-Conditional derivation}
\label{sec:si:joint_conditional_derivation}

We derive the Joint-to-Conditional estimator in Equation~\eqref{eq:joint_conditional} from the nine-state joint distribution $\boldsymbol{\pi}_\mathcal{M}^\mathrm{joint}(P,x)$. Recall that $\pi_{ab}^\mathrm{joint}(P,x)$ denotes the probability assigned to the joint state in which the antecedent $x$ has state $a\in\{T,F,N\}$ and the consequent $P$ has state $b\in\{T,F,N\}$.

Under Cooper--Cantwell semantics, the conditional $P\mid x$ takes the truth value of $P$ when $x$ is \texttt{True} or \texttt{Neither}, and is \texttt{Neither} when $x$ is \texttt{False}. Collapsing the nine joint states according to this truth table therefore gives the induced trivalent conditional distribution
\begin{align}
    \pi_T^{\mathrm{joint},C}(P,x)
    &=
    \pi_{TT}^\mathrm{joint}(P,x)
    +
    \pi_{NT}^\mathrm{joint}(P,x),
    \\
    \pi_F^{\mathrm{joint},C}(P,x)
    &=
    \pi_{TF}^\mathrm{joint}(P,x)
    +
    \pi_{NF}^\mathrm{joint}(P,x),
    \\
    \pi_N^{\mathrm{joint},C}(P,x)
    &=
    \pi_{TN}^\mathrm{joint}(P,x)
    +
    \pi_{NN}^\mathrm{joint}(P,x)
    +
    \pi_{FT}^\mathrm{joint}(P,x)
    +
    \pi_{FF}^\mathrm{joint}(P,x)
    +
    \pi_{FN}^\mathrm{joint}(P,x).
\end{align}
Here, the superscript $C$ denotes the conditional distribution induced from the joint probe.

Cantwell's non-bivalent probability rule assigns a sentence its probability of being \texttt{True} conditional on its having a determinate truth value~\cite{cantwell2006laws}. Applying this normalization to the induced conditional distribution yields
\begin{align}
    \widehat{\Pr}_\mathcal{M}^{\mathrm{joint}}(P\mid x)
    &=
    \frac{
    \pi_T^{\mathrm{joint},C}(P,x)
    }{
    \pi_T^{\mathrm{joint},C}(P,x)
    +
    \pi_F^{\mathrm{joint},C}(P,x)
    }
    \\
    &=
    \frac{
    \pi_{TT}^\mathrm{joint}(P,x)
    +
    \pi_{NT}^\mathrm{joint}(P,x)
    }{
    \pi_{TT}^\mathrm{joint}(P,x)
    +
    \pi_{TF}^\mathrm{joint}(P,x)
    +
    \pi_{NT}^\mathrm{joint}(P,x)
    +
    \pi_{NF}^\mathrm{joint}(P,x)
    },
\end{align}
recovering Equation~\eqref{eq:joint_conditional}. The numerator contains exactly the joint states that induce a \texttt{True} conditional, while the denominator contains all states that induce either a \texttt{True} or \texttt{False} conditional. The estimator is therefore undefined when the conditional is assigned probability one of being \texttt{Neither}; as described in Section~\ref{sec:methods:joint}, we treat denominators less than or equal to $10^{-12}$ as numerically zero and exclude the corresponding pairs.

This normalization also clarifies the distinction between the two conditional-probability estimators. The Direct Conditional estimator uses the raw \texttt{True}-state mass $\pi_T^\mathrm{direct}(P,x)$, whereas Joint-to-Conditional normalizes the induced \texttt{True} mass over the \texttt{True} and \texttt{False} conditional states.

%%%%%%%%%%%%%%%%%%%%%%%%%%%%%%%%%%%%%%%%%%%%%
%% Appendices: validating belief stability %%
%%%%%%%%%%%%%%%%%%%%%%%%%%%%%%%%%%%%%%%%%%%%%

\FloatBarrier

\begin{figure}[ht!]
\centering
\includegraphics[width=\textwidth]{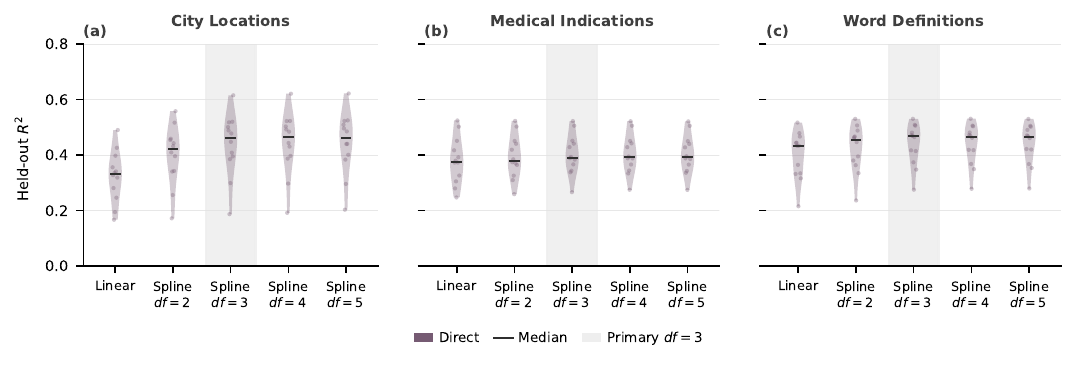}
\caption{
\textbf{Robustness of the natural cubic spline.}
Violin plots show the distribution across the $12$ instruction-tuned LLMs of held-out $R^2$ for predicting graded stability from individual belief probability using a linear baseline and natural cubic splines with $2$--$5$ degrees of freedom for \textbf{(a)} City Locations, \textbf{(b)} Medical Indications, and \textbf{(c)} Word Definitions. Horizontal black lines denote the median across models, and the shaded column marks the primary specification with $3$ degrees of freedom. Predictive performance is similar across the $3$--$5$ degree-of-freedom specifications, indicating that the main conclusion is not sensitive to the precise spline flexibility.
}
\label{si:fig:spline_robustness}
\end{figure}

\section{Supplementary validation of graded belief stability}
\label{sec:si:stability_validation}

\subsection{Robustness of the stability--belief-probability relationship}
\label{sec:si:stability_probability_robustness}

Our primary analysis models graded stability as a function of individual belief probability using a natural cubic spline with three degrees of freedom. To evaluate sensitivity to this choice, we repeat the analysis using a linear baseline and natural cubic splines with $2$, $3$, $4$, and $5$ degrees of freedom.

We use five-fold cross-validation to evaluate the cubic spline models. Propositions are randomly permuted using a deterministic seed derived from base seed $0$ and divided into five approximately equal folds. The same fold assignment is reused across spline specifications. Within each fold, we construct a centered natural cubic spline basis from the training propositions with fixed boundaries at $0$ and $1$, estimate its coefficients by ordinary least squares, and apply the resulting basis to the held-out propositions. The linear baseline is fit analogously using an intercept and linear term in $\pi_T(P)$. We apply the same inclusion criteria as in the primary analysis, requiring at least $50$ valid proposition-level stability scores and sufficient distinct values of $\pi_T(P)$ to fit the requested spline.

For the cross-model residual-agreement analysis in Section~\ref{sec:methods:stability_credence}, we use the out-of-fold residuals from the primary $df=3$ specification and retain model pairs sharing at least $50$ believed propositions. We generate $1{,}000$ permutation samples by independently shuffling residuals across proposition identities within each model while preserving model overlap, recompute the median pairwise Spearman correlation for each permutation, and calculate the one-sided Monte Carlo $p$-value using the standard $+1$ correction. Permutations use deterministic seeds derived from the same base seed $0$.

Across all three domains, the primary $df=3$ specification performs similarly to the more flexible $df=4$ and $df=5$ alternatives (Fig.~\ref{si:fig:spline_robustness}). The linear baseline and $df=2$ spline tend to yield lower held-out $R^2$, particularly for City Locations, indicating that some nonlinearity is useful. Increasing flexibility beyond the primary specification, however, produces little systematic improvement. The conclusion that individual belief probability predicts a meaningful but incomplete fraction of graded stability therefore does not depend on the exact spline complexity.

\subsection{Additional probabilistic-coherence analyses}
\label{sec:si:coherence}

\subsubsection{Exact pairwise coherence conditions}
\label{sec:si:coherence_proof}

We show that the four constraints in Equation~\eqref{eq:cck_constraints} are both necessary and sufficient for the three measured distributions associated with a single $(P,x)$ pair to admit a common trivalent joint representation. For compactness, we write
\begin{equation*}
    p_b = \pi_b(P), \qquad
    a_b = \pi_b(x), \qquad
    c_b = \pi_b^\mathrm{direct}(P,x),
    \qquad b\in\{T,F,N\}.
\end{equation*}
Let $J_{ab}$ denote the probability assigned by a latent joint distribution to the state in which $x$ has value $a$ and $P$ has value $b$, with the first coordinate corresponding to the antecedent. The row marginals of $J$ must equal $\boldsymbol{\pi}_\mathcal{M}(x)$ and its column marginals must equal $\boldsymbol{\pi}_\mathcal{M}(P)$.

Under the Cooper--Cantwell truth table, such a joint distribution induces the conditional distribution
\begin{align}
    c_T &= J_{TT} + J_{NT}, \\
    c_F &= J_{TF} + J_{NF}, \\
    c_N &= J_{TN} + J_{NN} + J_{FT} + J_{FF} + J_{FN}.
\end{align}

\paragraph{Necessity.}
Suppose first that a representing joint distribution $J$ exists. Combining its marginals with the induced conditional distribution gives
\begin{align}
    p_T-c_T &= J_{FT} \geq 0, \\
    p_F-c_F &= J_{FF} \geq 0, \\
    c_N-a_F &= J_{TN}+J_{NN} \geq 0, \\
    p_N+a_F-c_N &= J_{FN} \geq 0.
\end{align}
It follows that
\begin{equation*}
    c_T \leq p_T,
    \qquad
    c_F \leq p_F,
    \qquad
    a_F \leq c_N,
    \qquad
    c_N \leq p_N+a_F,
\end{equation*}
which are exactly the four constraints in Equation~\eqref{eq:cck_constraints}.

\paragraph{Sufficiency.} 
Conversely, suppose the four constraints hold. Define the vectors
\begin{align}
    \mathbf{r}
    &=
    \left(
    c_T,\;
    c_F,\;
    c_N-a_F
    \right),
    \\
    \mathbf{f}
    &=
    \left(
    p_T-c_T,\;
    p_F-c_F,\;
    p_N+a_F-c_N
    \right),
\end{align}
with coordinates ordered as $(T,F,N)$. The four coherence constraints imply that every coordinate of $\mathbf{r}$ and $\mathbf{f}$ is nonnegative. Because $\mathbf{p}$ and $\mathbf{c}$ are each normalized probability distributions,
\begin{equation*}
    \sum_{b\in\{T,F,N\}} r_b = 1-a_F,
    \qquad
    \sum_{b\in\{T,F,N\}} f_b = a_F,
    \qquad
    r_b+f_b=p_b.
\end{equation*}
The vector $\mathbf{f}$ can therefore be used as the \texttt{False}-antecedent row of the latent joint distribution, while $\mathbf{r}$ supplies the total mass of the \texttt{True}- and \texttt{Neither}-antecedent rows.

When $a_F<1$, define
\begin{equation}
\label{eq:si:cck_joint_construction}
    J_{Fb}=f_b,
    \qquad
    J_{Tb}=\frac{a_T}{1-a_F}r_b,
    \qquad
    J_{Nb}=\frac{a_N}{1-a_F}r_b,
    \qquad
    b\in\{T,F,N\}.
\end{equation}
All entries are nonnegative. Their row sums are $a_F$, $a_T$, and $a_N$, respectively, and because $a_T+a_N=1-a_F$, the column sum for each $b$ is
\begin{equation*}
    J_{Tb}+J_{Nb}+J_{Fb}=r_b+f_b=p_b.
\end{equation*}
Thus, $J$ has the required marginals. Moreover, the induced conditional has \texttt{True} and \texttt{False} masses $r_T=c_T$ and $r_F=c_F$, while its \texttt{Neither} mass is
\begin{equation*}
    a_F+r_N
    =
    a_F+(c_N-a_F)
    =
    c_N.
\end{equation*}
Hence, $J$ reproduces all three measured distributions.

The remaining case is $a_F=1$. Normalization then implies $a_T=a_N=0$. Because $a_F\leq c_N$ and $\mathbf{c}$ is normalized, $c_N=1$ and $c_T=c_F=0$. Setting
\begin{equation*}
    J_{Fb}=p_b,
    \qquad
    J_{Tb}=J_{Nb}=0
\end{equation*}
for every $b\in\{T,F,N\}$ therefore provides a valid joint distribution with the required marginals and an everywhere-\texttt{Neither} conditional.

The four inequalities in Equation~\eqref{eq:cck_constraints} are therefore necessary and sufficient for exact pairwise compatibility, provided that the three measured triples are themselves normalized nonnegative probability distributions.

\paragraph{Projection onto the coherent set.}
The pairwise coherent set $\mathcal{C}$ used to compute $d_\mathrm{CCK}$ is the intersection of three probability simplexes with the four closed linear half-spaces in Equation~\eqref{eq:cck_constraints}. It is therefore a nonempty, compact, and convex subset of $\mathbb{R}^9$. Consequently, every measured vector $\mathbf{v}_{P,x}$ has a unique Euclidean projection $\Pi_\mathcal{C}(\mathbf{v}_{P,x})$ onto this set, and Equation~\eqref{eq:cck_distance} is well defined.

\paragraph{Scope of the coherence analysis.}
These conditions characterize compatibility for a single $(P,x)$ pair. Testing them separately across pairs does not establish that all measured probabilities for an LLM can be generated by one global joint distribution over its entire belief system. Such simultaneous compatibility is a stronger global feasibility problem requiring a shared probability distribution over admissible trivalent valuations of all propositions. Our coherence analysis tests the pairwise condition only.

\subsubsection{Exact coherence}
\label{sec:si:coherence_results}

\begin{table*}[tb]
\centering
\small
\setlength{\tabcolsep}{3.2pt}
\renewcommand{\arraystretch}{1.05}
\resizebox{\textwidth}{!}{%
\begin{tabular}{llccccc}
\toprule
\textbf{Dataset}
& \textbf{Model}
& \textbf{Exact}
& $\boldsymbol{\pi_T^{\mathrm{direct}}(P,x) > \pi_T(P)}$
& $\boldsymbol{\pi_F^{\mathrm{direct}}(P,x) > \pi_F(P)}$
& $\boldsymbol{\pi_F(x) > \pi_N^{\mathrm{direct}}(P,x)}$
& $\boldsymbol{\pi_N^{\mathrm{direct}}(P,x) > \pi_N(P)+\pi_F(x)}$ \\
\midrule

City Locations
& llama-3.2-3b & $0.000\%$ & $2.7\%\;(0.095)$ & $90.4\%\;(0.154)$ & $3.4\%\;(0.122)$ & $96.6\%\;(0.207)$ \\
& llama-3.1-8b & $0.000\%$ & $5.2\%\;(0.054)$ & $88.0\%\;(0.051)$ & $5.8\%\;(0.047)$ & $94.2\%\;(0.080)$ \\
& llama-3.1-70b & $<0.001\%$ & $7.9\%\;(0.031)$ & $80.5\%\;(0.035)$ & $11.4\%\;(0.026)$ & $88.5\%\;(0.047)$ \\
& gemma-7b & $<0.001\%$ & $3.8\%\;(0.107)$ & $92.4\%\;(0.159)$ & $7.9\%\;(0.091)$ & $92.1\%\;(0.102)$ \\
& gemma-2-9b & $0.000\%$ & $3.1\%\;(0.050)$ & $93.9\%\;(0.087)$ & $11.0\%\;(0.036)$ & $89.0\%\;(0.046)$ \\
& gemma-2-27b & $0.000\%$ & $2.4\%\;(0.067)$ & $93.7\%\;(0.104)$ & $5.9\%\;(0.073)$ & $94.1\%\;(0.109)$ \\
& mistral-7b & $0.000\%$ & $3.5\%\;(0.097)$ & $91.7\%\;(0.077)$ & $8.8\%\;(0.053)$ & $91.2\%\;(0.093)$ \\
& mistral-12b & $0.004\%$ & $9.0\%\;(0.055)$ & $62.5\%\;(0.082)$ & $11.1\%\;(0.075)$ & $88.7\%\;(0.112)$ \\
& mistral-3.1-24b & $0.000\%$ & $8.8\%\;(0.027)$ & $86.0\%\;(0.074)$ & $21.2\%\;(0.022)$ & $78.8\%\;(0.068)$ \\
& qwen-2.5-7b & $<0.001\%$ & $4.6\%\;(0.052)$ & $88.0\%\;(0.125)$ & $16.6\%\;(0.052)$ & $83.4\%\;(0.112)$ \\
& qwen-2.5-14b & $0.000\%$ & $8.8\%\;(0.032)$ & $78.8\%\;(0.066)$ & $18.5\%\;(0.037)$ & $81.5\%\;(0.077)$ \\
& qwen-2.5-72b & $0.000\%$ & $8.4\%\;(0.031)$ & $74.4\%\;(0.056)$ & $15.0\%\;(0.039)$ & $85.0\%\;(0.077)$ \\

\addlinespace[2.5pt]

Medical Indications
& llama-3.2-3b & $0.000\%$ & $3.6\%\;(0.082)$ & $63.8\%\;(0.153)$ & $12.1\%\;(0.138)$ & $87.9\%\;(0.276)$ \\
& llama-3.1-8b & $0.008\%$ & $5.2\%\;(0.092)$ & $59.7\%\;(0.111)$ & $13.2\%\;(0.132)$ & $86.7\%\;(0.252)$ \\
& llama-3.1-70b & $0.000\%$ & $7.4\%\;(0.088)$ & $65.8\%\;(0.111)$ & $13.6\%\;(0.127)$ & $86.4\%\;(0.179)$ \\
& gemma-7b & $0.097\%$ & $2.1\%\;(0.095)$ & $60.6\%\;(0.152)$ & $20.8\%\;(0.132)$ & $79.0\%\;(0.250)$ \\
& gemma-2-9b & $0.000\%$ & $8.4\%\;(0.094)$ & $72.1\%\;(0.119)$ & $21.8\%\;(0.129)$ & $78.1\%\;(0.148)$ \\
& gemma-2-27b & $0.015\%$ & $9.4\%\;(0.103)$ & $68.9\%\;(0.100)$ & $20.6\%\;(0.134)$ & $79.3\%\;(0.150)$ \\
& mistral-7b & $<0.001\%$ & $7.3\%\;(0.085)$ & $63.8\%\;(0.090)$ & $21.6\%\;(0.134)$ & $78.4\%\;(0.183)$ \\
& mistral-12b & $0.200\%$ & $13.4\%\;(0.099)$ & $45.4\%\;(0.107)$ & $31.7\%\;(0.132)$ & $67.9\%\;(0.184)$ \\
& mistral-3.1-24b & $0.015\%$ & $11.9\%\;(0.096)$ & $58.2\%\;(0.054)$ & $19.8\%\;(0.134)$ & $80.1\%\;(0.128)$ \\
& qwen-2.5-7b & $0.035\%$ & $3.5\%\;(0.087)$ & $75.8\%\;(0.184)$ & $23.4\%\;(0.132)$ & $76.2\%\;(0.204)$ \\
& qwen-2.5-14b & $0.000\%$ & $6.3\%\;(0.088)$ & $72.9\%\;(0.125)$ & $25.3\%\;(0.127)$ & $74.7\%\;(0.178)$ \\
& qwen-2.5-72b & $0.000\%$ & $7.1\%\;(0.086)$ & $74.7\%\;(0.079)$ & $22.6\%\;(0.114)$ & $77.4\%\;(0.144)$ \\

\addlinespace[2.5pt]

Word Definitions
& llama-3.2-3b & $0.073\%$ & $6.6\%\;(0.091)$ & $64.2\%\;(0.100)$ & $5.7\%\;(0.106)$ & $93.9\%\;(0.223)$ \\
& llama-3.1-8b & $0.020\%$ & $4.6\%\;(0.086)$ & $77.5\%\;(0.103)$ & $6.8\%\;(0.125)$ & $93.0\%\;(0.177)$ \\
& llama-3.1-70b & $0.018\%$ & $6.5\%\;(0.085)$ & $73.7\%\;(0.103)$ & $4.7\%\;(0.121)$ & $95.1\%\;(0.185)$ \\
& gemma-7b & $0.206\%$ & $2.7\%\;(0.088)$ & $64.1\%\;(0.140)$ & $8.1\%\;(0.118)$ & $91.4\%\;(0.259)$ \\
& gemma-2-9b & $0.012\%$ & $8.2\%\;(0.109)$ & $82.3\%\;(0.135)$ & $10.1\%\;(0.123)$ & $88.8\%\;(0.120)$ \\
& gemma-2-27b & $0.003\%$ & $3.6\%\;(0.109)$ & $87.2\%\;(0.145)$ & $5.1\%\;(0.110)$ & $94.7\%\;(0.169)$ \\
& mistral-7b & $0.030\%$ & $4.7\%\;(0.094)$ & $72.5\%\;(0.097)$ & $6.4\%\;(0.124)$ & $93.2\%\;(0.207)$ \\
& mistral-12b & $0.054\%$ & $2.7\%\;(0.092)$ & $73.1\%\;(0.132)$ & $5.6\%\;(0.126)$ & $94.3\%\;(0.323)$ \\
& mistral-3.1-24b & $0.026\%$ & $7.0\%\;(0.101)$ & $76.0\%\;(0.103)$ & $7.8\%\;(0.125)$ & $91.8\%\;(0.173)$ \\
& qwen-2.5-7b & $0.080\%$ & $7.7\%\;(0.108)$ & $76.2\%\;(0.135)$ & $10.9\%\;(0.120)$ & $87.6\%\;(0.189)$ \\
& qwen-2.5-14b & $0.008\%$ & $6.2\%\;(0.093)$ & $79.7\%\;(0.132)$ & $9.0\%\;(0.113)$ & $90.7\%\;(0.162)$ \\
& qwen-2.5-72b & $0.021\%$ & $3.8\%\;(0.082)$ & $82.0\%\;(0.164)$ & $8.2\%\;(0.117)$ & $91.6\%\;(0.189)$ \\

\bottomrule
\end{tabular}%
}

\caption{
\textbf{Exact CCK coherence and constraint violations for Direct Conditional estimates using \texttt{sAwMIL}.}
For each LLM and domain, we report the percentage of $(P,x)$ pairs satisfying all four CCK constraints simultaneously within tolerance $10^{-8}$. Here, $0.000\%$ signifies exactly zero fully coherent pairs, while $<0.001\%$ means a small number of exactly coherent pairs exists. Each remaining column reports the percentage of pairs violating each constraint, with the mean excess across violating pairs shown in parentheses. Exact coherence is rare across all models and domains, with no model--domain combination exceeding $0.206\%$. Violations occur most frequently for the constraints on conditional \texttt{False} and \texttt{Neither} probability mass.
}
\label{tab:si:exact_coherence_sawmil}
\end{table*}

Table~\ref{tab:si:exact_coherence_sawmil} reports exact CCK-coherence rates and individual constraint violations for the primary Direct Conditional estimates obtained using \texttt{sAwMIL}. As described in Section~\ref{sec:methods:probabilistic_coherence}, a $(P,x)$ pair is classified as exactly coherent when all four CCK inequalities are satisfied within the numerical tolerance of $10^{-8}$. For each inequality, we additionally report how frequently it is violated and the mean signed excess among the pairs that violate it.

Exact coherence is rare for every model and domain (Tab.~\ref{tab:si:exact_coherence_sawmil}), with the largest observed rate equal to $0.206\%$. The two most frequently violated conditions are $\pi_F^{\mathrm{direct}}(P,x)\leq\pi_F(P)$ and $\pi_N^{\mathrm{direct}}(P,x)\leq\pi_N(P)+\pi_F(x)$, whereas violations of the remaining two inequalities occur substantially less often. Because exact coherence requires all four constraints to hold simultaneously, these frequent violations drive the near-zero exact-coherence rates reported in the main text and drive the distance-to-coherence analysis in Section~\ref{sec:methods:probabilistic_coherence}.

One likely contributor to this pattern is the scale of the corresponding individual-statement probabilities. Because $P$ is a probe-defined belief, $\pi_F(P)$ and $\pi_N(P)$ are often small. Similarly, because $\mathcal{X}_\mathcal{M}$ contains non-disbeliefs, its members $x$ often have small $\pi_F(x)$. Even modest conditional probability mass assigned to \texttt{False} or \texttt{Neither} can therefore violate the second or fourth constraint. The fourth constraint may additionally be sensitive to the Direct Conditional representation itself: in ``Given $x$, $P$,'' the conditioning proposition precedes $P$ in the autoregressive sequence, allowing the representation of the conditional to incorporate $x$ and potentially shift probability mass toward \texttt{Neither} relative to $P$ in isolation.

%%%%%%%%%%%%%%%%%%%%%%%%%%%%%%%%%%%%%%%%%%%%%
%%    Appendices: Behavioral challenge     %%
%%%%%%%%%%%%%%%%%%%%%%%%%%%%%%%%%%%%%%%%%%%%%

%\FloatBarrier

\begin{table*}[ht!]
\centering
\small
\setlength{\tabcolsep}{4.5pt}
\renewcommand{\arraystretch}{1.05}
\begin{tabular}{llccc}
\toprule
\textbf{Dataset} & \textbf{Model} & \textbf{$n$} & 
\textbf{$|\Delta \pi_T|$} & \textbf{$|\Delta \gamma|$} \\
\midrule

City Locations & llama-3.2-3b & $315$ & $0.00435$ $(0.01055)$ & $0.235$ $(0.173)$ \\
 & llama-3.1-8b & $305$ & $0.00181$ $(0.00556)$ & $0.090$ $(0.122)$ \\
 & llama-3.1-70b & $287$ & $0.00483$ $(0.02893)$ & $0.066$ $(0.116)$ \\
 & gemma-7b & $304$ & $0.00377$ $(0.01042)$ & $0.175$ $(0.175)$ \\
 & gemma-2-9b & $291$ & $0.00370$ $(0.01686)$ & $0.082$ $(0.141)$ \\
 & gemma-2-27b & $316$ & $0.00362$ $(0.01372)$ & $0.133$ $(0.143)$ \\
 & mistral-7b & $303$ & $0.00456$ $(0.01924)$ & $0.113$ $(0.157)$ \\
 & mistral-12b & $278$ & $0.00495$ $(0.01547)$ & $0.146$ $(0.189)$ \\
 & mistral-3.1-24b & $307$ & $0.00362$ $(0.01821)$ & $0.100$ $(0.151)$ \\
 & qwen-2.5-7b & $312$ & $0.00350$ $(0.01001)$ & $0.136$ $(0.157)$ \\
 & qwen-2.5-14b & $290$ & $0.00314$ $(0.01289)$ & $0.103$ $(0.159)$ \\
 & qwen-2.5-72b & $276$ & $0.00282$ $(0.01189)$ & $0.096$ $(0.178)$ \\

\addlinespace[2.5pt]

Medical Indications & llama-3.2-3b & $342$ & $0.00442$ $(0.00530)$ & $0.234$ $(0.185)$ \\
 & llama-3.1-8b & $344$ & $0.00389$ $(0.00518)$ & $0.248$ $(0.204)$ \\
 & llama-3.1-70b & $340$ & $0.00404$ $(0.00763)$ & $0.198$ $(0.181)$ \\
 & gemma-7b & $356$ & $0.00430$ $(0.00639)$ & $0.266$ $(0.217)$ \\
 & gemma-2-9b & $344$ & $0.00381$ $(0.00535)$ & $0.210$ $(0.206)$ \\
 & gemma-2-27b & $338$ & $0.00408$ $(0.00648)$ & $0.182$ $(0.204)$ \\
 & mistral-7b & $329$ & $0.00456$ $(0.00993)$ & $0.179$ $(0.223)$ \\
 & mistral-12b & $339$ & $0.00515$ $(0.00856)$ & $0.200$ $(0.204)$ \\
 & mistral-3.1-24b & $337$ & $0.00379$ $(0.00564)$ & $0.138$ $(0.176)$ \\
 & qwen-2.5-7b & $353$ & $0.00441$ $(0.00579)$ & $0.263$ $(0.206)$ \\
 & qwen-2.5-14b & $354$ & $0.00406$ $(0.00494)$ & $0.232$ $(0.206)$ \\
 & qwen-2.5-72b & $337$ & $0.00401$ $(0.00719)$ & $0.155$ $(0.191)$ \\

\addlinespace[2.5pt]

Word Definitions & llama-3.2-3b & $303$ & $0.00510$ $(0.00960)$ & $0.221$ $(0.233)$ \\
 & llama-3.1-8b & $316$ & $0.00408$ $(0.00787)$ & $0.169$ $(0.191)$ \\
 & llama-3.1-70b & $316$ & $0.00429$ $(0.00772)$ & $0.199$ $(0.205)$ \\
 & gemma-7b & $317$ & $0.00538$ $(0.00766)$ & $0.262$ $(0.230)$ \\
 & gemma-2-9b & $300$ & $0.00488$ $(0.01071)$ & $0.179$ $(0.190)$ \\
 & gemma-2-27b & $307$ & $0.00460$ $(0.01048)$ & $0.204$ $(0.200)$ \\
 & mistral-7b & $315$ & $0.00475$ $(0.00910)$ & $0.201$ $(0.218)$ \\
 & mistral-12b & $308$ & $0.00553$ $(0.01139)$ & $0.300$ $(0.254)$ \\
 & mistral-3.1-24b & $306$ & $0.00462$ $(0.00839)$ & $0.201$ $(0.218)$ \\
 & qwen-2.5-7b & $311$ & $0.00447$ $(0.00834)$ & $0.189$ $(0.199)$ \\
 & qwen-2.5-14b & $308$ & $0.00459$ $(0.00962)$ & $0.169$ $(0.187)$ \\
 & qwen-2.5-72b & $305$ & $0.00474$ $(0.00925)$ & $0.253$ $(0.226)$ \\

\bottomrule
\end{tabular}

\caption{\textbf{Behavioral matching quality.}
For each LLM and domain, $n$ is the number of matched proposition pairs, $|\Delta \pi_T|$ is the absolute difference in belief probability between the two propositions in each pair, and $|\Delta \gamma|$ is their absolute difference in graded stability. Values for $|\Delta \pi_T|$ and $|\Delta \gamma|$ report the mean with standard deviation in parentheses. Across domains and models, matching produces small differences in individual belief probability while retaining substantially larger differences in graded stability.}
\label{tab:si:behavioral_matching_sawmil}
\end{table*}

\section{Additional behavioral results}
\label{sec:si:behavioral}

\subsection{Matched-pairs analysis}
\label{sec:si:matched_pairs}

Table~\ref{tab:si:behavioral_matching_sawmil} reports diagnostics for the probability-matched pairs used in our primary behavioral analysis. As described in Section~\ref{sec:methods:behavioral_resilience}, matching is performed separately within each challenge-sequence block so that both propositions in a pair receive the same conversational challenges in the same order. The reported sample additionally requires both propositions' initial behavioral judgments to agree with their individual-statement probe classifications, as in the primary analysis. For each model, $n$ denotes the number of pairs, while $|\Delta\pi_T|$ and $|\Delta\gamma|$ quantify pairwise differences in individual belief probability and graded stability, respectively.

Across all three domains, matching yields small differences in individual belief probability while retaining substantially larger differences in graded stability. Mean $|\Delta\pi_T|$ is at most $0.00553$ across the reported model--domain combinations, whereas mean $|\Delta\gamma|$ ranges from $0.066$ to $0.300$. Thus, the matched pairs are closely aligned in individual belief probability while preserving meaningful variation in graded stability.

\subsection{Uncertainty in behavioral resilience effects}
\label{sec:si:behavioral_uncertainty}

Figure~\ref{fig:behavioral_resilience} reports $\pm 1$ bootstrap standard error to visualize the precision of each model--domain effect. Table~\ref{tab:si:behavioral_resilience_uncertainty} reports the corresponding $95\%$ percentile bootstrap confidence intervals for mean $\Delta M$. Mean $\Delta M$ is positive in $83.3\%$ of model--domain settings. The $95\%$ confidence interval is entirely positive in $11$ of $36$ $(30.6\%)$ settings, entirely negative in $1$ of $36$ $(2.8\%)$, and includes zero in the remaining  $24$ of $36$ $(66.7\%)$.

\begin{table*}[tb]
\centering
\small
\setlength{\tabcolsep}{4.5pt}
\renewcommand{\arraystretch}{1.05}
\begin{tabular}{llccc}
\toprule
\textbf{Dataset} & \textbf{Model} & \textbf{Mean $\Delta M$} & \textbf{Bootstrap SE} & \textbf{95\% CI} \\
\midrule
City Locations & llama-3.2-3b & $-0.0256$ & $0.0118$ & $[-0.0488,\,-0.0018]$ \\
 & llama-3.1-8b & $0.0220$ & $0.0308$ & $[-0.0402,\,0.0815]$ \\
 & llama-3.1-70b & $0.0001$ & $0.0049$ & $[-0.0098,\,0.0095]$ \\
 & gemma-7b & $0.0101$ & $0.0080$ & $[-0.0058,\,0.0256]$ \\
 & gemma-2-9b & $0.0180$ & $0.0077$ & $[0.0030,\,0.0329]$ \\
 & gemma-2-27b & $0.0784$ & $0.0144$ & $[0.0497,\,0.1069]$ \\
 & mistral-7b & $0.0426$ & $0.0147$ & $[0.0156,\,0.0730]$ \\
 & mistral-12b & $0.0938$ & $0.0179$ & $[0.0582,\,0.1289]$ \\
 & mistral-3.1-24b & $-0.0012$ & $0.0076$ & $[-0.0160,\,0.0137]$ \\
 & qwen-2.5-7b & $0.0107$ & $0.0095$ & $[-0.0080,\,0.0293]$ \\
 & qwen-2.5-14b & $0.0089$ & $0.0068$ & $[-0.0040,\,0.0227]$ \\
 & qwen-2.5-72b & $0.0107$ & $0.0064$ & $[-0.0015,\,0.0238]$ \\
\addlinespace[2.5pt]
Medical Indications & llama-3.2-3b & $-0.0070$ & $0.0153$ & $[-0.0368,\,0.0231]$ \\
 & llama-3.1-8b & $0.0255$ & $0.0307$ & $[-0.0357,\,0.0860]$ \\
 & llama-3.1-70b & $0.0097$ & $0.0115$ & $[-0.0131,\,0.0323]$ \\
 & gemma-7b & $0.0764$ & $0.0169$ & $[0.0438,\,0.1096]$ \\
 & gemma-2-9b & $0.0522$ & $0.0131$ & $[0.0265,\,0.0778]$ \\
 & gemma-2-27b & $0.0289$ & $0.0167$ & $[-0.0040,\,0.0611]$ \\
 & mistral-7b & $0.0256$ & $0.0213$ & $[-0.0160,\,0.0680]$ \\
 & mistral-12b & $0.0344$ & $0.0173$ & $[0.0003,\,0.0687]$ \\
 & mistral-3.1-24b & $0.0005$ & $0.0076$ & $[-0.0140,\,0.0156]$ \\
 & qwen-2.5-7b & $0.1025$ & $0.0147$ & $[0.0740,\,0.1313]$ \\
 & qwen-2.5-14b & $0.0081$ & $0.0125$ & $[-0.0162,\,0.0328]$ \\
 & qwen-2.5-72b & $-0.0058$ & $0.0112$ & $[-0.0273,\,0.0160]$ \\
\addlinespace[2.5pt]
Word Definitions & llama-3.2-3b & $0.0050$ & $0.0252$ & $[-0.0418,\,0.0561]$ \\
 & llama-3.1-8b & $0.0607$ & $0.0558$ & $[-0.0515,\,0.1659]$ \\
 & llama-3.1-70b & $0.0211$ & $0.0154$ & $[-0.0088,\,0.0510]$ \\
 & gemma-7b & $0.0408$ & $0.0139$ & $[0.0135,\,0.0681]$ \\
 & gemma-2-9b & $0.0346$ & $0.0164$ & $[0.0034,\,0.0672]$ \\
 & gemma-2-27b & $0.0145$ & $0.0228$ & $[-0.0302,\,0.0584]$ \\
 & mistral-7b & $0.0568$ & $0.0281$ & $[0.0009,\,0.1102]$ \\
 & mistral-12b & $0.0257$ & $0.0373$ & $[-0.0471,\,0.0995]$ \\
 & mistral-3.1-24b & $0.0206$ & $0.0148$ & $[-0.0088,\,0.0491]$ \\
 & qwen-2.5-7b & $0.0226$ & $0.0171$ & $[-0.0116,\,0.0555]$ \\
 & qwen-2.5-14b & $-0.0039$ & $0.0114$ & $[-0.0266,\,0.0179]$ \\
 & qwen-2.5-72b & $-0.0096$ & $0.0191$ & $[-0.0476,\,0.0266]$ \\
\bottomrule
\end{tabular}
\caption{\textbf{Uncertainty in behavioral resilience effects.} 
We report the mean behavioral movement difference $\Delta M$, bootstrap standard error, and $95\%$ percentile bootstrap confidence interval using the \texttt{sAwMIL} probe and Direct Conditional graded stability. Uncertainty is estimated from $10{,}000$ sequence-stratified matched-pair bootstrap resamples. Mean $\Delta M$ is positive in $83.3\%$ of model--domain settings; the $95\%$ confidence interval is entirely positive in $30.6\%$ of settings, entirely negative in $2.8\%$, and includes zero in the remaining $66.7\%$.}
\label{tab:si:behavioral_resilience_uncertainty}
\end{table*}

%%%%%%%%%%%%%%%%%%%%%%%%%%%%%%%%%%%%%%%%%%%%%
%%           Appendices: Scaling           %%
%%%%%%%%%%%%%%%%%%%%%%%%%%%%%%%%%%%%%%%%%%%%%

\FloatBarrier

\begin{figure}[ht!]
\centering
\includegraphics[width=\textwidth]{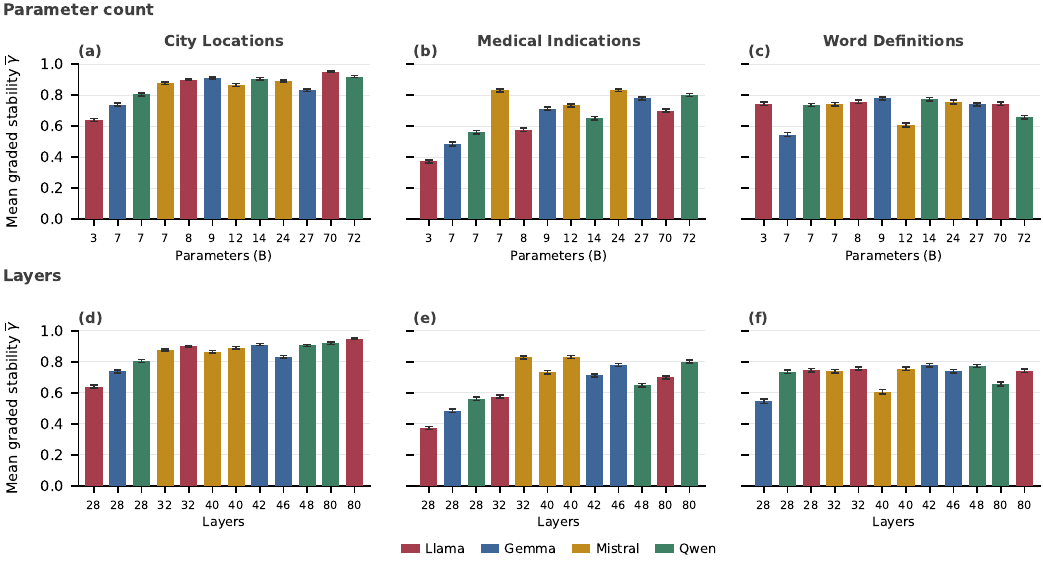}
\caption{
\textbf{Association of model scale with graded stability.} Bars show mean graded stability $\overline{\gamma}_\mathcal{M}$ for the $12$ instruction-tuned LLMs, ordered by nominal parameter count in \textbf{(a)} City Locations, \textbf{(b)} Medical Indications, and \textbf{(c)} Word Definitions, and by number of decoder layers in \textbf{(d)} City Locations, \textbf{(e)} Medical Indications, and \textbf{(f)} Word Definitions. Colors denote \texttt{Llama} (red), \texttt{Gemma} (blue), \texttt{Mistral} (yellow), and \texttt{Qwen} (green) model families. Error bars denote $\pm 1$ bootstrap standard error. Neither parameter count nor model depth shows a consistent relationship with graded stability.
}
\label{si:fig:model_scale}
\end{figure}

\section{Model scaling}
\label{sec:si:scaling}

The between-model variation in graded stability raises the possibility that $\gamma_\mathcal{M}(P)$ increases systematically with model scale. We examine this relationship using two measures of scale: parameter count and number of layers. For each domain, we compute the mean graded stability $\overline{\gamma}$ across each model's beliefs and order the $12$ instruction-tuned LLMs by nominal parameter count (Fig.~\ref{si:fig:model_scale}\textbf{(a--c)}) and model depth (Fig.~\ref{si:fig:model_scale}\textbf{(d--f)}).

We do not observe a consistent scaling relationship under either measure. Mean stability generally increases across some of the smaller models in City Locations and Medical Indications, but these trends are neither monotonic nor consistent across domains. Word Definitions shows little evidence of increasing stability with either parameter count or model depth. Moreover, models with the same nominal parameter count or the same number of layers can exhibit substantially different mean stability. Thus, although model scale may contribute to some of the within-family differences observed in Section~\ref{sec:results:stability_variation}, neither parameter count nor model depth provides a systematic explanation for variation in graded belief stability across models.

%%%%%%%%%%%%%%%%%%%%%%%%%%%%%%%%%%%%%%%%%%%%%
%%      Appendices: Robustness checks      %%
%%%%%%%%%%%%%%%%%%%%%%%%%%%%%%%%%%%%%%%%%%%%%

\FloatBarrier
\section{Robustness checks}
\label{sec:si:robustness}

\subsection{Base-model robustness}
\label{sec:si:base_robustness}

To test whether our primary findings depend on instruction tuning, we repeat the main-text analyses using the pretrained base counterparts of the $12$ instruction-tuned LLMs. The models are matched by family and parameter scale, with exact specifications reported in Section~\ref{sec:si:base_models}.

\begin{figure}[ht!]
\centering
\includegraphics[width=\textwidth]{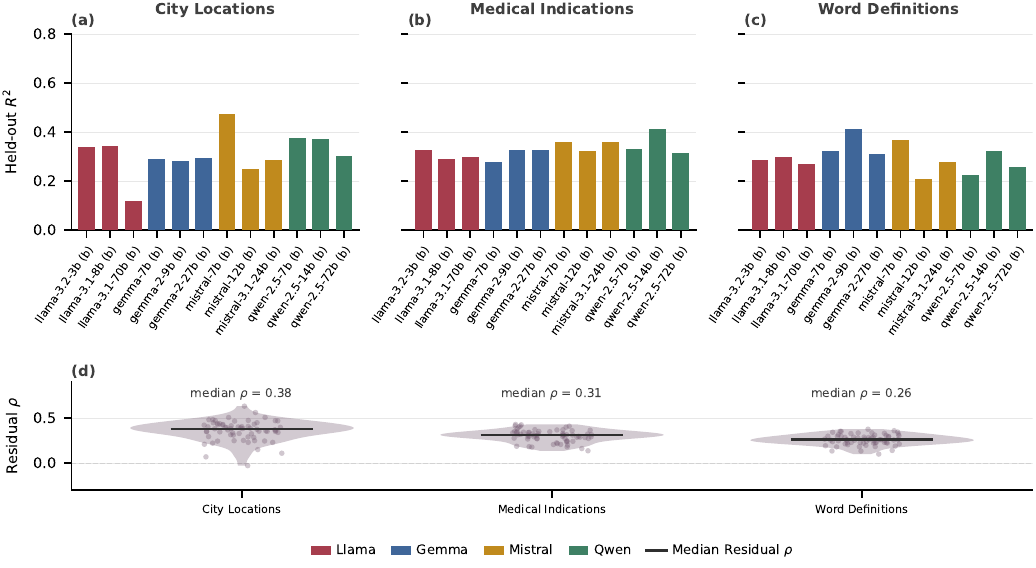}
\caption{
\textbf{Relationship between individual belief probability and graded stability in base models.}
We display the held-out $R^2$ for predicting graded stability from individual belief probability in \textbf{(a)} City Locations, \textbf{(b)} Medical Indications, and \textbf{(c)} Word Definitions for the $12$ matched pretrained base models from the \texttt{Llama} (red), \texttt{Gemma} (blue), \texttt{Mistral} (yellow), and \texttt{Qwen} (green) families. Panel \textbf{(d)} shows the distribution of pairwise Spearman correlations between models' residual graded-stability scores after removing the fitted belief-probability relationship, with black lines denoting median correlations. As with the instruction-tuned models, belief probability explains only part of graded stability and the remaining proposition-level structure is positively shared across models.
}
\label{si:fig:2_base}
\end{figure}

\subsubsection{Replication of primary analyses in base models}
\label{sec:si:base_results}

The principal representation-based findings are qualitatively similar in the base models. Individual belief probability continues to explain a meaningful but incomplete fraction of graded stability (Fig.~\ref{si:fig:2_base}), and the residual stability remaining after accounting for belief probability exhibits positive cross-model structure. Median residual correlations are $\rho=0.38$ for City Locations, $\rho=0.31$ for Medical Indications, and $\rho=0.26$ for Word Definitions.

\begin{figure}[ht!]
\centering
\includegraphics[width=\textwidth]{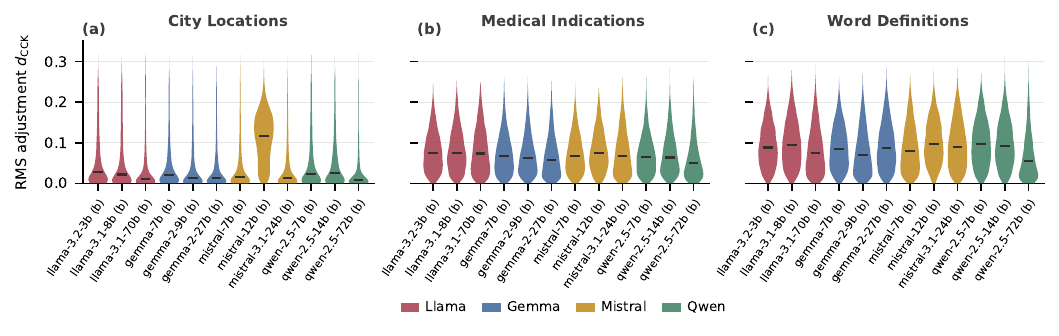}
\caption{
\textbf{Distance from probabilistic coherence in base models.}
Violin plots show the distribution across $(P,x)$ pairs of the RMS adjustment $d_{\mathrm{CCK}}$ required to project the measured probability distributions for individual statements and conditionals onto the nearest CCK-coherent probability system for \textbf{(a)} City Locations, \textbf{(b)} Medical Indications, and \textbf{(c)} Word Definitions for \texttt{Llama} (red), \texttt{Gemma} (blue), \texttt{Mistral} (yellow), and \texttt{Qwen} (green). Horizontal black lines denote within-model medians. Base models remain comparatively close to the coherent set, although distances and between-model heterogeneity are slightly larger than in the instruction-tuned models.
}
\label{si:fig:3_base}
\end{figure}

The approximate-coherence result also persists in the base models (Fig.~\ref{si:fig:3_base}). Most measured probability systems remain relatively close to the CCK-coherent set, with City Locations generally requiring the smallest adjustments and Medical Indications and Word Definitions showing larger distances.

\begin{figure}[ht!]
\centering
\includegraphics[width=\textwidth]{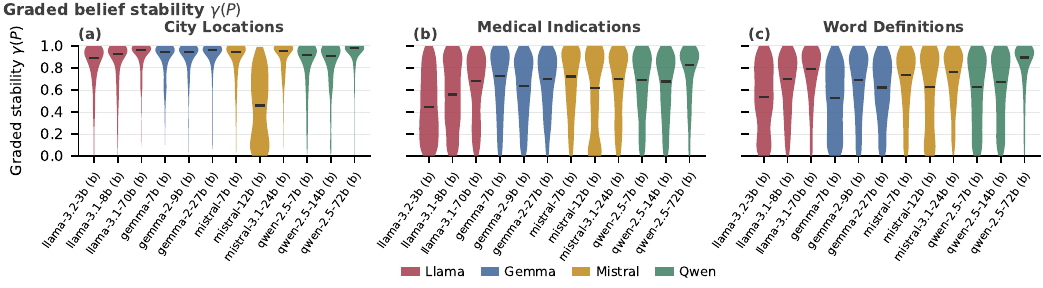}
\caption{
\textbf{Variation in graded belief stability across domains in base models.}
Violin plots show the proposition-level distribution of Direct Conditional graded stability for each of the pretrained base models in \textbf{(a)} City Locations, \textbf{(b)} Medical Indications, and \textbf{(c)} Word Definitions for \texttt{Llama} (red), \texttt{Gemma} (blue), \texttt{Mistral} (yellow), and \texttt{Qwen} (green). Horizontal black lines denote within-model medians. City Locations is generally concentrated at higher stability values, while Medical Indications and Word Definitions show broader and more model-dependent distributions, reproducing the main domain-level pattern with greater heterogeneity.
}
\label{si:fig:4_base}
\end{figure}

Domain-level variation in graded stability is likewise broadly preserved (Fig.~\ref{si:fig:4_base}). City Locations remains concentrated near high stability for most base models, whereas Medical Indications and Word Definitions generally exhibit broader distributions and more mass at intermediate stability values. The base models typically show greater between-model heterogeneity than the instruction-tuned models, particularly for the Medical Indications and Word Definitions domains. There are, however, several notable exceptions to the overall pattern. In City Locations, for example, \texttt{Llama-3.2-3B} is substantially more stable than in the instruction-tuned setting, whereas \texttt{Mistral-12B} is markedly less stable.

\begin{figure}[ht!]
\centering
\includegraphics[width=\textwidth]{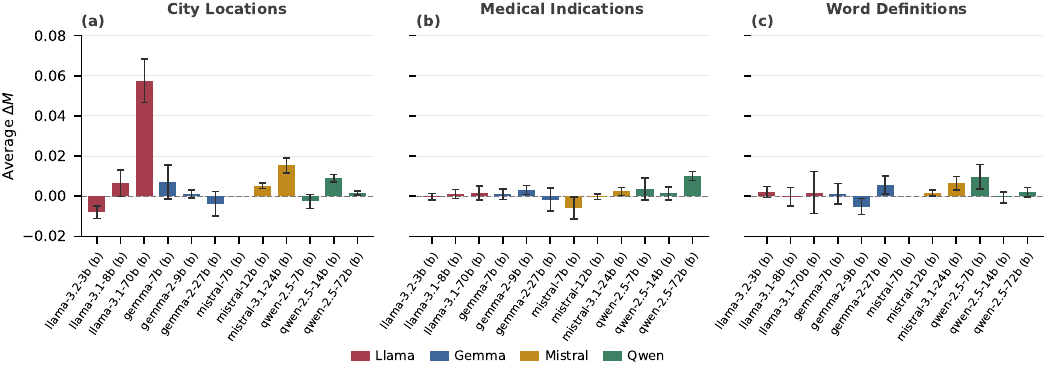}
\caption{
\textbf{Behavioral resilience among probability-matched beliefs in base models.}
Bars show the mean difference in behavioral movement $\Delta M$ between matched lower- and higher-stability beliefs for the pretrained base models in \textbf{(a)} City Locations, \textbf{(b)} Medical Indications, and \textbf{(c)} Word Definitions, with error bars denoting $\pm 1$ bootstrap standard error, for \texttt{Llama} (red), \texttt{Gemma} (blue), \texttt{Mistral} (yellow), and \texttt{Qwen} (green). Positive values indicate greater movement for the lower-stability belief. In contrast to the clearer directional association in instruction-tuned models, base-model effects are smaller and more heterogeneous, suggesting that the behavioral relationship is sensitive to instruction-following post-training.
}
\label{si:fig:5_base}
\end{figure}

The behavioral replication is weaker (Fig.~\ref{si:fig:5_base}). Unlike the instruction-tuned models, for which lower-stability beliefs exhibit greater mean movement in most model--domain settings, the base-model effects are generally smaller and cluster more closely around zero. We interpret this difference cautiously because the behavioral challenge paradigm is itself an instruction-following, multi-turn interaction. The pretrained base models have not undergone the instruction-following post-training of their matched instruction-tuned counterparts, and in our implementation the conversational history is therefore supplied to base models as an explicit plain-text transcript rather than through a model-specific chat template. The weaker base-model behavioral association may therefore reflect the mismatch between the challenge task and the models' training rather than a failure of the underlying representational stability measure.

\subsubsection{Association with instruction tuning}
\label{sec:si:base_vs_instruct}

\begin{figure}[ht!]
\centering
\includegraphics[width=\textwidth]{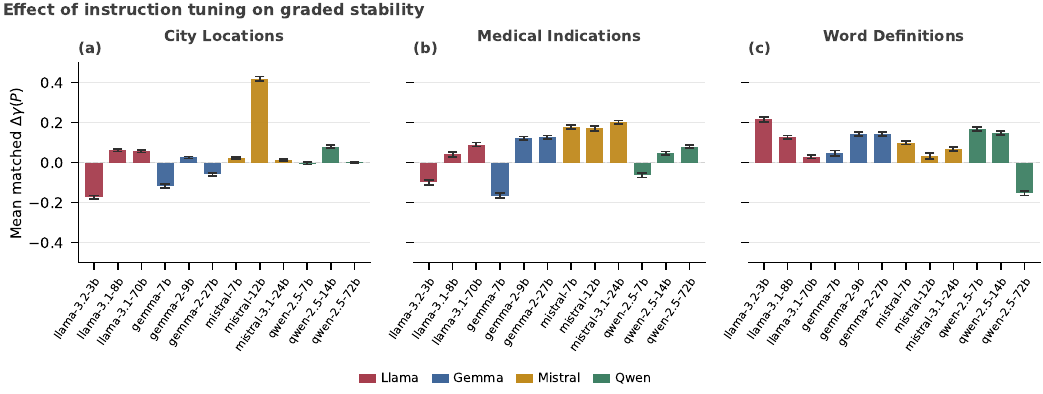}
\caption{
\textbf{Association between instruction tuning and graded belief stability.}
Bars show the mean matched difference $\Delta_{\mathrm{inst}}$ in graded stability between each instruction-tuned model and its pretrained base counterpart over propositions believed by both models for \textbf{(a)} City Locations, \textbf{(b)} Medical Indications, and \textbf{(c)} Word Definitions for \texttt{Llama} (red), \texttt{Gemma} (blue), \texttt{Mistral} (yellow), and \texttt{Qwen} (green). Positive values indicate greater graded stability after instruction tuning, and error bars capture $\pm 1$ standard error. Instruction-tuned models have higher measured stability
in $77.8\%$ of matched comparisons, but the magnitude and direction of the effect remain heterogeneous.
}
\label{si:fig:base_v_instruct}
\end{figure}

We next compare each base model directly with its instruction-tuned counterpart while holding belief identity fixed. For each matched model pair, we restrict the comparison to propositions believed by both models and define the average association with instruction tuning as
\begin{equation}
\label{eq:instruction_tuning_effect}   
    \Delta_{\mathrm{inst}} := \frac{1}{ |\mathcal{B}_{\mathrm{base}} \cap \mathcal{B}_{\mathrm{inst}}| } \sum_{ P\in \mathcal{B}_{\mathrm{base}} \cap \mathcal{B}_{\mathrm{inst}} } \left[ \gamma_{\mathrm{inst}}(P) - \gamma_{\mathrm{base}}(P) \right].
\end{equation}
Thus, $\Delta_{\mathrm{inst}}>0$ indicates that instruction tuning is associated with greater graded stability for the same set of shared beliefs. Across the $36$ matched model--domain comparisons, $28$ $(77.8\%)$ have $\Delta_{\mathrm{inst}}>0$ (Fig.~\ref{si:fig:base_v_instruct}). The effect is therefore positive more often than not, but its magnitude varies substantially across models and domains and several comparisons are negative. We interpret instruction tuning as exhibiting a positive directional tendency rather than a uniform stabilizing effect.

\FloatBarrier
\subsection{Conditional-estimator robustness}
\label{sec:si:joint_robustness}

Our primary analyses use the Direct Conditional estimator, which estimates conditional support from representations of statements of the form ``Given $x$, $P$.'' To test whether our results depend on this operationalization of $\Pr_\mathcal{M}(P\mid x)$, we repeat the principal analyses using the Joint-to-Conditional estimator described in Section~\ref{sec:methods:joint}. While Direct Conditional probes the conditional statement itself, Joint-to-Conditional represents $x$ and $P$ jointly and derives the corresponding conditional probability from the resulting nine-state distribution. We first examine an additional representational choice introduced by the Joint estimator, the ordering of $x$ and $P$ in the conjunction, before replicating the main analyses and directly comparing the two graded-stability estimates.

\begin{figure}[ht!]
\centering
\includegraphics[width=\textwidth]{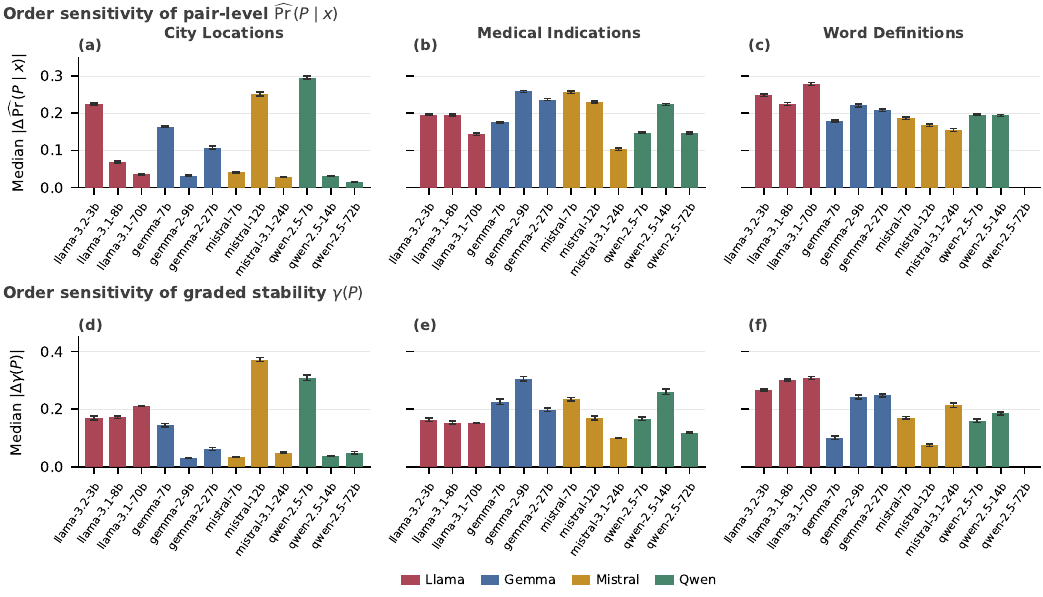}
\caption{
\textbf{Sensitivity of the Joint-to-Conditional estimator to conjunction ordering.}
We compare Joint estimates obtained using the templates ``$x$ and $P$'' and ``$P$ and $x$''. For each model, bars show the median absolute change in $\widehat{\Pr}^{\mathrm{joint}}_\mathcal{M}(P\mid x)$ across evaluated $(P,x)$ combinations when the conjunction order is reversed, for \textbf{(a)} City Locations, \textbf{(b)} Medical Indications, and \textbf{(c)} Word Definitions for \texttt{Llama} (red), \texttt{Gemma} (blue), \texttt{Mistral} (yellow), and \texttt{Qwen} (green). We additionally display \textbf{(d--f)} the median absolute change in $\gamma_\mathcal{M}^{\mathrm{joint}}(P)$ across target beliefs after graded stability is recomputed using the reversed-order estimates. Error bars denote $\pm 1$ bootstrap standard error. Reversing the conjunction order changes both the estimated conditional probabilities and the resulting graded-stability scores, revealing an order sensitivity specific to the Joint-to-Conditional operationalization.
}
\label{si:fig:joint_ordering}
\end{figure}

\subsubsection{Sensitivity to conjunction ordering}
\label{sec:si:joint_ordering}

The Joint-to-Conditional estimator requires a natural-language representation of the two statements being jointly probed. In our primary implementation, we use ``$x$ and $P$'', placing the conditioning proposition $x$ first and the target belief $P$ second. The calibrated Joint probe produces the nine-state distribution described in Section~\ref{sec:methods:joint}, from which we derive $\widehat{\Pr}^{\mathrm{joint}}_\mathcal{M}(P\mid x)$. To test whether this estimate depends on the conjunction order, we independently train and calibrate a second Joint probe using the reversed template ``$P$ and $x$.'' 

Figure~\ref{si:fig:joint_ordering} shows that this choice is consequential. Reversing the conjunction changes the estimated conditional probabilities, with the size of the change varying substantially across models and domains (Fig.~\ref{si:fig:joint_ordering}\textbf{(a--c)}). These differences also affect the downstream graded stability measure (Fig.~\ref{si:fig:joint_ordering}\textbf{(d--f)}). We treat this sensitivity as an additional measurement choice introduced by the Joint-to-Conditional estimator. An autoregressive LLM produces different hidden representations when the same two propositions appear in different token orders, and its training objective does not require the probe-relevant representations of ``$x$ and $P$'' and ``$P$ and $x$'' to be invariant. As such, the observed differences should not necessarily be interpreted as evidence about an underlying order-sensitive joint belief distribution. Rather, they show that the Joint operationalization inherits sensitivity to the linguistic representation used to elicit that distribution. We fix the order to ``$x$ and $P$'' throughout the analyses below, and Direct Conditional remains our primary estimator because it does not introduce this additional conjunction-ordering choice.

\begin{figure}[ht!]
\centering
\includegraphics[width=\textwidth]{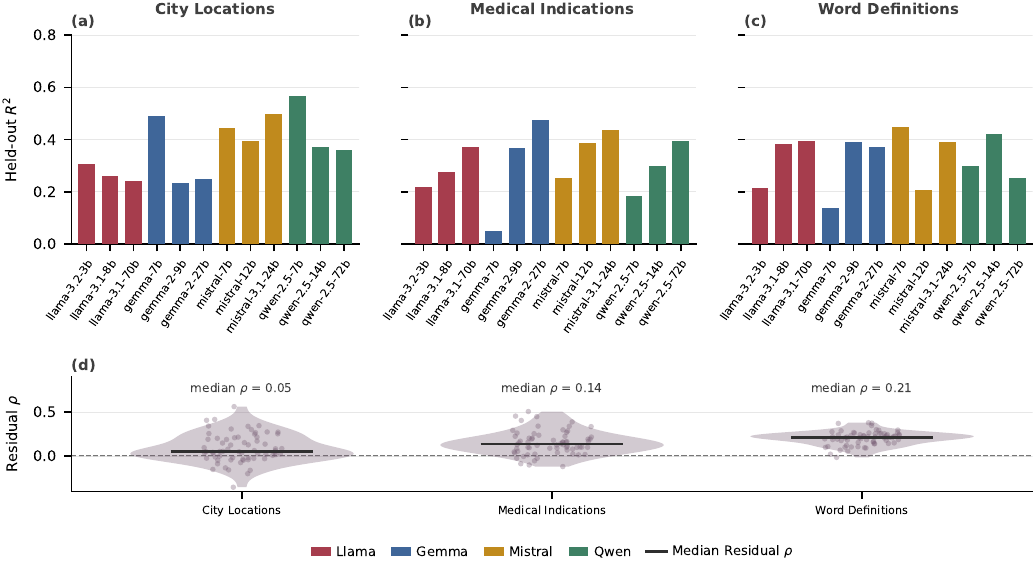}
\caption{
\textbf{Relationship between individual belief probability and Joint-to-Conditional graded stability.}
We display the held-out $R^2$ for predicting Joint-to-Conditional graded stability from individual belief probability in \textbf{(a)} City Locations, \textbf{(b)} Medical Indications, and \textbf{(c)} Word Definitions for the $12$ instruction-tuned LLMs for \texttt{Llama} (red), \texttt{Gemma} (blue), \texttt{Mistral} (yellow), and \texttt{Qwen} (green). Panel \textbf{(d)} shows the distribution of pairwise Spearman correlations between models' residual graded-stability scores after removing the fitted belief-probability relationship, with black lines denoting median correlations. As with Direct Conditional, individual belief probability explains only part of graded stability, and residual stability remains positively correlated across most model pairs.
}
\label{si:fig:2_joint}
\end{figure}

\subsubsection{Replication with Joint-to-Conditional}
\label{sec:si:joint_results}

\begin{figure}[ht!]
\centering
\includegraphics[width=\textwidth]{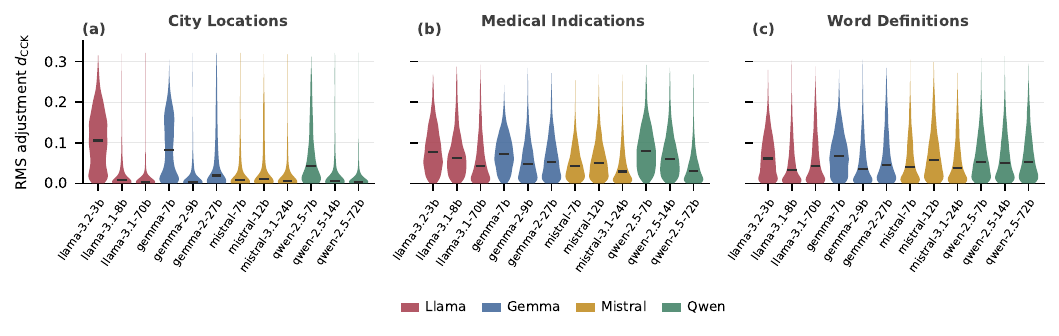}
\caption{
\textbf{Distance of Joint-to-Conditional probability estimates from probabilistic coherence.}
Violin plots show the distribution across $(P,x)$ combinations of the RMS adjustment $d_{\mathrm{CCK}}$ required to project measured probability distributions for individual statements and conditionals onto the nearest CCK-coherent probability system for \textbf{(a)} City Locations, \textbf{(b)} Medical Indications, and \textbf{(c)} Word Definitions for \texttt{Llama} (red), \texttt{Gemma} (blue), \texttt{Mistral} (yellow), and \texttt{Qwen} (green). Horizontal black lines denote within-model medians. As under Direct Conditional, the measured systems generally lie close to the CCK-coherent set, with City Locations showing the smallest distances for most models.
}
\label{si:fig:3_joint}
\end{figure}

\begin{figure}[ht!]
\centering
\includegraphics[width=\textwidth]{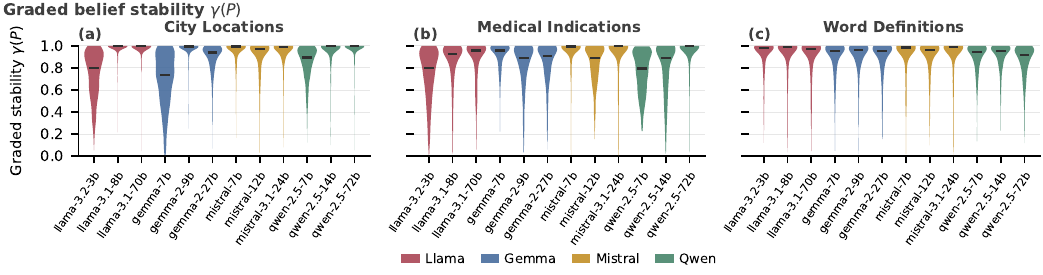}
\caption{
\textbf{Variation in Joint-to-Conditional graded belief stability across domains.}
Violin plots show the proposition-level distribution of Joint-to-Conditional graded stability for each of the $12$ instruction-tuned LLMs in \textbf{(a)} City Locations, \textbf{(b)} Medical Indications, and \textbf{(c)} Word Definitions for \texttt{Llama} (red), \texttt{Gemma} (blue), \texttt{Mistral} (yellow), and \texttt{Qwen} (green). Horizontal black lines denote within-model medians. Joint-to-Conditional stability is strongly concentrated near $\gamma=1$ across all three domains, producing weaker domain separation than under Direct Conditional.
}
\label{si:fig:4_joint}
\end{figure}

\begin{figure}[ht!]
\centering
\includegraphics[width=\textwidth]{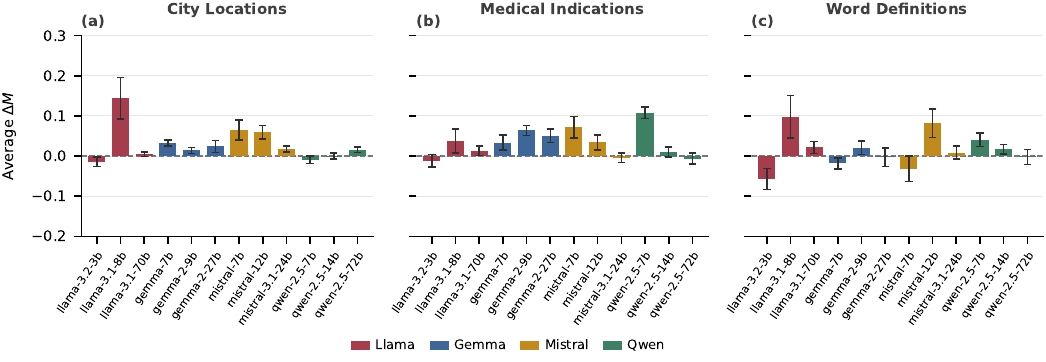}
\caption{
\textbf{Behavioral resilience using Joint-to-Conditional graded stability among probability-matched beliefs.}
Bars show the mean difference in behavioral movement $\Delta M$ between matched lower- and higher-stability beliefs for the $12$ instruction-tuned LLMs in \textbf{(a)} City Locations, \textbf{(b)} Medical Indications, and \textbf{(c)} Word Definitions, with error bars denoting $\pm 1$ bootstrap standard error, for \texttt{Llama} (red), \texttt{Gemma} (blue), \texttt{Mistral} (yellow), and \texttt{Qwen} (green). Positive values indicate greater movement for the lower-stability belief. Mean $\Delta M$ is positive in $26/36$ ($72.2\%$) model--domain settings, although the association is less consistent than under Direct Conditional.
}
\label{si:fig:5_joint}
\end{figure}

We repeat the main-text analyses using Joint-to-Conditional graded stability for the same $12$ instruction-tuned LLMs (Figs.~\ref{si:fig:2_joint}--\ref{si:fig:5_joint}). The principal representation-based findings are broadly preserved, while the absolute scale of graded stability and its behavioral association show greater sensitivity to the conditional estimator.

Individual belief probability remains an incomplete predictor of Joint-to-Conditional graded stability (Fig.~\ref{si:fig:2_joint}). Held-out $R^2$ remains well below one across every model and domain. Cross-model residual correlations are also predominantly positive, although less uniformly than under Direct Conditional: $71.2\%$ are positive for City Locations, $89.4\%$ for Medical Indications, and $98.5\%$ for Word Definitions. The corresponding median residual Spearman correlations are $\rho=0.05$, $\rho=0.14$, and $\rho=0.21$, respectively. Thus, the Joint estimator preserves the central result that graded stability contains systematic variation not captured by individual belief probability, while the strength of the shared residual structure is estimator-dependent.

Using the same CCK projection analysis as in Section~\ref{sec:methods:probabilistic_coherence}, but substituting the Joint-derived conditional estimates, we find that the measured probability systems again lie relatively close to the coherent set (Fig.~\ref{si:fig:3_joint}). Importantly, coherence is not guaranteed by the Joint construction: the individual-statement distributions for $P$ and $x$ are estimated independently rather than obtained as marginals of the probed joint distribution. City Locations is typically closest to the coherent set, while Medical Indications and Word Definitions show larger adjustments.

The clearest estimator-dependent effect appears in the absolute distribution of graded stability (Fig.~\ref{si:fig:4_joint}). Joint-to-Conditional scores are strongly concentrated near $\gamma=1$ across all three domains. City Locations remains highly stable, but Medical Indications and Word Definitions are also shifted substantially toward the upper end of the scale, reducing the domain separation observed under Direct Conditional. This difference is consistent with the construction of the estimators: Joint-to-Conditional renormalizes over determinate conditional states, whereas Direct Conditional allows Neither probability mass to reduce the estimated support for $P\mid x$. The pronounced domain ordering in the primary analysis should therefore be interpreted as partly dependent on the numerical operationalization of conditional support, even though Direct and Joint stability remain strongly related within models (Section~\ref{sec:si:operationalization_agreement}).

The behavioral association is weaker but remains more often positive than negative under Joint-to-Conditional (Fig.~\ref{si:fig:5_joint}). Mean $\Delta M$ is positive in $26/36$ ($72.2\%$) model--domain settings: $10/12$ for City Locations, $9/12$ for Medical Indications, and $7/12$ for Word Definitions. Thus, probability-matched lower-stability beliefs still tend to move more under conversational challenge, but the relationship is less consistent than under Direct Conditional.

\begin{figure}[ht!]
\centering
\includegraphics[width=\textwidth]{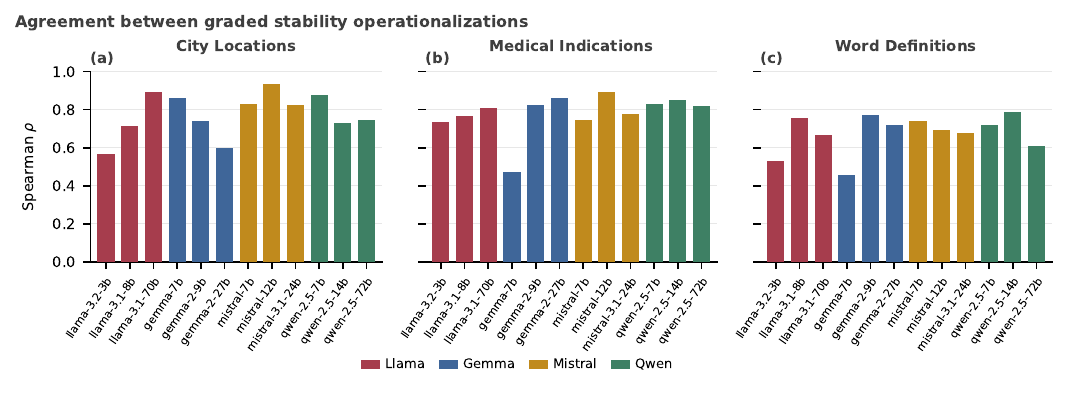}
\caption{
\textbf{Agreement between Direct Conditional and Joint-to-Conditional graded stability.}
Bars show the Spearman correlation between $\gamma_\mathcal{M}^{\mathrm{direct}}(P)$ and $\gamma_\mathcal{M}^{\mathrm{joint}}(P)$ for each instruction-tuned LLM in \textbf{(a)} City Locations, \textbf{(b)} Medical Indications, and \textbf{(c)} Word Definitions for \texttt{Llama} (red), \texttt{Gemma} (blue), \texttt{Mistral} (yellow), and \texttt{Qwen} (green). Correlations are positive for every model and domain, indicating substantial agreement in the relative ranking of beliefs by stability despite differences in the estimators' absolute scales.
}
\label{si:fig:operationalization_agreement}
\end{figure}

\subsubsection{Direct--Joint agreement}
\label{sec:si:operationalization_agreement}

Finally, we ask whether the two estimators preserve the relative ordering of beliefs even where their absolute stability values differ. For each model and domain, we compute the Spearman correlation between $\gamma_\mathcal{M}^{\mathrm{direct}}(P)$ and $\gamma_\mathcal{M}^{\mathrm{joint}}(P)$ over propositions with valid stability estimates under both approaches (Fig.~\ref{si:fig:operationalization_agreement}).

The two operationalizations produce positively correlated stability rankings in every model--domain setting. Median Spearman correlations across models are $\rho=0.78$ for City Locations, $\rho=0.81$ for Medical Indications, and $\rho=0.71$ for Word Definitions. Thus, despite their differences in representation and absolute numerical scale, Direct Conditional and Joint-to-Conditional generally agree on which beliefs are relatively more or less stable within a model.

\FloatBarrier

\begin{table*}[ht!]
\centering
\begin{tabular}{lcccccc}
\toprule
%\textbf{Model}
& \multicolumn{2}{c}{\textbf{City Locations}}
& \multicolumn{2}{c}{\textbf{Medical Indications}}
& \multicolumn{2}{c}{\textbf{Word Definitions}} \\

\textbf{Model}
&
\textbf{SVM} &
\textbf{Mass Mean} &
\textbf{SVM} &
\textbf{Mass Mean} &
\textbf{SVM} &
\textbf{Mass Mean} \\
\midrule
llama-3.2-3b & 11 & 7 & 12 & 11 & 11 & 11 \\
llama-3.1-8b & 25 & 17 & 14 & 13 & 13 & 14 \\
llama-3.1-70b & 76 & 36 & 65 & 37 & 29 & 30 \\
\addlinespace[2pt]
gemma-7b & 17 & 19 & 17 & 17 & 16 & 16 \\
gemma-2-9b & 22 & 23 & 20 & 21 & 20 & 19 \\
gemma-2-27b & 22 & 22 & 21 & 43 & 20 & 22 \\
\addlinespace[2pt]
mistral-7b & 14 & 16 & 14 & 15 & 16 & 15 \\
mistral-12b & 17 & 39 & 16 & 20 & 15 & 19 \\
mistral-3.1-24b & 19 & 21 & 19 & 17 & 18 & 20 \\
\addlinespace[2pt]
qwen-2.5-7b & 20 & 19 & 22 & 18 & 18 & 17 \\
qwen-2.5-14b & 27 & 30 & 31 & 30 & 29 & 23 \\
qwen-2.5-72b & 55 & 63 & 56 & 60 & 56 & 57 \\
\bottomrule
\end{tabular}%
\caption{
\textbf{Selected layers for the alternative probes.}
Zero-indexed layers selected independently for the \texttt{SVM} and \texttt{Mass Mean} probes for each instruction-tuned LLM. The selected probe-specific layer is subsequently used for the corresponding individual belief probability and Direct Conditional analyses.
}
\label{tab:si:selected_layers_other_probes}
\end{table*}

\begin{figure}[ht!]
\centering
\includegraphics[width=\textwidth]{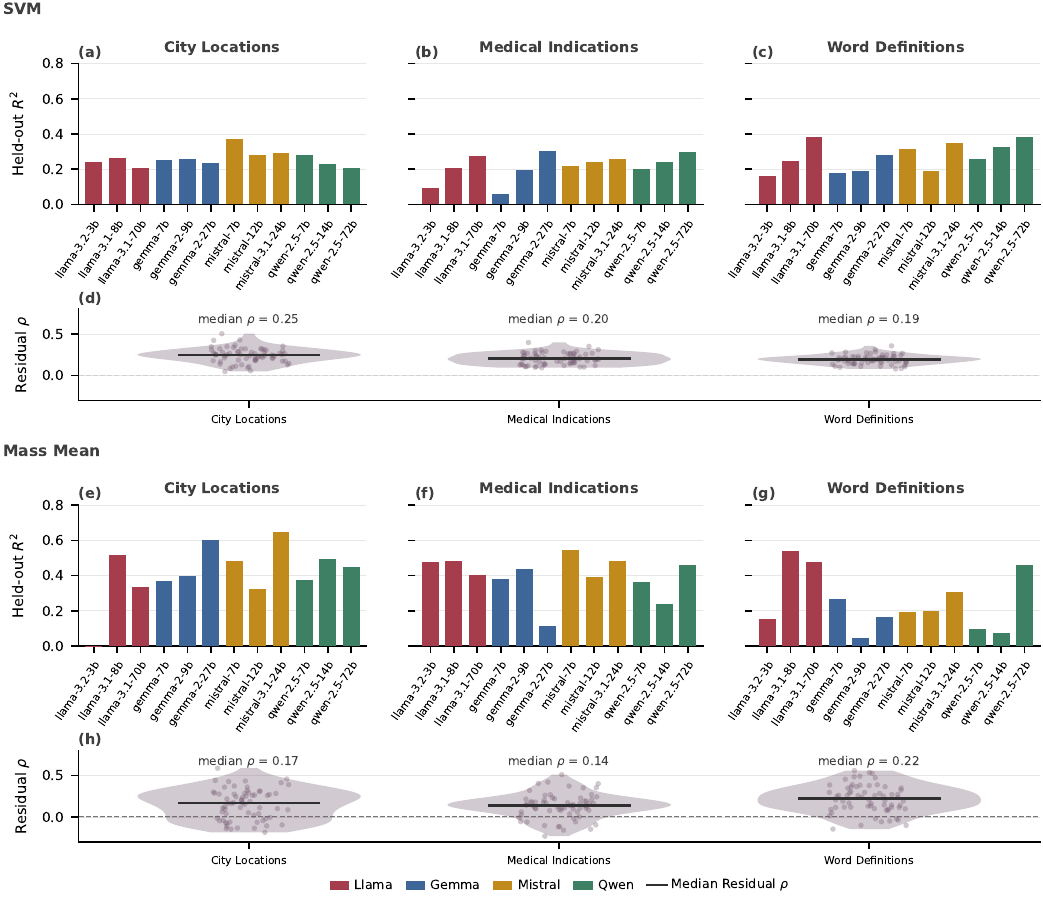}
\caption{
\textbf{Relationship between individual belief probability and graded stability under alternative probes.}
For the \texttt{SVM}, we show held-out $R^2$ for predicting graded stability from individual belief probability in \textbf{(a)} City Locations, \textbf{(b)} Medical Indications, and \textbf{(c)} Word Definitions, respectively, for \texttt{Llama} (red), \texttt{Gemma} (blue), \texttt{Mistral} (yellow), and \texttt{Qwen} (green). Panel \textbf{(d)} shows pairwise cross-model Spearman correlations between residual graded-stability scores. Panels \textbf{(e--h)} report the corresponding \texttt{Mass Mean} results. Black lines in \textbf{(d)} and \textbf{(h)} denote median residual correlations. Under both probes, individual belief probability leaves substantial proposition-level variation unexplained and residual stability remains largely positively shared across models.
}
\label{si:fig:2_probes}
\end{figure}

\begin{figure}[ht!]
\centering
\includegraphics[width=\textwidth]{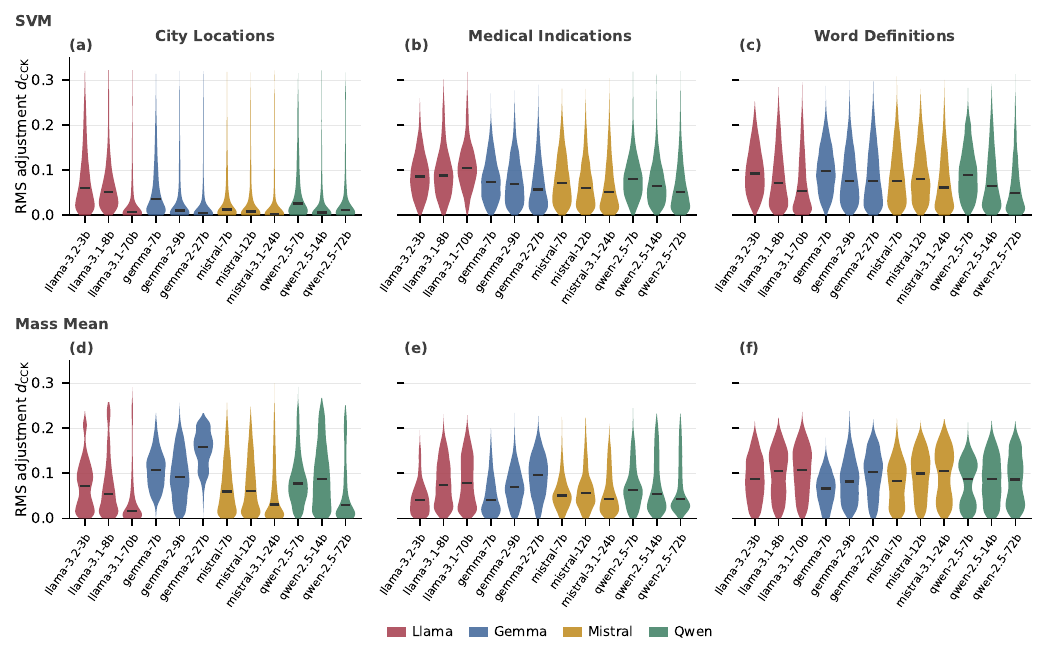}
\caption{
\textbf{Distance from probabilistic coherence under alternative probes.} Violin plots show the RMS adjustment $d_{\mathrm{CCK}}$ required to project measured probability distributions for individual statements and conditionals onto the nearest CCK-coherent system. Panels \textbf{(a--c)} show results for the \texttt{SVM} for \texttt{Llama} (red), \texttt{Gemma} (blue), \texttt{Mistral} (yellow), and \texttt{Qwen} (green), and panels \textbf{(d--f)} for \texttt{Mass Mean}, with columns corresponding to City Locations, Medical Indications, and Word Definitions. Horizontal black lines denote within-model medians. The \texttt{SVM} broadly reproduces the approximate-coherence pattern obtained with \texttt{sAwMIL}, whereas \texttt{Mass Mean} yields larger and more heterogeneous distances from the coherent set.
}
\label{si:fig:3_probes}
\end{figure}

\begin{figure}[ht!]
\centering
\includegraphics[width=\textwidth]{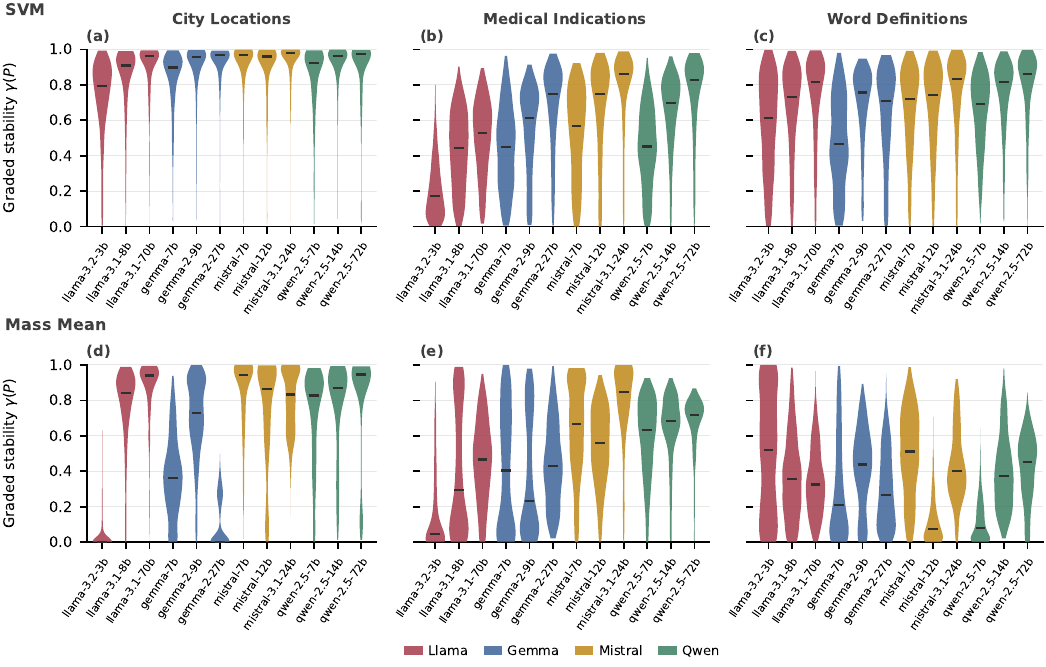}
\caption{
\textbf{Variation in graded belief stability under alternative probes.}
Violin plots show proposition-level Direct Conditional graded-stability distributions. Panels \textbf{(a--c)} show results for the \texttt{SVM} for \texttt{Llama} (red), \texttt{Gemma} (blue), \texttt{Mistral} (yellow), and \texttt{Qwen} (green) and panels \textbf{(d--f)} for \texttt{Mass Mean}, with columns corresponding to City Locations, Medical Indications, and Word Definitions. Horizontal black lines denote within-model medians. The \texttt{SVM} preserves the primary domain-level pattern, whereas \texttt{Mass Mean} produces more heterogeneous distributions and substantially weaker separation across domains.
}
\label{si:fig:4_probes}
\end{figure}

\begin{figure}[ht!]
\centering
\includegraphics[width=\textwidth]{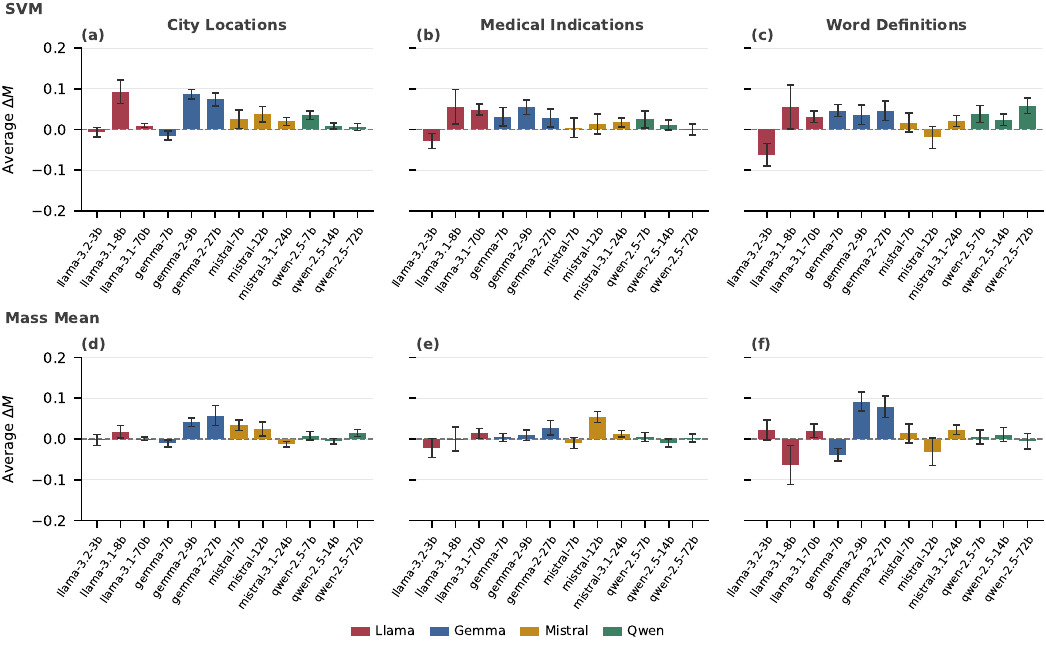}
\caption{
\textbf{Behavioral resilience under alternative probes.}
Bars show the mean behavioral movement difference $\Delta M$ between probability-matched lower- and higher-stability beliefs. Panels \textbf{(a--c)} show results for the \texttt{SVM} for \texttt{Llama} (red), \texttt{Gemma} (blue), \texttt{Mistral} (yellow), and \texttt{Qwen} (green) and panels \textbf{(d--f)} for \texttt{Mass Mean}, with columns corresponding to City Locations, Medical Indications, and Word Definitions. Error bars denote $\pm1$ bootstrap standard error, and positive values indicate greater movement for the lower-stability belief. The directional association remains visible across many \texttt{SVM} model--domain combinations but is weaker and less consistent under \texttt{Mass Mean}.
}
\label{si:fig:5_probes}
\end{figure}

\subsection{Alternative-probe robustness}
\label{sec:si:other_probes}

We repeat the principal analyses using two single-instance linear probes, an \texttt{SVM}~\cite{cortes1995support} and \texttt{Mass Mean}~\cite{marks2024geometry}, to evaluate whether the graded-stability results depend on the \texttt{sAwMIL} probe used in the main text. Both alternative probes operate on the hidden representation of the final non-padding token. For each probe, model, and dataset, we independently select the transformer layer that minimizes three-class calibration log loss using the procedure in Section~\ref{sec:si:layer_selection}. The resulting layers are reported in Table~\ref{tab:si:selected_layers_other_probes} and are subsequently used for both the individual-statement and Direct Conditional analyses.

For the \texttt{SVM} probe, we standardize each activation dimension to zero mean and unit variance using a \texttt{StandardScaler} fit only on the training representations. We then fit one \texttt{LinearSVC} per class in a one-versus-all construction, with $C=1.0$, $\ell_2$ regularization, squared-hinge loss, a maximum of $10{,}000$ iterations, convergence tolerance $10^{-4}$, no class weighting, and random seed $0$.

For \texttt{Mass Mean}, we construct one class-versus-rest direction for each class $j \in \{0,\ldots,K-1\}$. Let $\boldsymbol{\mu}_j$ denote the mean training representation for class $j$ and $\boldsymbol{\mu}_{\neg j}$ the mean training representation over all remaining classes. We compute the difference vector
\begin{equation}
    \Delta\boldsymbol{\mu}_j
    =
    \boldsymbol{\mu}_j-\boldsymbol{\mu}_{\neg j},
\end{equation}
and normalize $\Delta\boldsymbol{\mu}_j$ to the unit $\ell_2$ norm before scoring. The decision boundary is placed halfway between $\boldsymbol{\mu}_j$ and $\boldsymbol{\mu}_{\neg j}$, and statements are scored by their signed projection along the normalized $\Delta\boldsymbol{\mu}_j$ direction relative to this midpoint.

For both probes, the resulting class scores are converted to probability distributions using the multinomial logistic-regression calibration procedure described in Section~\ref{sec:si:probability_calibration}. The calibrator is fit only on the calibration split with $C=1.0$, the \texttt{L-BFGS} solver, a maximum of $5{,}000$ iterations, and tolerance $10^{-6}$. Direct Conditional probes are trained separately from the corresponding individual-statement probes on the same conditional training and calibration examples used for \texttt{sAwMIL}, while retaining the probe-specific layer selected from the sweep.

Across these analyses, the \texttt{SVM} largely reproduces the principal representation-based results obtained with \texttt{sAwMIL}, whereas \texttt{Mass Mean} is less consistent across several downstream analyses. Both alternative probes nevertheless preserve the central result that graded stability is not fully explained by the fitted relationship with individual belief probability (Fig.~\ref{si:fig:2_probes}). For the \texttt{SVM}, median cross-model residual correlations are $\rho=0.25$, $0.20$, and $0.19$ for City Locations, Medical Indications, and Word Definitions, respectively. The corresponding \texttt{Mass Mean} medians are $\rho=0.17$, $0.14$, and $0.22$. Thus, under both alternative probes, substantial proposition-level variation remains after accounting for individual belief probability, and models continue to show positively correlated residual stability.

The probabilistic-coherence results show a clearer difference between the two alternative probes (Fig.~\ref{si:fig:3_probes}). The \texttt{SVM} yields distance-to-coherence distributions broadly similar to those obtained with \texttt{sAwMIL}. By contrast, \texttt{Mass Mean} generally requires larger adjustments to reach the CCK-coherent set, particularly for City Locations and Word Definitions.

The domain-level characterization shows the same pattern (Fig.~\ref{si:fig:4_probes}). Under the \texttt{SVM}, City Locations remains concentrated near high graded stability, while Medical Indications and Word Definitions exhibit broader distributions extending further into intermediate and low stability values. This separation is substantially weaker under \texttt{Mass Mean}, with several City Locations models in particular receiving much lower graded-stability estimates than under either \texttt{sAwMIL} or the \texttt{SVM}.

Finally, the behavioral association is more robust under the \texttt{SVM} than under \texttt{Mass Mean} (Fig.~\ref{si:fig:5_probes}). \texttt{SVM}-based graded stability remains predominantly associated with greater behavioral resilience among probability-matched beliefs, although the effects are more heterogeneous than in the primary \texttt{sAwMIL} analysis. Under \texttt{Mass Mean}, the estimated effects are generally smaller and less consistent in direction.

The weaker \texttt{Mass Mean} results are consistent with known sensitivities of centroid-based truth directions. \texttt{Mass Mean} was introduced as a binary method that estimates a truth direction from the difference between the centroids of \texttt{True} and \texttt{False} representations~\cite{marks2024geometry}. Previous work applying this probe in a three-class setting found that it can become unstable when \texttt{Neither} examples are incorporated, since changes in the composition of the representation sets directly alter the estimated class centroids~\cite{dies2025}. Our trivalent implementation similarly constructs three class-versus-rest centroid directions for \texttt{True}, \texttt{False}, and \texttt{Neither} and calibrates the resulting scores into a shared multiclass probability distribution.

This sensitivity is especially important for graded stability because our downstream quantities depend on the absolute allocation of probability mass across all three classes, not merely on whether a binary truth direction ranks \texttt{True} statements above \texttt{False} statements. \texttt{Mass Mean} can therefore remain informative as a veracity probe while being less well suited to the trivalent probability estimation required here. We consequently view the strong \texttt{SVM} replication as evidence that the principal results are not specific to the multi-instance structure of \texttt{sAwMIL}, while the \texttt{Mass Mean} results identify a meaningful sensitivity to probes that are not naturally designed for multiclass probability estimation.

\end{appendices}

\end{document}